\documentclass[twoside,11pt]{article}

\usepackage{blindtext}

\usepackage{jmlr2e}
\usepackage{amsmath}
\usepackage{graphicx}
\usepackage{multirow}
\usepackage{tikz-dependency}
\usepackage{diagbox}
\usepackage{array}
\usepackage{algorithm}
\usepackage{algorithmicx}
\usepackage{algpseudocode}
\usepackage{mathtools}
\usepackage{algorithm}
\usepackage{amsmath,amssymb}
\usepackage{multicol}
\usepackage{fdsymbol}
\usepackage{multirow}
\usepackage{mathtools}
\usepackage[normalem]{ulem}
\usepackage{hyperref}
\usepackage{xcolor}
\usepackage{subcaption}
\usepackage{booktabs}
\usepackage{lastpage}

\newcolumntype{H}{@{}>{\lrbox0}l<{\endlrbox}}

\definecolor{darkblue}{rgb}{0, 0, 0.5}
\hypersetup{colorlinks=true,citecolor=darkblue, linkcolor=darkblue, urlcolor=darkblue}

\usepackage{lipsum}

\newcommand{\reals}{\mathbb{R}}

\newcommand{\zz}{\mathbf{z}} %  noise
\newcommand{\ttt}{\mathbf{t}}

\newcommand{\modelM}{\mathbb{M}}

\usepackage{amsmath,amsfonts,bm}

\def\eqref#1{equation~\ref{#1}}
\def\1{\bm{1}}

\DeclareMathAlphabet{\mathsfit}{\encodingdefault}{\sfdefault}{m}{sl}
\SetMathAlphabet{\mathsfit}{bold}{\encodingdefault}{\sfdefault}{bx}{n}

\def\sT{{\mathbb{T}}}

\def\sW{{\mathbb{W}}}

\jmlrheading{24}{2023}{1-\pageref{LastPage}}{1/23; Revised
10/23}{10/23}{23-0074}{Nadir Durrani, Fahim Dalvi and Hassan Sajjad}
\ShortHeadings{Discovering Salient Neurons in Deep NLP Models}{Durrani, Dalvi and Sajjad}
\firstpageno{1}

\begin{document}

\title{Discovering Salient Neurons in deep NLP models}

\author{\name Nadir Durrani \email ndurrani@hbku.edu.qa \\
       \addr Qatar Computing Research Institute \\
       Hamad Bin Khalifa University \\
       Doha, Qatar
       \AND
       \name Fahim\ Dalvi \email faimaduddin@hbku.edu.qa \\
       \addr Qatar Computing Research Institute \\
       Hamad Bin Khalifa University \\
       Doha, Qatar
              \AND
       \name Hassan \ Sajjad \email hsajjad@dal.ca \\
       \addr Faculty of Computer Science \\
       Dalhousie University \\
       Halifax, Nova Scotia, Canada
      }

\editor{Alexander Clark}

\maketitle

\begin{abstract}%   <- trailing '%' for backward compatibility of .sty file

While a lot of work has been done in understanding representations learned within deep NLP models and what knowledge they capture, work done towards analyzing individual neurons is relatively sparse. We present a technique called \textbf{Linguistic Correlation Analysis} to extract salient neurons in the model, with respect to any extrinsic property, with the goal of understanding how such knowledge is preserved within neurons. We carry out a fine-grained analysis to answer the following questions: (i) can we identify subsets of neurons in the network that learn a specific linguistic property? (ii) is a certain linguistic phenomenon in a given model localized (encoded in few individual neurons) or distributed across many neurons? (iii) how redundantly is the information preserved? (iv) how does fine-tuning pre-trained models towards downstream NLP tasks impact the learned linguistic knowledge? (v) how do models vary in learning different linguistic properties? Our data-driven, quantitative analysis illuminates interesting findings: (i) we found small subsets of neurons that can predict different linguistic tasks; (ii) neurons capturing basic lexical information, such as suffixation, are localized in the lowermost layers; (iii) neurons learning complex concepts, such as syntactic role, are predominantly found in middle and higher layers; (iv) salient linguistic neurons are relocated from higher to lower layers during transfer learning, as the network preserves the higher layers for task-specific information; (v) we found interesting differences across pre-trained models regarding how linguistic information is preserved within them; and (vi) we found that concepts exhibit similar neuron distribution across different languages in the multilingual transformer models. Our code is publicly available as part of the NeuroX toolkit \citep{dalvi-etal-2023-neurox}.\footnote{\url{https://github.com/fdalvi/NeuroX}}
\end{abstract}


\section{Introduction}

The advent of deep neural networks and their opaque nature has prompted a new field of research focused on analyzing these models. However, this revolution comes at the expense of transparency. Unlike traditional models that rely on manually designed features, we now face a lack of understanding regarding two key aspects: i) what knowledge the underlying representations capture, and ii) how this information is utilized in the decision-making process of deep NLP models. Addressing these questions is crucial for understanding the inner workings of neural networks and ensuring fairness, trust, and accountability when applying AI solutions.
To gain insights into these questions, researchers have pursued two main avenues of investigation: i) Concept Analysis and ii) Attribution Analysis. The former thrives on post-hoc decomposability, analyzing representations to uncover linguistic (and non-linguistic) phenomena learned as the network is trained for any NLP task \citep{belinkov:2017:acl,conneau2018you,tenney-etal-2019-bert}. The latter characterizes the role of model components in a specific prediction \citep{linzen_tacl,gulordava-etal-2018-colorless,marvin-linzen-2018-targeted}. Recently, efforts have been made to connect these two lines of work and determine whether the encoded knowledge is actually used by the model \citep{feder-etal-2021-causalm,elazar-etal-2021-amnesic}.

\textbf{Concept Analysis} aims at understanding the learned representations with respect to human interpretable concepts. For example analyzing if the model is aware of language morphology, anaphora or more subtle concepts like gender etc. Such an analysis can be potentially useful for not only debugging the systems for errors \citep{MODE,lertvittayakumjorn-etal-2020-find}, but improving the design choices and system performance \citep{Wambsganss2021ImprovingEA}. To this end the researchers have analyzed contextualized representations trained within deep NLP models w.r.t concepts varying from word morphology \citep{vylomova2016word,dalvi:2017:ijcnlp,durrani-etal-2019-one} to high level concepts such as structure \citep{shi-padhi-knight:2016:EMNLP2016,linzen2016assessing}  and semantics \citep{qian-qiu-huang:2016:P16-11,belinkov:2017:ijcnlp} or more generic properties such as sentence length \citep{adi2016fine}. Compared to the holistic view that we obtain from representation analysis, neuron analysis offers several benefits. By analyzing individual neurons, we gain a more fine-grained understanding of the network's representations. This allows us to identify specific neurons or groups of neurons that capture particular phenomena or concepts \citep{karpathy2015visualizing,li-etal-2016-visualizing,kadar-etal-2017-representation,lakretz-etal-2019-emergence}. Such insights go beyond understanding the mechanics of neural networks and have numerous applications. For example understanding which neurons are responsible for certain behaviors enables us to intervene and manipulate the network's output accordingly \citep{bau2018identifying}. Neuron analysis can facilitate model distillation by identifying less salient neurons \citep{rethmeier2019txray}. It can contribute to efficient feature-based transfer learning \citep{dalvi-2020-CCFS} by understanding the functional role of different neurons. Finally, how information is distributed across the architectures can guide architecture search. 

We present a method called as the \textbf{\emph{Linguistic Correlation Analysis (LCA)}}, to extract salient neurons in the model with respect to any extrinsic property. Our methodology is a 3-step process --- given a neural network model: i) use the pre-trained model to extract contextualized feature representations for words, ii) train a classifier using the extracted features to predict a property of interest, iii) extract a ranking based on the weights assigned to each neuron. The ranking serves as a proxy to measure the importance of neurons with respect to the understudied property. We select top $N\%$ neurons from the ranked list, that maintain oracle performance i.e. accuracy on the task using the entire representation. 

The \texttt{LCA} method is based on the diagnostic classifier framework \citep{Hupkes,belinkov:2017:acl}. However, rather than examining the overall representations learned by the neural network, we use it to discover individual neurons w.r.t their role in encoding specific linguistic features.

We apply the \texttt{LCA} method to delve into the core linguistic phenomena encompassing morphology, syntax, and semantics, essential aspects of language learning, deemed crucial for any sophisticated NLP task. However, our method can be seamlessly used for any task-specific concept where annotated data is available. We analyze the representations trained from  neural language models of which we study the recent transformer based models. We use \texttt{LCA} to understand: i) how concepts are learned within neurons and ii) how these neurons distribute across the network. More specifically we probe for the following questions:

\begin{itemize}
    \item  Question: Do the individual neurons in the transformers capture linguistic information and which parts of the network learn more about certain linguistic phenomena?
    \item Finding: Neurons that capture tasks such as predicting shallow lexical information (such as suffixes) or word morphology are predominantly found in the lower layers and those learning more complex properties such as syntactic information are found in the higher layers. \textbf{[Section \ref{sec:layerWise}]}
    \item Question: Is certain linguistic phenomenon in a given model localized or distributed across many neurons? And how redundantly is the information preserved?
    \item Finding: Closed-class properties like \emph{interjections} (morphological category) and \emph{Months of year} (semantic category) are concentrated within a smaller number of neurons, whereas open-class properties such as nouns (morphological category) and \emph{location} (semantic category) are spread across multiple neurons. \textbf{[Section \ref{sec:propertyWise}]}
    \item Finding: A subset of neurons that learn a particular concept can be discovered, however, concepts are redundantly learned and many subsets of neurons that carry the same information coexist in the network. \textbf{[Sections \ref{sec:ablation},\ref{sec:redundancy}]}

    \item Question: How do various architectures differ in learning these properties?
    \item Finding: Some models (such as BERT) are more distributed while others (such as XLNet) exhibit disjoint set of neurons  designated for different linguistic properties. \textbf{[Section \ref{sec:layerWise}]} 
    \item Question: How does the information redistributes across the layers during transfer learning? 
    \item Finding: The differences between architectures became more pronounced when the models were tuned towards the GLUE tasks. We observed that linguistic knowledge regresses to lower layers in RoBERTa and XLNet, unlike BERT, where it is still preserved in the higher layers. \textbf{[Section \ref{sec:fineTuning}]}
    \item Question: Do multilingual models with shared knowledge exhibit distinct behavior?
    \item Finding: Multilingual models exhibit a similarity in distributing neurons across layers, when learning concepts in different languages. \textbf{[Section \ref{sec:multilingual}]}
    
\end{itemize}

\noindent This paper builds upon the work presented in \citet{dalvi:2019:AAAI} by expanding on several aspects: i) we extend the analysis towards transformer models including multilingual BERT and XLM-RoBERTa \citep{conneau2020unsupervised}, ii) we used the algorithm to discover minimal set of salient neurons that learn a linguistic property and carry out several analyses on top of this. For example: a) to understand how salient neurons distribute across the layers and linguistic properties, b) how redundantly is the information preserved, c) how it redistributes when the models are fine-tuned towards a downstream task. 

% This article is organized into the following sections. Section \ref{sec:methodology} describes \emph{Linguistic Correlation Analysis} method. Section \ref{sec:setup} describes our experimental setup, the linguistic tasks and pre-trained models studied in this work. Section \ref{sec:eval} validates the correctness of our method.
% Section \ref{sec:analysis} provides analysis of our results and the discusses the findings. Section \ref{sec:literature} provides an account of the related work and the last section concludes the paper.

\section{Linguistic Correlation Analysis}
\label{sec:methodology}

\label{sec:approach-supervised}
\begin{figure}[t]
	\centering
	\includegraphics[scale=0.60]{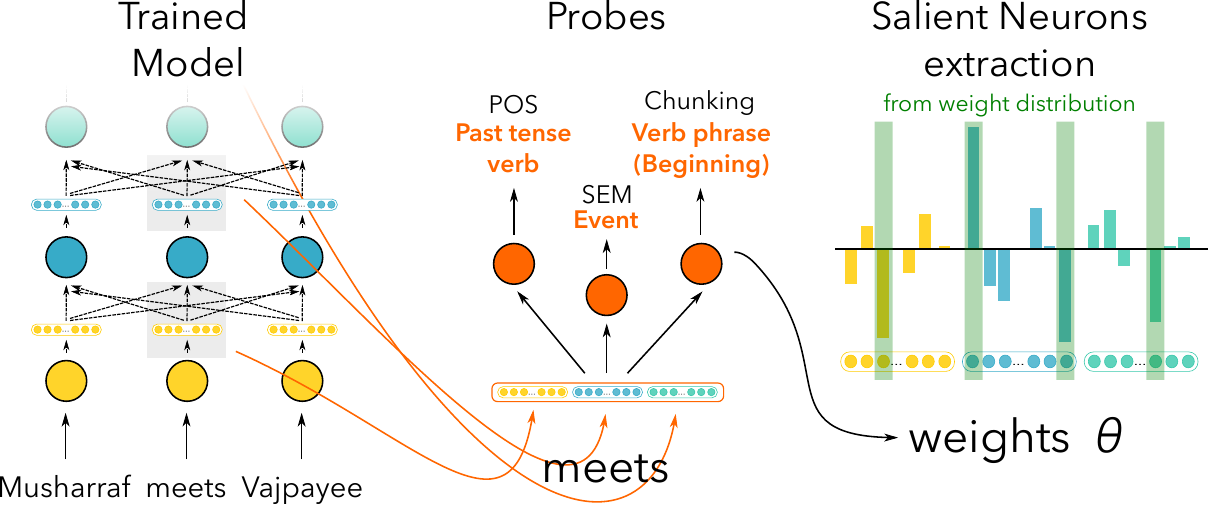}
	\caption{Linguistic Correlation Analysis: Extract neuron activations from a trained model, train a classifier and use weights of the classifier to extract salient neurons.}
    %\vspace{-4mm}
	\label{fig:supervised-method}
\end{figure}

The  LCA algorithm follows a three step process (See Figure \ref{fig:supervised-method}): i) first we extract activations from the trained neural network model, ii) we then use these as static features to train an auxiliary classifier towards an extrinsic property of interest, %Once the classifier is trained, 
iii) lastly we use %its
the weights of the trained classifier as a proxy to the describe saliency of a neuron w.r.t the %extrinsic 
property. 

\begin{algorithm}[h]
	\caption{Probe training (Section 2.1)}
	\label{alg:probe_training}
	\begin{algorithmic}[1]
        \Function{TrainProbe}{$X$, $y$, $\lambda_1$, $\lambda_2$}
            \State Initialize learning rate $\eta=0.001$, number of epochs $N=10$
            \State Initialize parameters $\theta$ (weights $W$ and bias $b$) randomly
            \For{epoch $ = 1$ to $N$}
                \For{each training example $(w_i, t_{w_i})$ in $X$} \Comment{Can be done over mini-batches as well}
                    \State $L(\theta) = -\sum_i \log P_{\theta}(t_{w_i} | w_i) + \lambda_1 \|\theta\|_1 + \lambda_2 \|\theta\|^2_2$ \Comment{Equation from \ref{subsec:model}}
                    \State $d\theta \gets \frac{\partial L(\theta)}{\partial \theta}$
                    \Comment{Compute the gradient of the loss function with respect to $W$ and $b$}
                    \State $\theta \gets \theta - \eta \cdot d\theta$
                    \Comment{Update the parameters $W$ and $b$}
                \EndFor
            \EndFor
            \State \Return $\mathrm{probe_\theta}$ \Comment{trained probe}
        \EndFunction
    \end{algorithmic}
\end{algorithm}

\subsection{Model}
\label{subsec:model}

\begin{algorithm}[h]
    \caption{Grid Search (Section 2.2)}
    \label{alg:grid_search}
    \begin{algorithmic}[1]
        \Function{GridSearch}{$X$, $y$, $\alpha$, $\beta$, $M$}
            \State searchList $\gets $ [$1^{-6}, 1^{-5}, 1^{-4}, 1^{-3}, 1^{-2}, 1^{-1}, 0$]
            \State $\lambda^*_1, \lambda^*_2 \gets $ null
            \Comment{Best regularization parameters}
            \State bestScore $\gets -\infty$
            \For{$\lambda_1$ in searchList}
                \For{$\lambda_2$ in searchList}
                    \State probe $\gets $ \Call{TrainProbe}{$X$, $y$, $\lambda_1$, $\lambda_2$}
                    \State rankedNeurons $\gets$ \Call{GetNeuronRanking}{probe}
                    \State $\mathrm{X_{top}} \gets $ \Call{AblateBottomNeurons}{$X$, rankedNeurons, $M$}
                    \Comment{Set all neurons to zero except top $M$ neurons}
                    \State $\mathrm{X_{bottom}} \gets $ \Call{AblateTopNeurons}{$X$, rankedNeurons, $M$}
                    \Comment{Set all neurons to zero except bottom $M$ neurons}
                    \State $\mathrm{probe_{top}} \gets $ \Call{TrainProbe}{$\mathrm{X_{top}}$, $y$, $\lambda_1$, $\lambda_2$}
                    \State $\mathrm{probe_{bottom}} \gets $ \Call{TrainProbe}{$\mathrm{X_{bottom}}$, $y$, $\lambda_1$, $\lambda_2$}
                    \State $\mathrm{probe_{noreg}} \gets $ \Call{TrainProbe}{$X$, $y$, 0, 0}
                    \State score $\gets $ $\alpha \cdot $ (\Call{A}{$\mathrm{probe_{top}}$} - \Call{A}{$\mathrm{probe_{bottom}}$}) - $\beta \cdot $ (\Call{A}{$\mathrm{probe_{noreg}}$} - \Call{A}{probe}) 
                    \Comment{\textbf{A} computes the accuracy of the given probe}
                    \If {score $>$ bestScore}
                        \State bestScore $\gets$ score
                        \State $\lambda^*_1 \gets \lambda_1$
                        \State $\lambda^*_2 \gets \lambda_2$
                    \EndIf
                \EndFor
            \EndFor
            \State \Return $\lambda^*_1, \lambda^*_2$
        \EndFunction
    \end{algorithmic}
\end{algorithm}

Formally, consider a pre-trained neural language model $\mathbf{M}$ with $L$ layers: $\{l_1, l_2, \ldots, l_L\}$. Given a dataset $\sW=\{w_1, w_2, ..., w_N\}$ %consisting of $N$ words
with %a 
the corresponding %set of linguistic 
annotations $\sT=\{t_{w_1}, t_{w_2}, ..., t_{w_N}\}$, we map each word $w_i$ in the data $\sW$ to a sequence of latent representations: $\sW \overset{\modelM}{\mapsto} \zz = \{\zz_1, \dots, \zz_n\}$. The representations can %either 
be extracted from the entire model, %or just 
from an individual layer or any component of the model (attention head, FFN layer etc). The probe  is trained by minimizing the following loss function:
%
%\vspace{-2mm}
\begin{equation}
\mathcal{L}(\theta) = -\sum_i \log P_{\theta}(t_{w_i} | w_i) + \lambda_1 \|\theta\|_1 + \lambda_2 \|\theta\|^2_2 \nonumber
\end{equation}
%\vspace{-2mm}
%
where $P_{\theta}(t_{w_i} | w_i) = \frac{\exp (\theta_l \cdot \zz_i)}{\sum_{l'} \exp (\theta_{l'} \cdot \zz_i)}$
 is the probability that word $i$ is assigned property $t_{w_i}$. The weights \mbox{$\theta \in \reals^{D \times |T|}$} are learned with gradient descent (Algorithm \ref{alg:probe_training}). Here $D$ is the dimensionality of the latent representations $\zz_i$ and $T$ is the set of tags (properties) in the linguistic tag set, which the classifier is predicting, $|T|$ is the number of such tags. The trained probe can also be represented as a function $X \mapsto y$ where $X$ is the combination of all the word representations ($w_i$) and $y$ is the list of associated labels from $T$.
 The terms $\lambda_1 \|\theta\|_1$ and $\lambda_2 \|\theta\|^2_2$ correspond to $L1$ and $L2$ regularization. This combination also known as elastic-net \citep{Zou05regularizationand}. It is useful to strike a balance between identifying very focused localized features ($L1$) versus  distributed neurons ($L2$). We use a \textbf{grid search} algorithm  to find the most appropriate set of lambda values. 

\subsection{Grid Search}
\label{sec:search}

To find the best values for $\lambda_1$ and $\lambda_2$, we conduct a weight ablation experiment on the trained classifier using a search algorithm (refer to Algorithm \ref{alg:grid_search}). Following classifier training, we identify the top and bottom $M$ features from a ranked list generated by a neuron ranking algorithm. The remaining features are then set to zero. The score for each pair of ($\lambda_1$, $\lambda_2$) is calculated as follows:

\begin{equation}
{S}(\lambda_1, \lambda_2) =  \alpha ( A_t - A_b) - \beta (A_z - A_l) \nonumber
\end{equation}

\noindent here, $A_t$ represents the accuracy of the classifier retaining top neurons and masking the rest, $A_b$ is the accuracy retaining bottom neurons, $A_z$ stands for the accuracy of the classifier trained using all neurons but without regularization, and $A_l$ represents the accuracy with the current lambda set. The first term ensures the selection of a lambda set where accuracies of top and bottom neurons are significantly different, while the second term ensures a preference for weights that result in minimal loss in classifier accuracy due to regularization.\footnote{Higher L1 values encourage sparsity in the classifier, leading to the selection of a smaller set of focused neurons. However, this often results in poorer task performance. We set $M=20\%$ and $\alpha,\beta=0.5$.}

\begin{algorithm}[h]
    \caption{Neuron Ranking Extraction (Section 2.3)}
    \label{alg:ranking_extraction}
    \begin{algorithmic}[1]
        \Function{GetNeuronRanking}{probe}
            \State ordering $\gets$ []
            \Comment{{\footnotesize ordering will store the neurons in order of decreasing importance}}
            \For{$p=1$ \textbf{to} $100$ \textbf{by} $\alpha$}
                \Comment{{\footnotesize $p$ is the percentage of the weight mass. We start with a very small value and incrementally move towards 100\%.}}
                \State tnpt $\gets$ \Call{GetTopNeuronsPerTag}{$\theta$, $p$}
                \Comment{{\footnotesize tnpt contains the top neurons per tag using the threshold $p$}}
                \State topNeurons $\gets \bigcup\limits_{i=1}^{L} \mathrm{tnpt_{i}}$
                \State newNeurons $\gets$ topNeurons $\setminus$ ordering
                \State ordering.append(newNeurons)
            \EndFor
            \State \Return ordering
        \EndFunction
    \end{algorithmic}
\end{algorithm}

\subsection{Neuron Ranking Algorithm}
\label{sec:algorithm}

Deriving neuron rankings is analogous to subset selection, which is a well-studied field in machine learning. Classical subset selection methods either enumerate all possible subsets to select the best features, or perform step wise selection of features by training a model to find the next best feature at each step \cite[Section 3.3]{hastie2013elements}. However, these techniques become infeasible when the number of total features is large, as in the case of neuron selection. In this work, we derive inspiration of Shrinkage Methods \cite[Section 3.4]{hastie2013elements} and use the internal weights of a trained classifier to extract top neurons (and subsequently a ranking of neurons). This avoids any expensive computation per feature, and also allows the usage of a single trained model to extract independent rankings for each class in a multi-class scenario. 

Formally, given the trained weights of the classifier denoted as $\theta \in \reals^{D \times |T|}$, our objective is to establish a ranking for the $D$ neurons within the model $\modelM$. For a specific property $\ttt \in T$, we sort the weights $\theta_{\ttt} \in \reals^D$ based on their absolute values, sorting them in descending order. Consequently, the neuron with the highest absolute weight in $\theta_{\ttt}$ is positioned at the top of our ranking. We identify the top $n$ neurons (pertaining to the specific property in consideration) that cumulatively contribute to some percentage of the total weight mass as \textit{salient neurons}.
% Formally, given 
% the trained weights of the classifier \mbox{$\theta \in \reals^{D \times |T|}$}, we want to extract a ranking of the $D$ neurons in the model $\modelM$. For the property $\ttt \in T$, we sort the weights $\theta_{\ttt} \in \reals^D$ by their absolute values in descending order. Hence the neuron with the highest corresponding absolute weight in $\theta_{\ttt}$ appears at the top of our ranking. We consider the top $n$ neurons (for the individual property under consideration) that cumulatively contribute to some percentage of the total weight mass as \textit{salient neurons}. %\sout{In order to get a set of \textit{salient neurons} from this ranking, we consider the top $n$ neurons that cumulatively contribute to some percentage of the total weight mass.}
%Next, 
%In order 
To extract a ranking of neurons w.r.t. all of the tags in $T$, we use an iterative process  described in Algorithm \ref{alg:ranking_extraction}. We start with a small percentage of the total weight mass and choose the most salient neurons for each tag $\ttt$, and increase this percentage iteratively, adding newly discovered top neurons to our ordering. Hence, the salient neurons for each label $\ttt$ appears at the top of the ordering. The order in which the neurons are discovered indicates their importance to the property set $T$.

\begin{algorithm}[h]
    \caption{Minimal Neuron Subset (Section 2.4)}
    \label{alg:minimal_neuron_subset}
    \begin{algorithmic}[1]
        \Function{GetMinimumNeurons}{$X$, $y$, $\delta$}
            \State $\lambda^*_1, \lambda^*_2 \gets$ \Call{GridSearch}{$X$, $y$, $0.5$, $0.5$, $20\%$}
            \State probe $\gets $ \Call{TrainProbe}{$X$, $y$, $\lambda^*_1$, $\lambda^*_2$}
            \State $\mathrm{O_{acc}} \gets $ \Call{A}{probe} \Comment{Oracle accuracy}
            \State rankedNeurons $\gets$ \Call{GetNeuronRanking}{probe}
            \State $N \gets 0$ \Comment{Number of selected neurons}
            \Repeat
                \State $N \gets N+1$
                \State $\mathrm{X_{top}} \gets $ \Call{SelectTopNeurons}{$X$, rankedNeurons, $N$}
                \Comment{Select top $N$ neurons}
                \State $\mathrm{probe_{selected}} \gets $ \Call{TrainProbe}{$\mathrm{X_{top}}$, $y$, $\lambda^*_1$, $\lambda^*_2$}
                \State $\mathrm{S_{acc}} \gets $ \Call{A}{$\mathrm{probe_{selected}}$} \Comment{Selected neurons accuracy}
            \Until{$N < $ length(rankedNeurons) or {$\mathrm{O_{acc}} - \mathrm{S_{acc}} < \delta$}}
            \State \Return N
        \EndFunction
    \end{algorithmic}
\end{algorithm}

\subsection{Minimal Neuron Selection} 
\label{sec:minimumNeurons}

%Once we have obtained the best regularization lambdas, 

Our approach involves an iterative process to identify minimal neurons for any downstream task: i) train a classifier to predict the task using all the neurons (call it \emph{Oracle}), ii) obtain a neuron ranking based on the ranking algorithm described above,  iii) select the top $N$ neurons from the ranked list and retrain a classifier using these selected neurons. This process is repeated, with $N$ gradually increasing by $1\%$ at each iteration, until the classifier achieves an accuracy close to, but not less than a specified threshold compared to the performance of the Oracle. In our experiments, we utilized a threshold of $1\%$ loss compared to Oracle, to guide this iterative process.

\subsection{Selectivity}
\label{sec:selectivity}

A potential pitfall in the approach of probing classifiers is the challenge of determining whether the probe faithfully reflects the underlying property of the representation or simply learns the specific task. \citet{hewitt-liang-2019-designing} introduced control tasks to investigate the influence of training data and lexical memorization on probing experiments. They propose \emph{Selectivity} as a criterion to put a ``linguistic task's accuracy in context with the probe’s capacity to memorize from word types''. An effective probe should exhibit high accuracy in the linguistic task and low accuracy in the control task. We used selectivity to ensure that our probes are a true reflection of the learned representation.

The control tasks for our probing classifiers are defined by mapping each word type $x_i$ to a randomly sampled behavior $C(x_i)$, from a set of numbers \{$1 \dots T$\} where $T$ is the size of tag set to be predicted in the linguistic task. The sampling is done using the empirical token distribution of the linguistic task, so the marginal probability of each label is similar. The selectivity is defined as: $\delta = A_t - A_{t_c}$ i.e. difference of classifier accuracy when trained towards the task at hand and accuracy when trained towards control task drawn from the same distribution.

\section{Experimental Setup}
\label{sec:setup}

\vspace{2mm}
\noindent \textbf{Pre-trained Neural Language Models:}
We experimented with 3 transformer models: BERT \citep{devlin-etal-2019-bert}, RoBERTa \citep{liu2019roberta} and XLNet \citep{yang2019xlnet} using the base versions (13 layers and 768 dimensions). This architectural choice allowed for a compelling comparison between auto-encoder and auto-regressive models.  Additionally, we explored cross-language experiments by employing mBERT and XLM-RoBERTa \citep{conneau2020unsupervised}, the multilingual variants of BERT and RoBERTa, respectively.

\begin{table*}[t]
        \caption{Example sentence with different word-level annotations.}
	\centering
	%\footnotesize
    % The CCG supertags are taken from \citep{W17-4707}; POS and semantic tags are our own annotation.}
	\resizebox{\columnwidth}{!}{
	\begin{tabular}{c|cccccccc}
	\toprule
		Words & Musharraf  & meets & Vajpayee  & in  & the & city & of  & Islamabad \\ %& reading \\
		\midrule
		Suffix & -- & s & -- & -- & -- & -- & -- & -- \\
		POS  & NP & VBZ & NP & IN & DT & NN & IN & NP \\ % & VBG \\
		SEM  & PER & ENS & PER & REL & DEF & REL & REL & GEO \\ %& EXS \\ 
		Chunk & B-NP & B-VP & B-NP & B-PP & B-NP & I-NP & B-PP & B-NP \\
		CCG & NP & ((S[dcl]$\backslash$NP)/PP)/NP & NP  & PP/NP  & NP/N  & N  & (NP$\backslash$NP)/NP  & NP
        \\
        %\midrule
       	\bottomrule
	\end{tabular}
	}
    % \caption{Example sentence with different word-level annotations.}
	\label{tab:example-annotation}
\end{table*}

\begin{figure}[t]
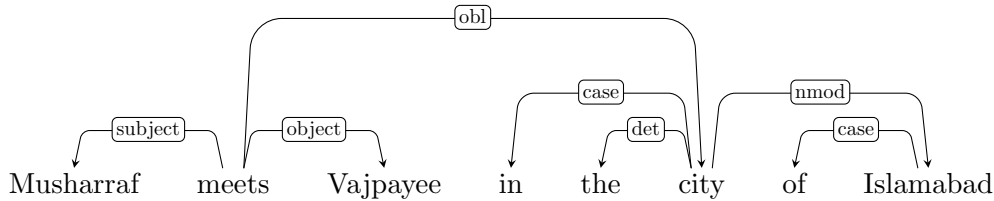

  \centering
    \begin{dependency}[theme=default]%[arc edge, text only label, label style={above}]
      \begin{deptext}[column sep=.6cm, font=\normalsize]
      Musharraf \& meets \& Vajpayee \& in \& the \& city \& of \& Islamabad \\
      \end{deptext}
      \depedge{2}{1}{subject} %nsubj}
      \depedge{2}{3}{object}
      \depedge{6}{4}{case}% mark}
      \depedge{6}{5}{det} %obj}
      \depedge{2}{6}{obl} %conj}
      \depedge{8}{7}{case}% cc}
      \depedge{6}{8}{nmod}
    \end{dependency}
  \caption{Syntactic relations according to the Universal Dependencies formalism. Here ``Musharraf'' and ``Vajpayee'' are the subject and object of ``meets'', respectively, \texttt{obl} refers to an oblique relation of the locative modifier, \texttt{nmod} denotes the genitive relation, the prepositions ``in'' and ``of'' are treated as case-marking elements, and ``the'' is a determiner. See \url{https://universaldependencies.org/guidelines.html} for detailed definitions. }
  \label{fig:syn}  
\end{figure}

%\paragraph{Language Tasks:} 
\vspace{2mm}
\noindent \textbf{Language Tasks:}
We evaluated our method on 6  linguistic tasks: suffix prediction, parts-of-speech tagging using 
the Penn TreeBank \citep{marcus-etal-1993-building}, syntactic chunking using CoNLL 2000 shared task dataset \citep{tjong-kim-sang-buchholz-2000-introduction}, CCG super-tagging using CCGBank \citep{hockenmaier2006creating}, syntactic dependency labeling with Universal Dependencies data-set %\cite{UD-11234/1-1983}
and semantic tagging using the Parallel Meaning Bank data \citep{abzianidze-EtAl:2017:EACLshort}. See Table \ref{tab:example-annotation} and Figure \ref{fig:syn} for sample annotations for different tasks. We used standard splits for training, development and test data (See Table \ref{tab:dataStats} in Appendix). 
For multilingual experiments, we annotated a small portion of multi-parallel news data \citep{bojar-etal-2014-findings} for English, German and French using RDRPOSTagger \citep{nguyen-EtAl:2014:Demos} (See Table \ref{tab:dataStats2} in Appendix for statistics).

\vspace{2mm}

\noindent \textbf{Transfer Learning:} For the transfer-learning experiments, the models (BERT, XLNet, and RoBERTa) were fine-tuned towards GLUE tasks. Specifically, we experimented with SST-2 for sentiment analysis with the Stanford sentiment tree-bank \citep{socher-etal-2013-recursive}, MNLI for natural language inference \citep{williams-etal-2018-broad}, QNLI for Question NLI \citep{rajpurkar-etal-2016-squad}, RTE for recognizing textual entailment \citep{Bentivogli09thefifth}, MRPC for Microsoft Research paraphrase corpus \citep{dolan-brockett-2005-automatically}, and STS-B for the semantic textual similarity benchmark \citep{cer-etal-2017-semeval}. All models were fine-tuned using identical settings, and we conducted three independent runs to identify the best-performing model.

\vspace{2mm}

\noindent \textbf{Classifier Settings:} We utilized a linear probing classifier with elastic-net regularization, employing a categorical cross-entropy loss function and optimizing it with Adam \citep{kingma2014adam}. The training process involved shuffled mini-batches of size 512 and was stopped after 10 epochs. The regularization weights were trained using a grid-search algorithm. In the case of sub-word based models, we considered the last activation value as the representative of the word, as also done for the embeddings \citep{liu-etal-2019-linguistic,durrani-etal-2019-one}. Linear classifiers are commonly chosen for analyzing deep NLP models due to their enhanced interpretability \citep{qian-qiu-huang:2016:EMNLP2016, belinkov-etal-2020-analysis}. \cite{hewitt-liang-2019-designing} have also demonstrated that linear probes exhibit higher ``Selectivity'' (which we also observed), a desirable characteristic for more interpretable probes. Linear probes are especially important for our method since we employ the learned weights as a measure of each neuron's importance.

\section{Evaluation}
\label{sec:eval}

Before presenting our analysis, we evaluate the efficacy of our neuron ranking. We validate \texttt{LCA} using i) ablation study, ii) classifier retraining, iii) selectivity, and iv) via qualitative evaluation.

\subsection{Ablation Study}
\label{sec:ablation}

\begin{table}[t]
\caption{Ablation Study: Selecting all, top, random and bottom 20\% neurons and zeroing-out remaining to evaluate classifier accuracy on blind test (averaged over 3 runs). In the last column we select 20\% neurons based on weight magnitude of the neurons.}
\centering					
\footnotesize
%\resizebox{\columnwidth}{!}{									
    \begin{tabular}{l|ccc|ccc|ccc}									
    \toprule									
        & \multicolumn{3}{c}{\textbf{POS}} & \multicolumn{3}{c}{\textbf{SEM}} & \multicolumn{3}{c}{\textbf{Chunking}} \\  		
    \midrule
    & BERT & XLNet & RoBERTa & BERT & XLNet & RoBERTa & BERT & XLNet & RoBERTa \\
    \toprule
    All & 96.16 & 96.19 & 96.72 & 92.11  & 92.68 & 93.13 & 95.01 & 94.15 & 95.09 \\
    Top & 90.16 & 92.28 & 91.93 & 84.32 & 90.70 & 90.16  & 89.01 & 89.16 & 89.63\\
    Random & 28.45 & 58.17 & 44.56 & 64.28 & 72.14 & 66.15 & 75.83 & 75.26 & 75.30  \\
    Bottom & 16.86 & 44.64 & 23.09 & 59.02 & 25.37 & 34.35 & 66.82 & 46.66 & 54.11 \\
    \midrule
    Wt. Mag. & 67.89 & 64.73 & 71.06  & 61.56 & 68.86 & 64.79 & 68.31 & 75.44 & 73.22 \\
    \bottomrule
    \end{tabular}
\label{tab:classifier_ablation_mask_out}						    %\vspace{-4mm}
\end{table}

 Given a classifier trained to predict a linguistic task $T$, we extract a ranked list of neurons and conduct ablation experiments to validate the rankings. In these experiments, we ``zero-out'' all activations in the test, except for the selected $M\%$ neurons. We evaluate the rankings by retaining the top, random, and bottom 20\% neurons. Additionally, we establish another baseline in which we selected neurons based on their weight magnitude and ablated the remaining neurons. It is worth noting that the choice of 20\% is arbitrary and not experimentally explored; it was used solely to demonstrate the effectiveness of the rankings and to select the best lambdas (see Section \ref{sec:search}). 
 
 Table \ref{tab:classifier_ablation_mask_out} illustrates the efficacy of our rankings. Retaining the top 20\% neurons yields high prediction accuracy, in contrast to retaining the bottom or random 20\% neurons. Notably, the accuracy of the \texttt{Random} neurons, although significantly lower compared to the \texttt{Top} neurons, is still relatively high in some cases (e.g., for the ``Chunking'' task, which is related to predicting syntax). We observed that when the underlying task is complex, the related information is more distributed across the network, leading to redundancy. The issue of redundancy in pre-trained models will be revisited in Section \ref{sec:redundancy}.  This \texttt{Weight Magnitude} baseline demonstrated similar performance to randomly selecting neurons in most cases.

\begin{table}[t]									
\centering
\caption{Selecting minimal number of neurons for each downstream NLP task. Accuracy numbers reported on blind test-set (averaged over three runs) -- Neu$_t$ = Percentage of top selected neurons, Acc$_a$ = Accuracy using all neurons, Acc$_t$ = Accuracy using selected neurons after retraining the classifier using selected neurons, Sel = Difference between linguistic task and control task accuracy when classifier is trained on all neurons (Sel$_a$) and top neurons (Sel$_t$).%\textbf{hs: remove accuracy numbers from the table}
}
\footnotesize
%\resizebox{\columnwidth}{!}{									
    \begin{tabular}{l|ccc|ccc|ccc}									
    \toprule									
        & \multicolumn{3}{c}{\textbf{POS}} & \multicolumn{3}{c}{\textbf{SEM}} & \multicolumn{3}{c}{\textbf{Chunking}} \\   		
    \midrule
    & BERT & XLNet & RoBERTa & BERT & XLNet & RoBERTa & BERT & XLNet & RoBERTa \\
    \toprule
    Neu$_t$ & 5\% & 5\% & 5\% & 5\% & 5\% & 5\% & 10\% & 10\%  & 10\% \\
    \midrule
    Acc$_a$ & 96.16 & 96.19 & 96.72 & 92.11 & 92.68 & 93.13 & 95.09 & 94.15 & 94.68 \\
    Acc$_t$ & 95.92 & 96.49 & 96.48 & 92.18 & 92.72 & 92.96 & 94.99 & 94.62 & 94.80 \\
    \midrule
    Sel$_a$ & 14.49 & 23.53 & 17.93 & 14.08 & 12.98 & 13.33 & 16.30 & 22.77 &  18.39 \\
    Sel$_t$ & 31.84 & 31.98 & 31.91 & 27.17 & 26.83 & 27.00 & 29.19 & 28.42 & 28.39 \\
    \midrule
    Probeless & 95.45 & 96.18 & 94.90 & 90.87 & 91.35 & 89.99 & 90.36 & 91.80 & 91.16 \\
    %Wt. Mag. & 93.87 & 93.87 & 94.19 & 90.78 & 88.61 & 89.65 & 90.20 & 91.29 & 90.75 \\
    \bottomrule
    \end{tabular}
 %   }
%  \caption{Selecting minimal number of neurons for each downstream NLP task. Accuracy numbers reported on blind test-set (averaged over three runs) -- Neu$_t$ = Percentage of top selected neurons, Acc$_a$ = Accuracy using all neurons, Acc$_t$ = Accuracy using selected neurons after retraining the classifier using selected neurons, Sel = Difference between linguistic task and control task accuracy when classifier is trained on all neurons (Sel$_a$) and top neurons (Sel$_t$).%\textbf{hs: remove accuracy numbers from the table}
% }							
\label{tab:accuracy}					    %\vspace{-4mm}
\end{table}

\subsection{Minimal Neuron Subset}
\label{sec:minimalNeuronRetrain}

In the previous section, we conducted an experiment where we masked out most of the classifier neurons, leaving only 20\% of them, in order to evaluate the accuracy of the rankings. Now we leverage these rankings to select the minimum number of neurons required for each linguistic task. To achieve this, we employ Algorithm \ref{alg:minimal_neuron_subset} to obtain a ranking of neurons. We then iteratively select a percentage, denoted as $N\%$, of the highest-ranked neurons that achieve comparable accuracy (using a threshold $\delta = 1.0$) to using the entire network (all the features). By analyzing these significant neurons, we can highlight two aspects: i) the parts of the learned network that predominantly capture different linguistic phenomena, and ii) the extent to which information is localized or distributed across different properties. A summary of the results is provided in Table~\ref{tab:accuracy}.

Firstly, we demonstrate that for all tasks, selecting a subset of the top N\% neurons and retraining the classifier can achieve similar or even better accuracy compared to using all the neurons (Acc$_a$) as static features. In the case of lexical tasks such as \textbf{POS} or \textbf{SEM} tagging, a very small number of neurons (approximately 5\%, i.e., 499 out of 9984\footnote{When training the classifier, we create a large feature vector (768 $\times$ 13 -- hidden dimensions $\times$ number of layers) by concatenating the layers.}) were found to be sufficient to achieve a comparable accuracy (Acc$_t$) to the oracle accuracy (Acc$_a$). On the other hand, syntactic tasks like \textbf{Chunking} tagging, which model non-local dependencies, required a larger set of neurons (999 out of 9984) to achieve the same level of accuracy.
We also examined more complex syntactic tasks, namely CCG super tagging and Syntactic dependency relations, which are discussed later in the paper. These tasks required an even larger set of neurons (15\%) to achieve performance close to the oracle accuracy. Additionally, we included %two other baselines 
another baseline for comparison. We employed the Probeless method \citep{antverg2022on} to generate rankings of neurons for each task and subsequently retrained the classifier by selecting the top N\% neurons based on these rankings. Our findings demonstrate that the Probeless method exhibited similar performance to our approach in the POS tagging task, while our method (LCA) outperformed Probeless in the other two tasks.

%and dependency labeling %\cite{UD-11234/1-1983} and found that more complex tasks require even larger set of neurons. Please see Appendix for results if interested. 

\subsection{Selectivity}

% Although selectivity is not directly used to evaluate the efficacy of our ranking, it is generally considered important to demonstrate that the probe is truly reflecting the knowledge contained within the understudied representation and not its ability to memorize the task.  We use \emph{Selectivity} to further demonstrate that our probes (trained using the entire representation and selected neurons) do not memorize from word types but highlight the knowledge captured within the underlying representation. Recall that an effective probe is recommended to achieve high linguistic task accuracy and low control task accuracy.  The results (see Table \ref{tab:accuracy}) show that selectivity with top neurons ($Sel_t$) is much higher than selectivity with all neurons $Sel_a$. It is evident that using all the neurons may contribute to memorization whereas higher selectivity with selected neurons indicates less memorization and efficacy of our neuron selection. 

Although selectivity is not directly used to evaluate the efficacy of our ranking, it is generally considered important to demonstrate that the probe truly reflects the knowledge contained within the understudied representation and not its ability to memorize the task. We use \emph{Selectivity} to further demonstrate that our probes (trained using the entire representation and selected neurons) do not memorize word types but highlight the knowledge captured within the underlying representation. Recall that an effective probe is recommended to achieve high linguistic task accuracy and low control task accuracy. The results (see Table \ref{tab:accuracy}) show that selectivity with top neurons ($Sel_t$) is much higher than selectivity with all neurons ($Sel_a$). It is evident that using all the neurons may contribute to memorization, whereas higher selectivity with selected neurons indicates less memorization and the efficacy of our neuron selection.

%We achieve high selectivity when selecting 500 neurons as in the case of POS and SEM. The chunking  requires double the neurons. 

\subsection{Qualitative Evaluation}
\label{sec:qa}

Visualizations have been used effectively to gain qualitative insights on analyzing neural networks \citep{karpathy2015visualizing,kadar-etal-2017-representation}. We carry a qualitative evaluation and analysis of the selected neurons using the \texttt{NeuroX} visualization tool \citep{dalvi-etal-2023-neurox}. We do this by visualizing: i) the activations of the top neurons w.r.t a property, ii) or analyzing the top $n$ words that  activates a neuron. 

\begin{figure*}[!ht]
	\centering
    \begin{subfigure}[b]{0.65\linewidth}
    % \centering
    \includegraphics[width=0.85\linewidth]{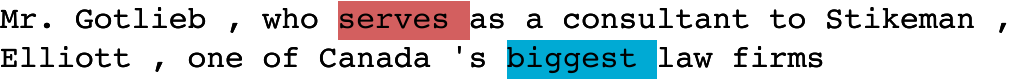}
    
    \vspace{0.2cm}
    
    \includegraphics[width=0.95\linewidth]{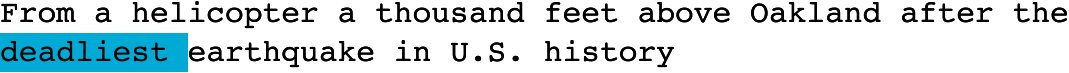}
    
    \vspace{0.2cm}
    
    \includegraphics[width=1\linewidth]{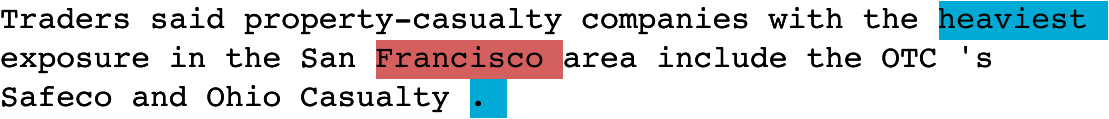}
    \caption{\label{fig:jjs} Superlative Adjective Neuron (ROBERTA) -- Layer 1: 193}
    \end{subfigure}
     \begin{subfigure}[b]{0.65\linewidth}
    % \centering
    \vspace{0.2cm}
    
    \includegraphics[width=0.95\linewidth]{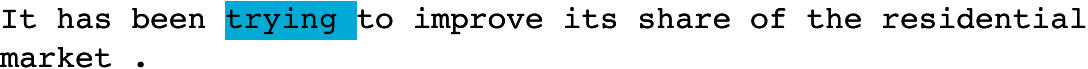}

    \vspace{0.2cm}
    
    \includegraphics[width=0.90\linewidth]{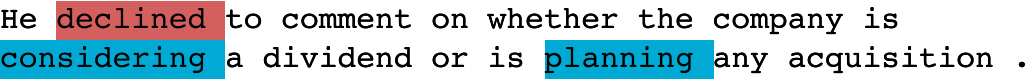}

    \vspace{0.2cm}

    \includegraphics[width=0.95\linewidth]{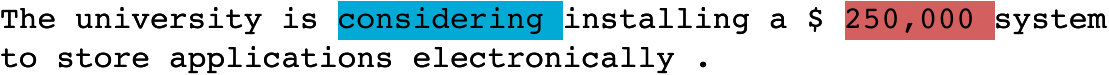}

    \caption{\label{fig:gerunds} Gerund Verb Neuron (ROBERTA) -- Layer 1: 750}
    \end{subfigure}
    \caption{\label{fig:vis} Visualizations (POS) -- Neuron 193 in Layer activates positively for superlative adjectives, Neuron 750 activates positively for gerund verbs.}
\end{figure*}

\begin{figure*}[!ht]
	\centering
    \begin{subfigure}[b]{0.65\linewidth}
    % \centering
    \includegraphics[width=0.95\linewidth]{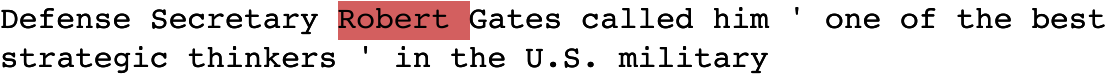}

    \vspace{0.3cm}

    \includegraphics[width=0.95\linewidth]{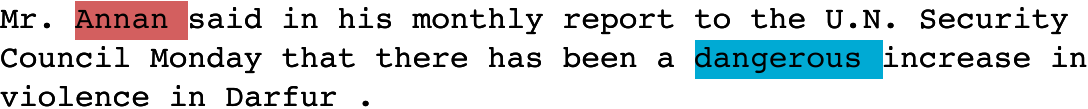}

    \vspace{0.3cm}

    \includegraphics[width=0.95\linewidth]{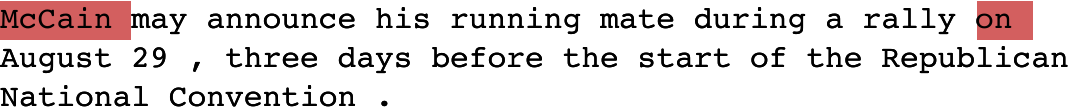}
    \caption{\label{fig:person} Person Name Neuron (XLNet) -- Layer 2: 651}
    \end{subfigure}
     \begin{subfigure}[b]{0.65\linewidth}
    % \centering
    \vspace{0.2cm}
    \includegraphics[width=0.82\linewidth]{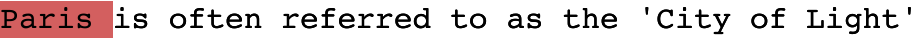}

    \vspace{0.2cm}

    \includegraphics[width=0.95\linewidth]{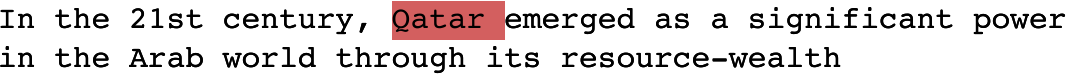}

    \vspace{0.2cm}

    \includegraphics[width=0.84\linewidth]{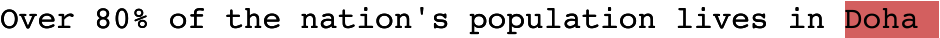}
    \caption{\label{fig:place} Place Neuron (XLNet) -- Layer 2: 115}
    \end{subfigure}
    \caption{\label{fig:vis2} Visualizations (SEM) -- Neuron 651 in layer 2 activates negatively for person names, Neuron 115 is a place neuron.}
\end{figure*}

\begin{figure*}[ht]
	\centering
    \includegraphics[width=0.85\linewidth]{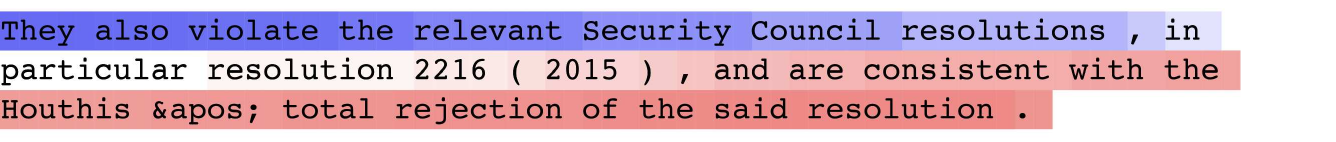}
    \caption{\label{fig:position} Position Neuron: Activates positively in the beginning, becomes neutral in the middle and negatively towards the end of sentence.}
\end{figure*}

\noindent Figures \ref{fig:vis}-\ref{fig:position} visualize the activations of the top neurons for a few properties. It shows how single neurons can focus on very specific linguistic properties. Figure \ref{fig:jjs} shows a neuron that activates positively for superlative adjectives (e.g. biggest, deadliest, heaviest). Figure \ref{fig:gerunds} shows a neuron that activates positively for gerund verbs (e.g. trying, considering etc).  Similarly Figure \ref{fig:person} shows a neuron that activates negatively for person name (e.g. Robert, Annan and Mcain etc). Figure \ref{fig:place} shows a neuron that activates negatively for places, specifically countries and cities (e.g. Paris, Qatar, Doha etc). We are able to identify neurons learning a specific property. Although our results are focused on linguistic tasks, the \texttt{LCA} method %methodology 
%is general for 
can be used to probe any property for which %supervision can be created by labeling
labeled data can be created. %the data. 
For instance, \citet{mousi2023llms} used LLMs to annotate encoded concepts within pre-trained language models and carried out neuron analysis using these annotations. Here, we trained a classifier to predict position of the word, i.e., identify if a given word is at the beginning, middle, or end of the sentence. As shown in Figure \ref{fig:position}, the top neuron identified by this classifier activates with high positive value at the beginning (blue), moves to zero in the middle (white), and gets a high negative value at the end of the sentence (red).

Another way to qualitatively validate the salient neurons identified by our method is to visualize the top words that activate a given neuron. Table \ref{tab:neuronExamples} shows salient neurons w.r.t a property in different models along with their top-5 activating words in the test-set. Looking at the top-5 activating words for a neuron, we can validate that the identified neuron indeed learns that property. These results also show how visualization can be combined with any neuron ranking method to give us a quick insight into the property a neuron has learned.

\begin{table}[ht]
  \caption{Ranked list of words for some individual neurons, VBD: Past-tense verb, VBG: Gerund Verb, VBZ: Third person singular, CD: Numbers, LOC: Location, ORG: Organization, PER: Person, YOC: Year of the century.}
\centering					
\footnotesize
%\resizebox{\columnwidth}{!}{									
    \begin{tabular}{l|l|l|l}									
    \toprule									
\textbf{Neuron}   & \textbf{Property} & \textbf{Model} & \textbf{Top-5 words} \\		
\midrule
     Layer 9: 624 & VBD & RoBERTa & supplied, deposited, supervised, paled, summoned \\
     Layer 2: 750 & VBG & RoBERTa & exciting, turning, seeing, owning, bonuses \\
     Layer 0: 249 & VBG & BERT & requiring, eliminating, creates, citing, happening \\
     Layer 1: 585 & VBZ & XLNet & achieves, drops, installments, steps, lapses, refunds \\
     Layer 2: 254 & CD & RoBERTa & 23, 28, 7.567, 56, 43 \\
     Layer 5: 618 & CD & BERT & 360, 370, 712, 14.24, 550 \\
     Layer 1: 557 & LOC & XLNet & Minneapolis, Polonnaruwa, Mwangura, Anuradhapura, Kobe \\
     Layer 5: 343 & ORG & RoBERTa & DIA, Horobets, Al-Anbar, IBRD, GSPC \\
     Layer 10: 61 & PER & RoBERTa & Grassley, Cornwall, Dalai, Bernanke, Mr.Yushchenko \\
     Layer 6: 132 & PER & BERT & Nick, Manie, Troy, Sam, Leith \\
     Layer 2: 343 & YOC & BERT &  1897, 1918, 1901, 1920,  Alam \\
    \bottomrule
    \end{tabular}
    % \caption{Ranked list of words for some individual neurons, VBD: Past-tense verb, VBG: Gerund Verb, VBZ: Third person singular, CD: Numbers, LOC: Location, ORG: Organization, PER: Person, YOC: Year of the century}

\label{tab:neuronExamples}						
\end{table}

\section{Analysis}
\label{sec:analysis}

Now that we have established the efficacy of our method, both empirically and qualitatively, let us analyze the discovered salient neurons. Specifically we want to study the following questions: i) how do the salient neurons distribute across the layers? ii) how do they distribute across the properties, iii) how redundant are the neurons w.r.t a concept?  iv) how transfer learning impacts linguistic knowledge in deep NLP models? and v) how much overlap in neurons do we see in the multilingual models?

\subsection{Distribution of Neurons Across Layers}
\label{sec:layerWise}

Work done in interpreting deep NLP models often study the distribution of linguistic knowledge across layers by training classifiers for each layer individually \citep{liu-etal-2019-linguistic,kim2020pretrained,belinkov-etal-2020-analysis}. In this study, we investigate how neurons selected from the entire network distribute across various layers of the model. This analysis provides an alternative perspective on which layers predominantly contribute to different tasks.

\begin{figure*}[t]
    \centering
    
    \begin{subfigure}[b]{0.32\linewidth}
    \centering
    \includegraphics[width=\linewidth]{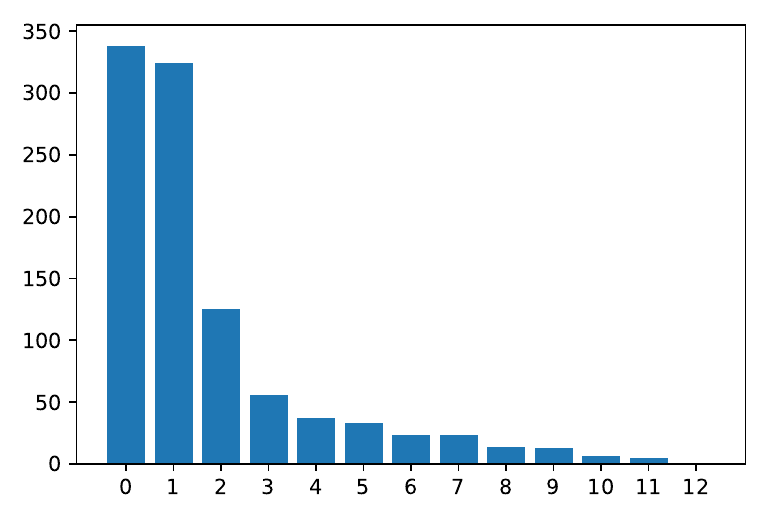}
    \caption{Suffix -- BERT}
    \label{fig:suffbert}
    \end{subfigure}
    \begin{subfigure}[b]{0.32\linewidth}
    \centering
    \includegraphics[width=\linewidth]{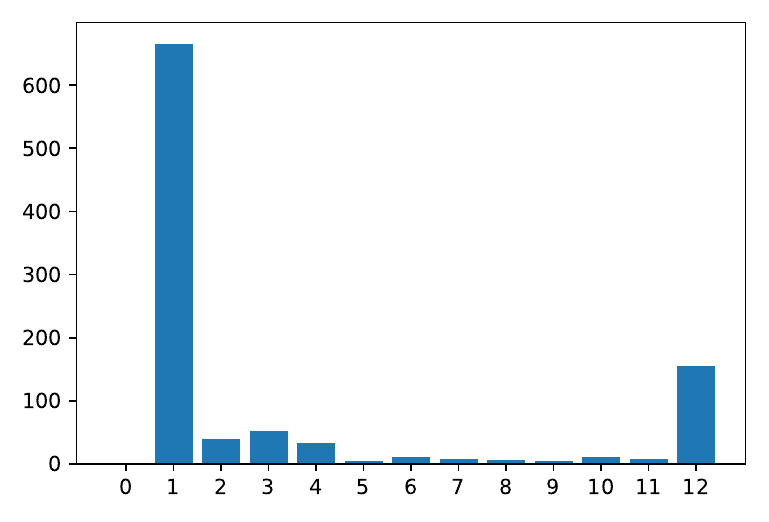}
    \caption{Suffix -- XLNet}
    \label{fig:suffxlnet}
    \end{subfigure}    
    \begin{subfigure}[b]{0.32\linewidth}
    \centering
    \includegraphics[width=\linewidth]{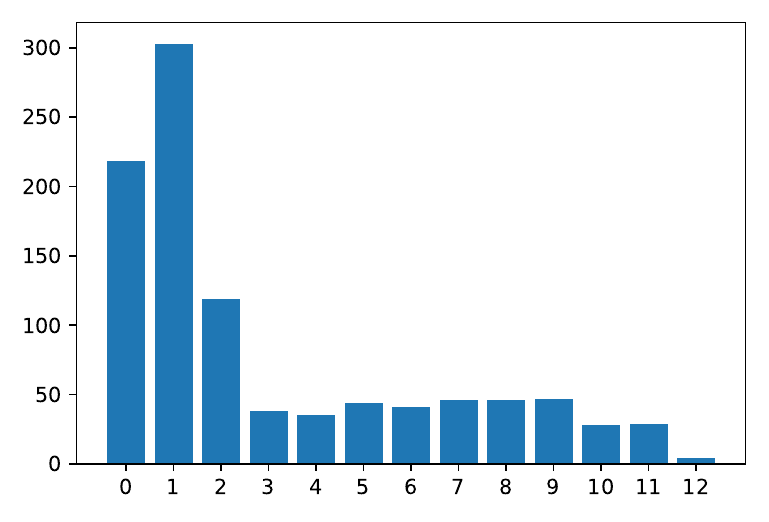}
    \caption{Suffix -- RoBERTa}
    \label{fig:suffroberta}
    \end{subfigure}
    
    \begin{subfigure}[b]{0.32\linewidth}
    \centering
    \includegraphics[width=\linewidth]{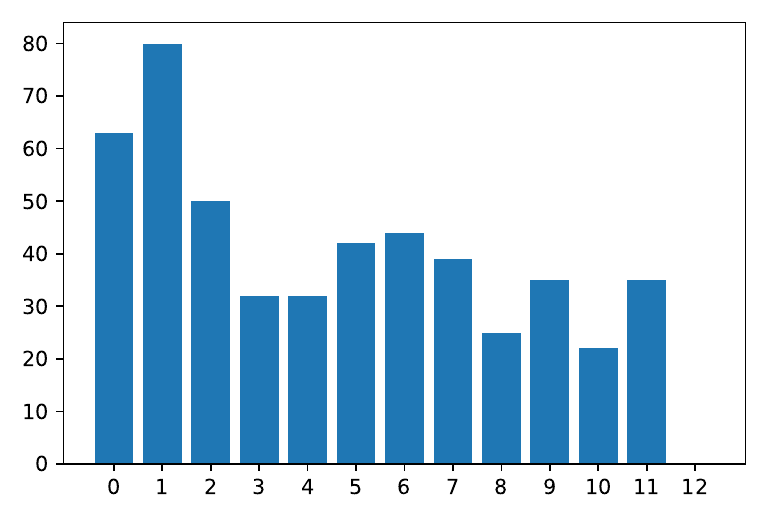}
    \caption{POS -- BERT}
    \label{fig:posbert}
    \end{subfigure}
    \begin{subfigure}[b]{0.32\linewidth}
    \centering
    \includegraphics[width=\linewidth]{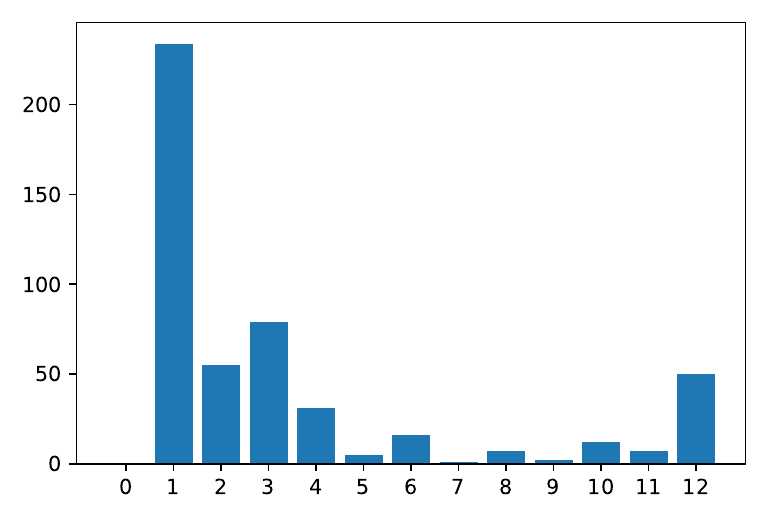}
    \caption{POS -- XLNet}
    \label{fig:posxlnet}
    \end{subfigure}    
    \begin{subfigure}[b]{0.32\linewidth}
    \centering
    \includegraphics[width=\linewidth]{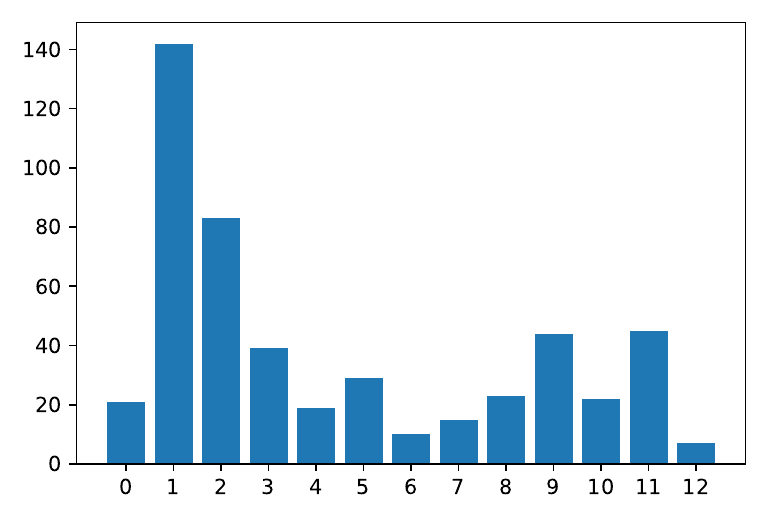}
    \caption{POS -- RoBERTa}
    \label{fig:posroberta}
    \end{subfigure}
    
    \begin{subfigure}[b]{0.32\linewidth}
    \centering
    \includegraphics[width=\linewidth]{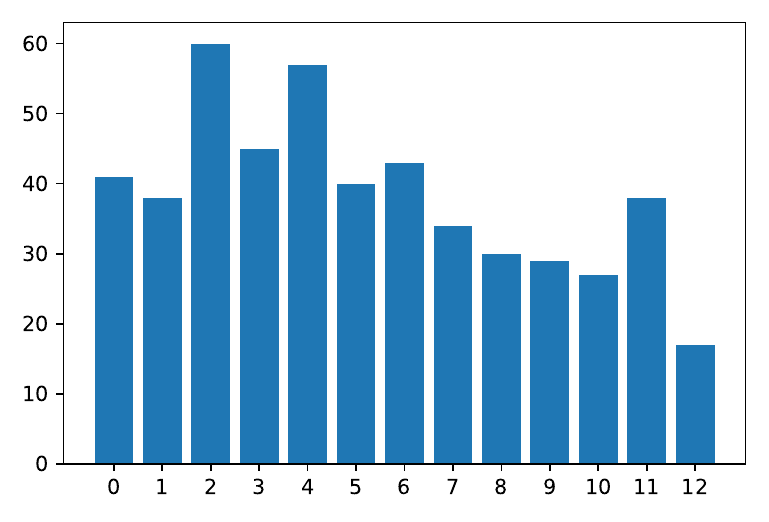}
    \caption{SEM -- BERT}
    \label{fig:sembert}
    \end{subfigure}
    \begin{subfigure}[b]{0.32\linewidth}
    \centering
    \includegraphics[width=\linewidth]{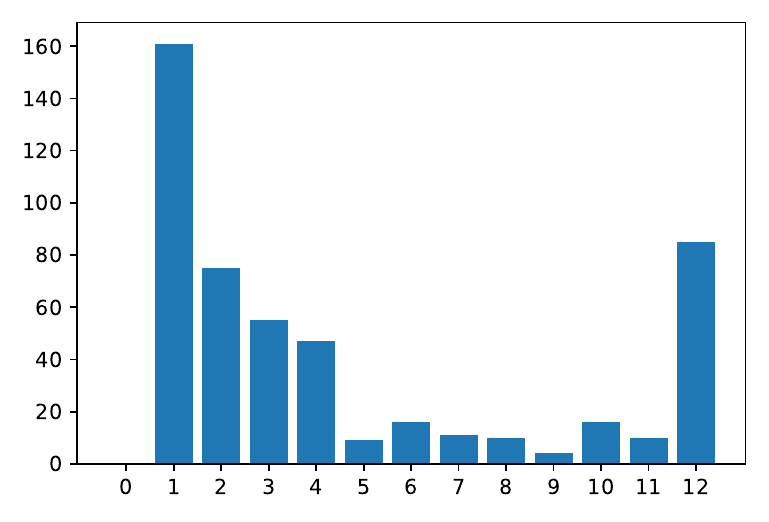}
    \caption{SEM -- XLNet}
    \label{fig:semxlnet}
    \end{subfigure}    
    \begin{subfigure}[b]{0.32\linewidth}
    \centering
    \includegraphics[width=\linewidth]{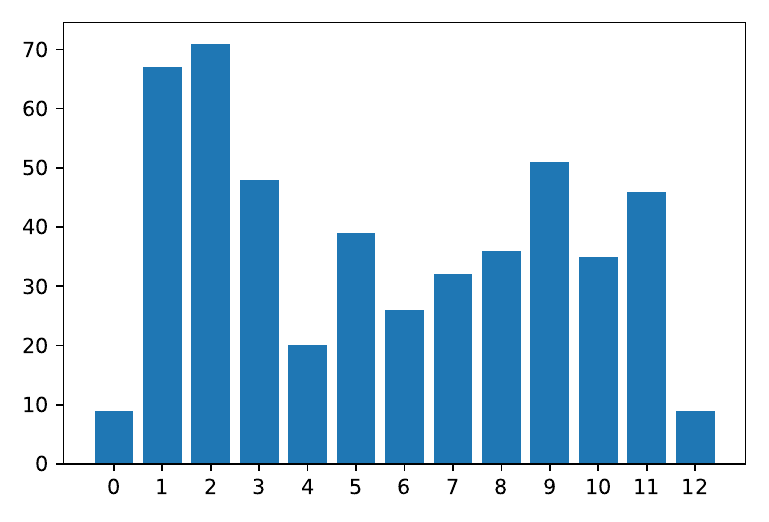}
    \caption{SEM -- RoBERTa}
    \label{fig:semroberta}
    \end{subfigure}

    % \begin{subfigure}[b]{0.30\linewidth}
    % \centering
    % \includegraphics[width=\linewidth]{figures/BERT-Chunk.png}
    % \caption{Chunking -- BERT}
    % \label{fig:chunkingbert}
    % \end{subfigure}
    % \begin{subfigure}[b]{0.30\linewidth}
    % \centering
    % \includegraphics[width=\linewidth]{figures/XLNet-Chunk.png}
    % \caption{Chunking -- XLNet}
    % \label{fig:chunkingxlnet}
    % \end{subfigure}    
    % \begin{subfigure}[b]{0.30\linewidth}
    % \centering
    % \includegraphics[width=\linewidth]{figures/RoBERTa-Chunk.png}
    % \caption{Chunking -- RoBERTa}
    % \label{fig:chunkingcalypso}
    % \end{subfigure}

    \begin{subfigure}[b]{0.32\linewidth}
    \centering
    \includegraphics[width=\linewidth]{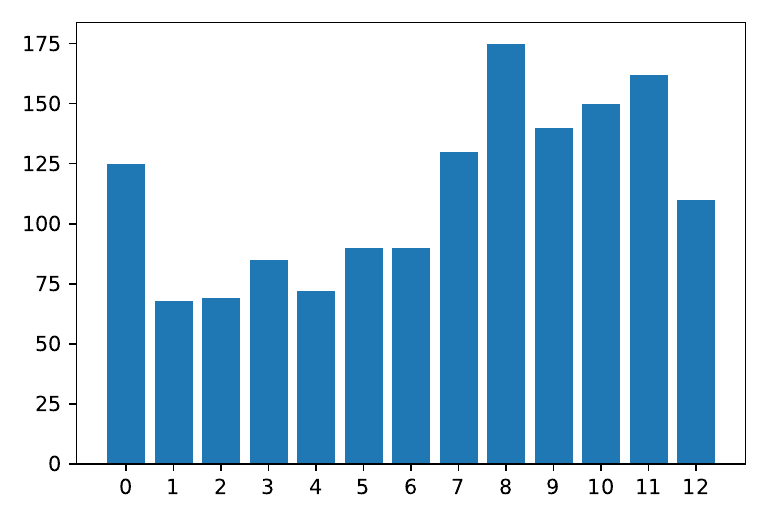}
    \caption{CCG -- BERT}
    \label{fig:ccgbert}
    \end{subfigure}
    \begin{subfigure}[b]{0.32\linewidth}
    \centering
    \includegraphics[width=\linewidth]{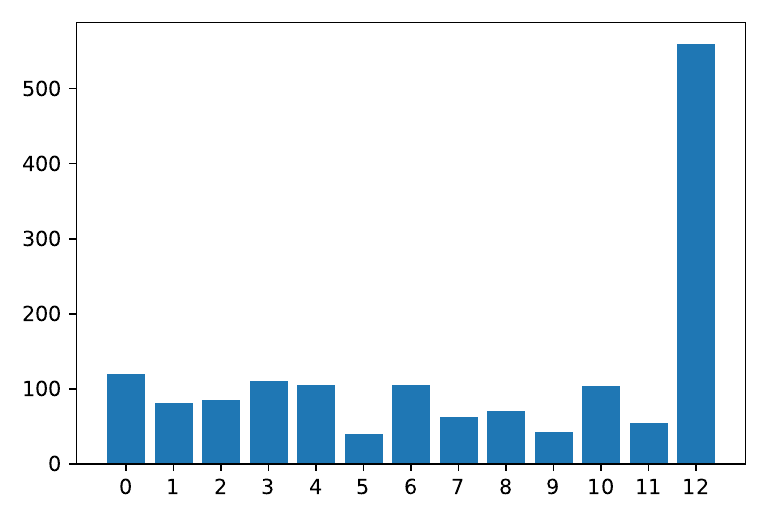}
    \caption{CCG -- XLNet}
    \label{fig:ccgxlnet}
    \end{subfigure}    
    \begin{subfigure}[b]{0.32\linewidth}
    \centering
    \includegraphics[width=\linewidth]{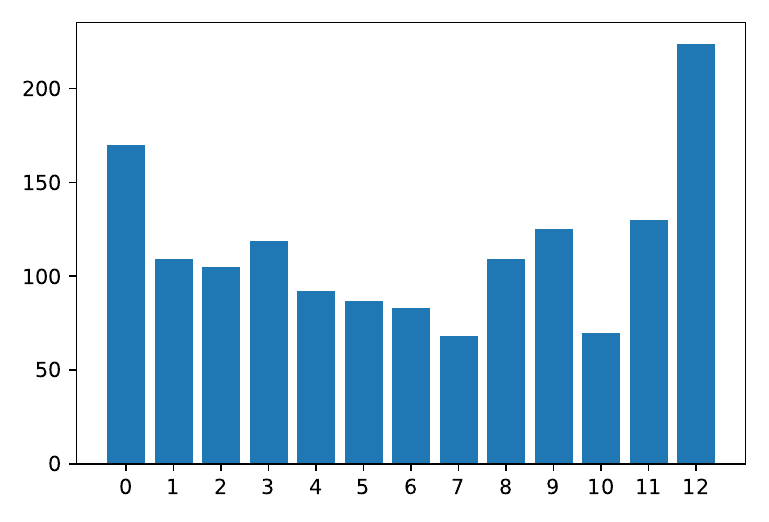}
    \caption{CCG -- RoBERTa}
    \label{fig:ccgcalypso}
    \end{subfigure}
    \caption{How top neurons distribute across different layers for each task? X-axis = Layer number, Y-axis = Number of neurons selected from that layer.}
    \label{fig:layerwise}
%\vspace{-10pt}
\end{figure*}

We explore two additional tasks in this section. A simple task of predicting the suffix of a word given its representation, 
%is a shallow task of predicting suffix of a word given representation 
and the second is a more complex task predicting CCG tag of a word.
CCG captures syntactic information globally within a sentence at the word level by assigning a label to each word, indicating its syntactic role. These annotations can be conceptualized as a function that processes and produces syntactic categories (refer to Table \ref{tab:example-annotation} for an example). 

The findings presented in Figure \ref{fig:layerwise} reveal insights across various tasks and transformer models. Neurons responsible for predicting suffixes, a shallow linguistic task, are primarily concentrated in the first two layers of the models. Similarly, abstract concepts like learning morphology (POS tagging) are predominantly captured within these initial layers as well. These patterns indicate that the initial layers of the network are focused on reconstructing word structure, which is disassembled during the subword segmentation in the pre-processing stage. As the complexity of the task increases, such as learning lexical semantics, we observe that neurons in middle and higher middle layers become more active. 
At the opposite end of the spectrum, the intricate task of modeling syntax (CCG supertagging) is primarily handled by neurons in the final layers.

\textbf{How do the architectures compare in terms of the distribution of concept neurons?} We notice that while BERT adheres to the observed patterns, the top neurons in BERT are spread out more evenly across the layers compared to XLNet, where the top neurons for a specific task are concentrated in fewer layers. This indicates that in BERT, each layer contains neurons specialized in learning specific language properties, whereas in XLNet, certain designated layers are specialized for these properties. Unlike auto-encoder-based BERT models, neurons in the embedding layer consistently show minimal contribution in XLNet across all tasks. Let's examine these distinctions for each concept:

In the case of predicting suffixes, we clearly see that the neurons capturing the information are learned in the first two layers. It is interesting to note that RoBERTa maintains this information consistently in the network. For the task of learning parts-of-speech, most layers in BERT contributed towards the top neurons, while the distribution is dominated by lower layers in XLNet, with an exception of XLNet not choosing any neurons from the embedding layer. RoBERTa exhibits similar behavior to XLNet with embedding layer contributing small number of neurons. However, in this case higher layers also contribute to the most salient neurons. On the task of semantic tagging, similar to POS, all layers of BERT contributed to the list of top neurons. However, the lower-middle layers showed the most contribution. This is in line with 
\cite{liu-etal-2019-linguistic} 
who found middle and higher middle layers to give optimal results for the semantic tagging task.  On XLNet and RoBERTa, the lower layers after the embedding layer got the largest share of the top neurons of SEM. This trend is consistent across other tasks,  i.e., the core linguistic information is learned earlier in the network with an exception of BERT, which distributes information across the network. 

% Compared to the lexical tasks discussed above, that are dominantly captured in the initial layer and middle layers, notice a shift in pattern when learning the syntactic task. The salient neurons appear in the higher layers. For XLNet and RoBERT, most number of salient neurons were contributed from the final layer. Even for BERT that displays more distributed behavior, higher layers contributed more neurons than the initial layers.  Our overall findings resonate with previous work done by \cite{liu-etal-2019-linguistic} and \cite{tenney-etal-2019-bert} who analyzed contextualized representations across layers to study how various linguistic concepts are learned at different layers. Here we used our neuron analysis method to complement their study. 

In contrast to the lexical tasks discussed earlier, which are mainly captured in the initial and middle layers, there is a noticeable shift in pattern when it comes to learning syntactic tasks. Salient neurons for syntactic tasks emerge in the higher layers. In XLNet and RoBERTa, the majority of salient neurons were contributed by the final layer. Even in the case of BERT, which exhibits a more distributed behavior, higher layers contribute more neurons than the initial layers. These findings align with previous research %conducted by 
\cite{liu-etal-2019-linguistic,tenney-etal-2019-bert} on analyzing contextualized representations across layers to explore how various linguistic concepts are learned at different layers. Our neuron analysis method serves as a complementary approach to their study, reinforcing the overall understanding of the layer-wise distribution of linguistic knowledge in transformer models.

\subsection{Distribution of Neurons Across Properties}
\label{sec:propertyWise}

\begin{figure*}[!ht]
	\centering
    \includegraphics[width=0.80\linewidth]{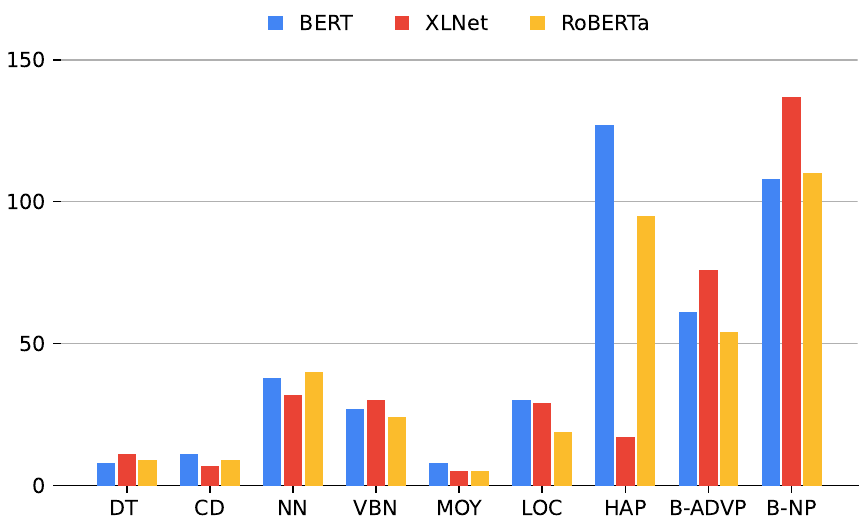}
    \caption{\label{fig:loc-dist} Localized vs. Distributed Properties: DT = Determiner,  CD = Numbers, NN = Noun, VBN = Past participle verb, MOY, Month of Year, LOC = Location, HAP = Event (happening), B-ADVP = Beginning of Adverb Phrase, B-NP = Beginning of Noun Phrase, Y-axis = number of neurons assigned to this property}
\end{figure*}

The Linguistic Correlation Analysis provides a ranking for a property set (e.g., Chunking tagging) as well as for individual labels (e.g., Chunking:B-VP). The ranking algorithm extracts neurons for each label $l$ in task $T$, sorted based on absolute weights. The final rankings are obtained by selecting neurons from each label using the neuron ranking algorithm as described in Section \ref{sec:algorithm}. The distribution of neurons, allows us to analyze how localized or distributed a property is, based on the number of neurons that are selected for each label in the task. Figure \ref{fig:loc-dist} shows various properties alongside the number of neurons (in the minimal set) required to capture each property. We observed that functional words, also known as closed class categories such as determiners or numbers, are localized in as few as 10 neurons. In contrast, open class words like nouns and verbs are distributed across a larger number of neurons. Open class categories occur in a variety of contexts, necessitating more neurons to capture diverse linguistic nuances. This observation held consistently across different properties within various tasks, as well as across different models and languages. For detailed results, please refer to Appendix \ref{sec:appendix:neuronSpread}.

\vspace{2mm}
\noindent \textbf{How do individual properties distribute across layers?} 
Previously, our analysis focused on examining each linguistic task as a whole. Now, we study the individual properties (e.g., adjectives) to understand whether they are localized or widely distributed across layers in different architectures. We observed intriguing similarities across architectures. For instance, neurons predicting the foreign words (\textbf{FW}) property were predominantly localized in the final layers (BERT: 13, XLNet: 11, RoBERTa: 11) in the architectures under investigation. In contrast, neurons capturing common class words such as adjectives (\textbf{JJ}) and locations (\textbf{LOC}) were localized in lower layers (BERT: 0, XLNet: 1, RoBERTa: 1), indicating that these common class words might be determined at a lexical level.
However, we did find variations in certain cases. For example, personal pronouns (\textbf{PRP}) or plural nouns (\textbf{NNS}) in POS tagging and event class (\textbf{EXC}) in semantic tagging were handled at different layers across different architectures.

% Previously we analyzed each linguistic task in totality. We now study whether individual properties (e.g., adjectives) are localized or well distributed across layers in different architectures. We observed interesting cross architectural similarities, for example the neurons that predict the foreign words (\textbf{FW}) property were predominantly localized in final layers (BERT: 13, XLNET: 11, RoBERTa: 11) of the network in all the understudied architectures. In comparison, the neurons that capture common class words such as adjectives (\textbf{JJ}) and locations (\textbf{LOC}) are localized in lower layers (BERT: 0, XLNET: 1, RoBERTa:1) implying that the common class words are perhaps determined on a lexical level.
% In some cases, we did find variance, for example personal pronouns (\textbf{PRP}) or plural nouns (\textbf{NNS}) in POS tagging and event class (\textbf{EXC}) in semantic tagging were handled at different layers across different architectures.

\subsection{Redundancy}
\label{sec:redundancy}

% Neural networks are designed to be distributed in nature and are therefore innately redundant. Over-parameterization in pretrained models with a combination of various training and design choices, such as use of drop-outs causes further redundancy of information. An emerging body of work has shown that these models do not require all the representational power lent by the rich architectural choices during inference \citep{sixteenHeads}. This hypothesis has led many researchers to prune components during inference without a significant loss in model performance \citep{voita-etal-2019-analyzing,poorBERT}. 

Neural networks are inherently distributed and, as a result, exhibit natural redundancy. Pretrained models, due to over-parameterization and the incorporation of techniques like drop-outs, further contribute to this redundancy. A growing body of research suggests that these models don't utilize the full representational capacity offered by their complex architectures during inference \citep{sixteenHeads}. This hypothesis has motivated researchers to prune components during inference, a process that has been shown to not significantly impact model performance \citep{voita-etal-2019-analyzing,sajjad2023:csl}.

In Section \ref{sec:minimalNeuronRetrain}, we identified the most salient neurons for various linguistic tasks. Now we will explore whether the remaining neurons in the network contain redundant information. To conduct this experiment, we randomly selected $N$ neurons from the model and retrained a classifier to predict specific properties. We set $N$ equal to the number of top neurons selected for the task. The results presented in Table \ref{tab:redundancy} demonstrate that using the same number of randomly selected neurons yields very similar performance (within a margin of less than 3\%) compared to the classifier trained with the top neurons. Surprisingly, even the least significant neurons carry sufficient information to predict the tasks accurately. This aligns with the findings of \citet{prasanna-etal-2020-bert}, who revealed the high re-trainability of even the poorest sub-networks in BERT. These results have practical implications for network pruning and efficient feature-based transfer learning \cite{dalvi-2020-CCFS}. %This approach offers a promising alternative to the traditional fine-tuning method \citep{peters-etal-2019-tune}.

% In section \ref{sec:minimalNeuronRetrain}, we identified the most salient neurons for different linguistic tasks. Here we probe, if the other neurons in the network exhibit 
% posses the same information redundantly? To carry out this experiment, we randomly sample $N$ neurons from the model and retrain classifier towards the task of predicting certain property. We choose an $N$ equal to the number of top neurons selected for the task. Table \ref{tab:redundancy} shows that using the same number of randomly selected neurons attain a very close performance ($< 3\%$) compared to the classifier retrained using the top neurons. Even bottom neurons possess enough information to predict the tasks. Our findings resonate with  \citet{prasanna-etal-2020-bert} who also showed that even the worst sub-networks in BERT to be highly re-trainable. These findings entail useful application for pruning of network and efficient feature-based transfer learning, which has been considered as a viable alternative to the fine-tuning method \citep{peters-etal-2019-tune}. 
%We carried out an elaborate study along this line in
%explored this idea in 
%\citeauthor{dalvi-2020-CCFS}, where we use redundancy analysis to reduce the number of layers required in a forward pass. Then we remove task-specific redundant neurons using Linguistic Correlation Analysis described in this paper. We show that one can reduce the feature set to less than 100 neurons for several tasks while maintaining more than 97\% of the performance on the GLUE tasks.

\begin{table}[t]									
\centering
 \caption{Selecting minimal number of neurons for each downstream NLP task. Accuracy numbers reported on blind test-set -- Neu$_t$ = Percentage of top selected neurons, Acc$_a$ = Accuracy using all neurons, Acc$_t$ = Accuracy using selected neurons after retraining the classifier using selected neurons, Acc$_r$ = Accuracy using randomly selected neurons after retraining the classifier, Acc$_b$ = Accuracy using bottom neurons after retraining the classifier.
}
\footnotesize
%\resizebox{\columnwidth}{!}{									
    \begin{tabular}{l|ccc|ccc|ccc}									
    \toprule									
    & \multicolumn{3}{c}{\textbf{POS}} & \multicolumn{3}{c}{\textbf{SEM}} & \multicolumn{3}{c}{\textbf{Chunking}} \\    		
    \midrule
    & BERT & XLNet & RoBERTa & BERT & XLNet & RoBERTa & BERT & XLNet & RoBERTa \\
    \toprule
    Neu$_t$ & 5\% & 5\% & 5\% & 5\% & 5\% & 5\% & 10\% & 10\%  & 10\% \\
    \midrule
    Acc$_a$ & 96.16 & 96.19 & 96.72 & 92.11 & 92.68 & 93.13 & 95.09 & 94.15 & 94.68 \\
    Acc$_t$ & 95.92 & 96.49 & 96.48 & 92.18 & 92.72 & 92.96 & 94.99 & 94.62 & 94.80 \\
    Acc$_r$ & 94.12 & 95.19 & 94.38 & 90.01 & 90.37 & 91.17 & 92.57 & 91.63 & 93.10 \\
    Acc$_b$ & 93.20 & 91.20 & 93.21 & 90.00 & 88.80 & 90.70 & 91.60 & 92.35 & 92.11 \\
    \bottomrule
    \end{tabular}
 %   }
%  \caption{Selecting minimal number of neurons for each downstream NLP task. Accuracy numbers reported on blind test-set -- Neu$_t$ = Percentage of top selected neurons, Acc$_a$ = Accuracy using all neurons, Acc$_t$ = Accuracy using selected neurons after retraining the classifier using selected neurons, Acc$_r$ = Accuracy using randomly selected neurons after retraining the classifier, Acc$_b$ = Accuracy using bottom neurons after retraining the classifier.
% }							
\label{tab:redundancy}					    %\vspace{-4mm}
\end{table}

\subsection{Transfer Learning}
\label{sec:fineTuning}

Continuing our neuron analysis of linguistic knowledge learned within transformer-based language models, we dive into further experiments employing our \texttt{LCA} method to investigate the redistribution of morpho-syntactic and semantic knowledge within these pretrained models as they undergo fine-tuning for GLUE tasks \citep{wang-etal-2018-glue}. Specifically, we focus on extracting the most salient neurons related to specific linguistic properties (e.g., POS task), from both the pre-trained model and its fine-tuned counterpart. By comparing the distribution of these neurons across the network, we gain valuable insights into how linguistic knowledge evolves during the fine-tuning process. Below we discuss notable findings from this set of experiments:

\begin{table}[t]									
\centering
 \caption{Minimal top and bottom number of neurons for each downstream NLP task before (Baseline) and after fine-tuning. Task$_t$ = Accuracy using top selected neurons after retraining the classifier using selected neurons, Task$_b$ = Accuracy using bottom neurons after retraining the classifier using selected neurons. }
\footnotesize
%\resizebox{\columnwidth}{!}{									
    \begin{tabular}{l|ccc|ccc|ccc}									
    \toprule									
    & \multicolumn{3}{c}{\textbf{POS}} & \multicolumn{3}{c}{\textbf{SEM}} & \multicolumn{3}{c}{\textbf{Chunking}} \\   		
    \midrule
    & BERT & XLNet & RoBERTa & BERT & XLNet & RoBERTa & BERT & XLNet & RoBERTa \\
    \toprule
    Baseline$_t$ & 95.92 & 96.49 & 96.48 & 92.18 & 92.72 & 92.96 & 94.99 & 94.62 & 94.80 \\
    Baseline$_b$ & 93.20 & 91.20 & 93.21 & 90.00 & 88.80 & 90.70 & 91.60 & 92.35 & 92.11 \\
    \midrule
    MRPC$_t$ & 95.91 & 95.24 & 95.23 & 92.2 & 92.33 & 91.53 & 94.97 & 93.8 & 94.5 \\
    MRPC$_b$ & 93.13 & 78.83 & 91.11 & 90.04 & 72.33 & 88.14 & 91.92 & 88.0 & 91.9 \\
    \midrule
    QNLI$_t$ & 96.09 & 94.71 & 95.81 & 92.09 & 91.41 & 91.54 & 94.33 & 93.99 &  94.08 \\
    QNLI$_b$ & 93.47 & 10.39 & 84.31 & 90.01 & 17.84 & 78.67 & 91.78 & 79.37 & 90.71 \\
        \midrule
    MNLI$_t$ & 95.81 & 94.93 & 95.45 & 91.9 & 91.34 & 91.43 & 93.81 & 92.86  & 93.24 \\
    MNLI$_b$ & 93.03 & 41.13  & 84.85 & 89.97 & 43.05 & 79.02 & 91.33 & 85.02 & 89.53 \\
        \midrule
    RTE$_t$ & 95.93 & 95.24 & 95.63 & 92.0 & 91.36 & 91.56 & 94.89 & 94.01 & 94.33 \\
    RTE$_b$ & 93.11 & 87.96 & 90.42 & 90.03 & 80.01 & 86.94 & 92.01 & 89.98 & 90.91 \\
        \midrule
    SST$_t$ & 95.92 & 95.02 & 95.67 & 92.1 & 91.33 & 91.51 & 94.48 & 93.74 & 94.17 \\
    SST$_b$ & 93.38 & 35.22 & 60.43 & 89.93 & 34.83 & 60.73 & 91.91 & 80.34 & 82.86 \\
        \midrule
    STS$_t$ & 95.94 & 95.15 & 95.76 & 92.1 & 91.68 & 91.52 & 94.89 & 93.82 & 94.31 \\
    STS$_b$ & 93.12 & 88.14 & 85.91 & 90.03  & 83.78 & 79.56 & 92.01 & 89.93 & 88.89 \\
    \bottomrule
    \end{tabular}
 %   }
 % \caption{Minimal top and bottom number of neurons for each downstream NLP task before (Baseline) and after fine-tuning. Task$_t$ = Accuracy using top selected neurons after retraining the classifier using selected neurons, Task$_b$ = Accuracy using bottom neurons after retraining the classifier using selected neurons }	
\label{tab:fineTuning}								
\end{table}

\vspace{2mm}

\noindent \textbf{Information becomes less distributed in the fine-tuned XLNet and RoBERTa models post fine-tuning.} 
Table~\ref{tab:fineTuning} presents the classifier accuracy results obtained by selecting the most (top) and least (bottom) N\% salient neurons across various tasks and models. Comparing the performance of these neurons in fine-tuned models to the baseline model, a significant decrease in accuracy is evident, particularly in the cases of RoBERTa and XLNet. This decline implies that the information stored in the baseline models is more redundant, given that the bottom neurons still retain linguistic knowledge, as indicated in Section \ref{sec:redundancy} on redundancy. In contrast, the fine-tuned models exhibit more localized and less distributed information, a consequence of their specific adaptation for downstream NLP tasks. Noteworthy is the finding that the bottom neurons in the fine-tuned BERT model experienced minimal change, suggesting that linguistic information remains redundant and distributed in BERT even after the fine-tuning process.
% Table~\ref{tab:fineTuning} displays the accuracy of the classifier when selecting the most (top) and least (bottom) N\% salient neurons across different tasks and models. We observed a significant drop in performance for the bottom neurons in the fine-tuned models compared to the baseline model, specifically in the case of RoBERTa and XLNet. These results indicate that the information in the baseline models is more redundant, as the bottom neurons still preserve linguistic knowledge (as seen in the results from Section \ref{sec:redundancy} on redundancy). On the contrary, the fine-tuned models exhibit more localized and less distributed information, as they are specifically fine-tuned for a downstream NLP task. Notably, the bottom neurons in the fine-tuned BERT changed the least, suggesting that linguistic information remains redundant and distributed in BERT even after fine-tuning.

\vspace{2mm}

\begin{figure*}[ht]
    \centering
    \begin{subfigure}[b]{0.32\linewidth}
    \centering
    \includegraphics[width=\linewidth]{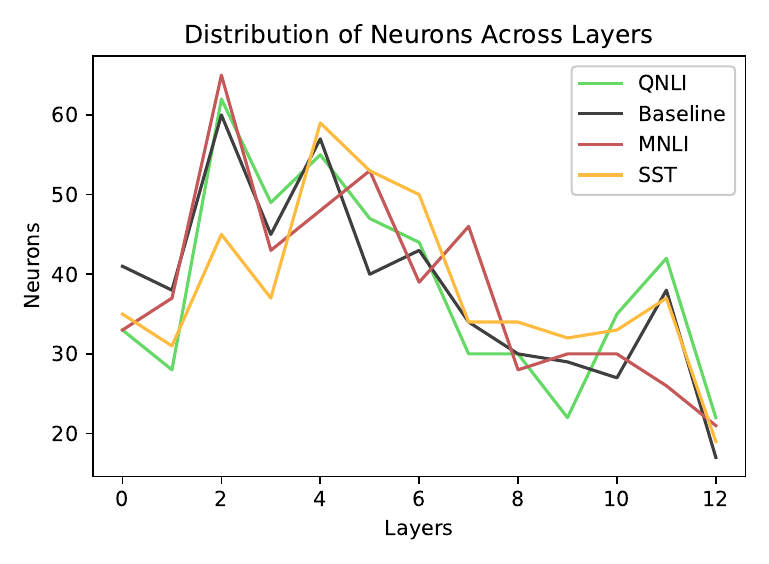}
    \caption{BERT -- SEM}
    \label{fig:bert_sem_selected}
    \end{subfigure}
    \begin{subfigure}[b]{0.32\linewidth}
    \centering
    \includegraphics[width=\linewidth]{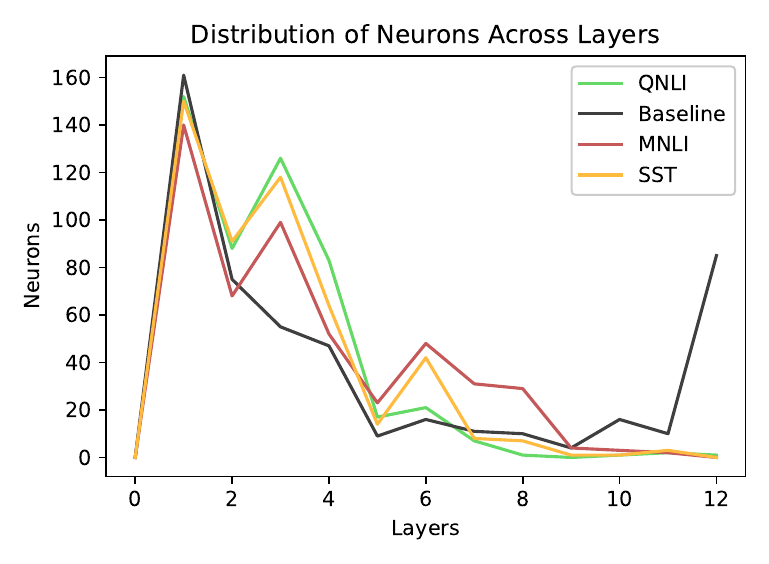}
    \caption{XLNet -- SEM}
    \label{fig:xlnet_sem_selected}
    \end{subfigure}
    \begin{subfigure}[b]{0.32\linewidth}
    \centering
    \includegraphics[width=\linewidth]{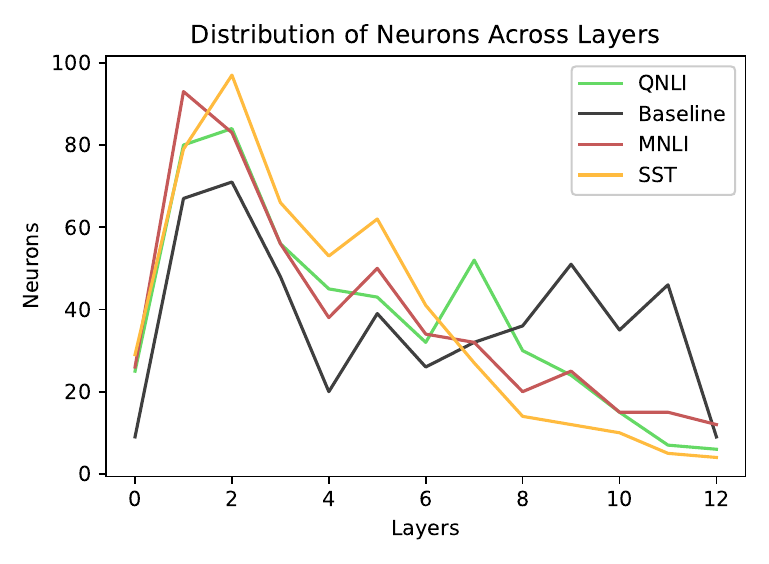}
    \caption{RoBERTa -- SEM}
    \label{fig:roberta_sem_selected}
    \end{subfigure}
    \centering
    \begin{subfigure}[b]{0.32\linewidth}
    \centering
    \includegraphics[width=\linewidth]{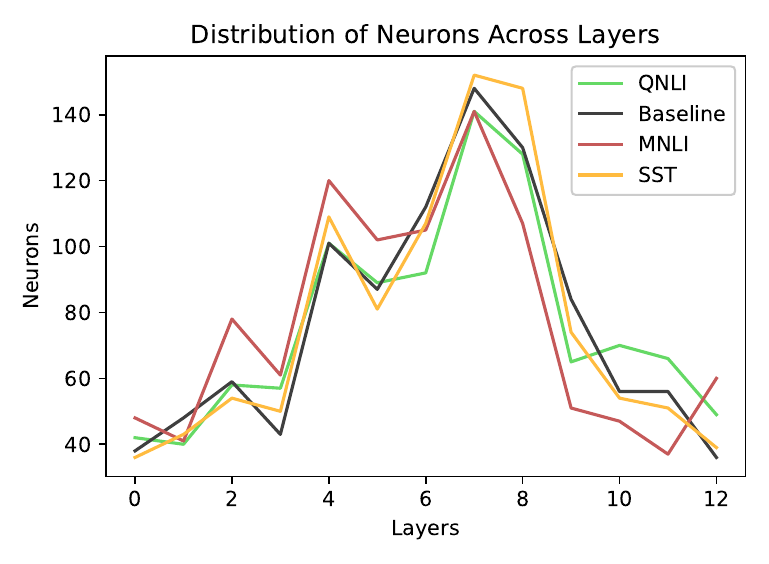}
    \caption{BERT -- Chunking}
    \label{fig:bert_chunking_neurons}
    \end{subfigure}
    \begin{subfigure}[b]{0.32\linewidth}
    \centering
    \includegraphics[width=\linewidth]{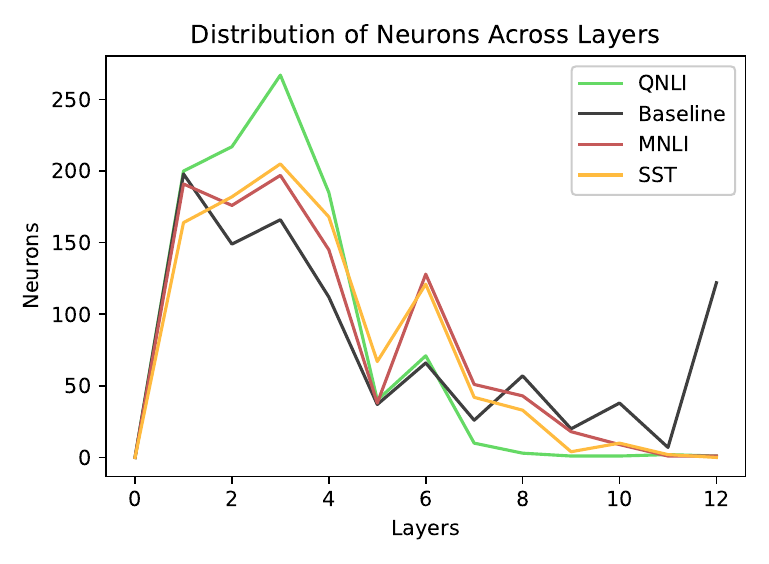}
    \caption{XLNet -- Chunking}
    \label{fig:xlnet_chunking_neurons}
    \end{subfigure}
    \begin{subfigure}[b]{0.32\linewidth}
    \centering
    \includegraphics[width=\linewidth]{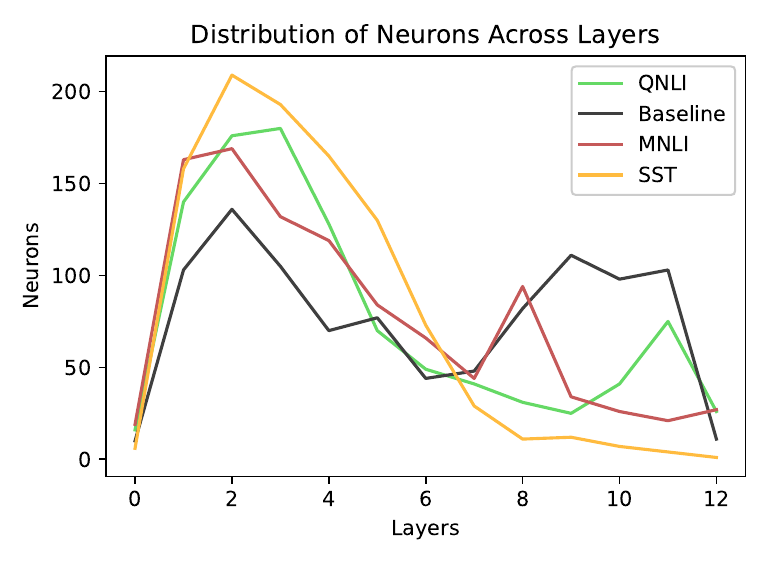}
    \caption{RoBERTa -- Chunking}
    \label{fig:roberta_chunking_neurons}
    \end{subfigure}
    \caption{Distribution of top neurons across layers. Salient linguistic neurons regress towards lower layers in XLNet and RoBERTa models as the models are fine-tuned towards GLUE tasks.}
    \label{fig:layerwise_neurons_selectedtasks}
%\vspace{-10pt}
\end{figure*}

\noindent \textbf{How do salient neurons distribute across the network layers?} In the earlier section (Section \ref{sec:layerWise}), we explored the distribution of salient neurons across network layers for each task. Now, our focus shifts to examining how this distribution changes in the fine-tuned models. The results for selected GLUE tasks are depicted in Figure \ref{fig:layerwise_neurons_selectedtasks}.\footnote{For a comprehensive list of tasks and linguistic properties, please refer to the Figure \ref{fig:layerwise_neurons_appendix} in the Appendix.} The graph illustrates the number of neurons across layers in both the baseline model and each fine-tuned model for different downstream NLP tasks. Notably, in RoBERTa and XLNet, the most salient neurons %shift from higher layers to 
are now found in the lower layers. This is particularly striking in \emph{Roberta-SST} and \emph{XLNet-QNLI} (refer to Figures \ref{fig:roberta_chunking_neurons} and \ref{fig:xlnet_chunking_neurons}), where the number of salient chunking neurons significantly increases in the lower layers while decreasing in the higher layers compared to the baseline. These observations signify a trend where neurons in lower layers bear more responsibility, focusing on learning task-specific information, while the higher layers become specialized in this regard.\footnote{\citet{neuralCollapse} recently demonstrated how neural collapse pushes structured features to the lower layers of the network. \citet{durrani-etal-2022-latent} made similar observations when studying how latent spaces evolve as the models are fine-tuned towards downstream NLP tasks.} In contrast, BERT does not exhibit this behavior. Unlike XLNet, where linguistic properties localize in the lower layers, BERT distributes information evenly across the network, as previously noted in Section \ref{sec:layerWise}, where we found linguistic properties in XLNet to be localized in the lower layers, while BERT disperses information throughout all layers. Our neuron analysis proves valuable in understanding the essential linguistic concepts for downstream tasks. A more detailed exploration would involve pinpointing the specific concepts regressed to lower layers versus those retained in higher layers. However, this in-depth analysis remains a topic for future research.

\subsection{Multilingual Models}
\label{sec:multilingual}

%Languages share orthography, morphological, syntactic and semantic patterns along with many other linguistic phenomenon. 

Recently, there has been growing interest in cross-lingual transfer learning, particularly in scenarios transitioning from high-resource to zero-shot or low-resource contexts. In this study, we investigate the effectiveness of multilingual BERT \citep{devlin-etal-2019-bert} and XLM-RoBERTa \citep{conneau2020unsupervised} using our Linguistic Correlation Analysis. We revisit our analysis focusing on parts-of-speech tagging, a task for which annotations are readily available in multiple languages. Additionally we use syntactic dependency labeling, the task of assigning label to an arc that connects two words in a syntactic dependency tree. A dependency is a directed bi-lexical relation between a head and its dependent, or modifier. %Following \citeauthor{belinkov-etal-2020-linguistic}, 
We simply concatenate the activations of the words in question and train classifiers on the concatenated vectors.\footnote{We also explored variations i) using just the activation of the head or modifier word, ii) use average vector of the two. Both variations resulted in sub-optimal classifier accuracy compared to using concatenation.}  
%Dependency structures are attractive to study for three main reasons. First, dependency formalisms have become increasingly popular in NLP in recent years, and much work has been devoted to developing large annotated datasets for these formalisms. The Universal Dependencies dataset %\cite{UD-11234/1-1983} 
%that is used in this paper has been especially influential.
We experiment with English, German and French. We used monolingual models including FlauBERT \citep{le-etal-2020-flaubert-unsupervised} and BERT-base-German-cased \citep{chan-etal-2020-germans} for French and German respectively. Below we discuss notable findings
from this set of experiments:

\begin{table}[t]									
\centering
 \caption{Selecting salient neurons for POS and Syntactic Dependency Labeling tasks. Accuracy numbers reported on blind test-set (averaged over three runs) -- Acc$_a$ = Accuracy using all neurons, Acc$_{t,r,b}$ = Accuracy using top, bottom, and randomly selected neurons after retraining the classifier using selected neurons.
}
\footnotesize
%\resizebox{\columnwidth}{!}{									
    \begin{tabular}{l|ccc|ccc|ccc}									
    \toprule									
    & \multicolumn{3}{c}{\textbf{POS English}} & \multicolumn{3}{c}{\textbf{POS German}} & \multicolumn{3}{c}{\textbf{POS French}} \\    		
    \midrule
    & BERT & mBERT & XLM-R & BERT & mBERT & XLM-R & BERT & mBERT & XLM-R \\
    \toprule
   Acc$_a$ & 94.82 & 94.48 & 94.70 & 93.37 & 93.50 &  93.91 & 96.28 & 95.84 & 96.22 \\
   Acc$_t$ & 94.52 & 94.60 & 94.17 & 93.82 & 93.42 & 93.77 & 96.24 & 95.59 & 95.39 \\
   Acc$_r$ & 93.96 & 94.55 & 94.00 & 93.40 & 93.30 & 93.63 & 95.50 & 95.31 & 95.43 \\
   Acc$_b$ & 93.52 & 94.48 & 92.20 & 93.32 & 93.01 & 92.04 & 95.04 & 95.04 & 94.27 \\
   \midrule
   & \multicolumn{3}{c}{\textbf{Dependency English}} & \multicolumn{3}{c}{\textbf{Dependency German}} & \multicolumn{3}{c}{\textbf{Dependency French}} \\    		
    \midrule
    & BERT & mBERT & XLM-R & BERT & mBERT & XLM-R & BERT & mBERT & XLM-R \\
    \toprule
   Acc$_a$ & 91.90 & 91.50 & 92.14 & 91.56 & 91.45 &  92.00 & 93.52 & 94.30 & 94.62 \\
   Acc$_t$ & 91.18 & 91.52 & 92.05 & 91.62 & 91.35 & 92.06 & 92.86 & 94.33 & 94.53 \\
   Acc$_r$ & 90.90 & 91.10 & 91.46 & 90.77 & 90.87 & 91.73 & 92.62 & 93.81 & 94.06 \\
   Acc$_b$ & 88.01 & 88.00 & 89.94 & 87.67 & 87.18 & 90.18 & 91.09 & 91.81 & 92.55 \\

    \bottomrule
    \end{tabular}
 %   }
%  \caption{Selecting salient neurons for POS and Syntactic Dependency Labeling tasks. Accuracy numbers reported on blind test-set (averaged over three runs) -- Acc$_a$ = Accuracy using all neurons, Acc$_{t,r,b}$ = Accuracy using top, bottom, and randomly selected neurons after retraining the classifier using selected neurons
% }							
\label{tab:accuracyMultilingual}					    %\vspace{-4mm}
\end{table}

\vspace{4mm}
\noindent \textbf{Salient Neurons:}  We found that a subset of neurons (5\% for POS tagging as before)  can be extracted to obtain comparable accuracy ($Acc_t$) to the oracle ($Acc_a$) which uses the entire network (See Table \ref{tab:accuracyMultilingual}). For a more complex task of predicting dependency relation given head and modifier word, a subset of 15\% neurons was found to be sufficient. 

\vspace{2mm}
\noindent \textbf{Redundancy:} Despite the fact that multilingual models are designed to learn multiple languages simultaneously, and the encoding space is shared, these models exhibited significant redundancy, similar to their monolingual counterparts. It is notable how random and bottom N\% neurons retain retrainable information crucial for restoring the majority of the performance. Although this redundancy enhances the model's robustness, it also complicates the task of isolating and controlling specific concepts within the network. %For example \citeauthor{bau2018identifying} were able to manipulate very few translations correctly by identifying salient neurons related to verbs, gender and tense, in a neural MT system trained from English into various languages. 

\begin{figure*}[t]
   \begin{subfigure}[b]{0.30\linewidth}
    \centering
    \includegraphics[width=\linewidth]{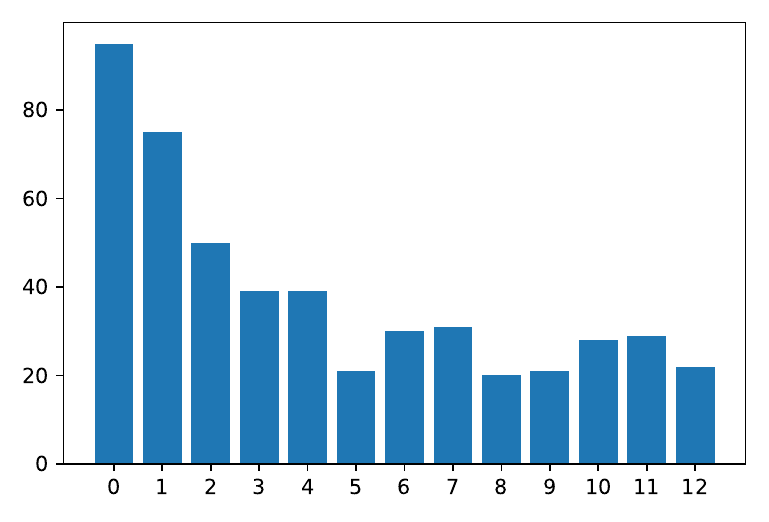}
    \caption{BERT}
    \label{fig:e-bert}
    \end{subfigure}
    \begin{subfigure}[b]{0.30\linewidth}
    \centering
    \includegraphics[width=\linewidth]{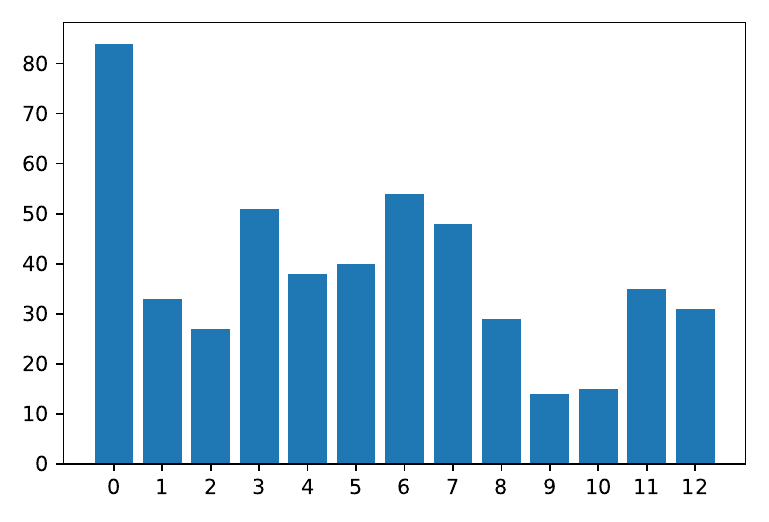}
    \caption{Multilingual BERT}
    \label{fig:e-mbert}
    \end{subfigure}    
    \begin{subfigure}[b]{0.30\linewidth}
    \centering
    \includegraphics[width=\linewidth]{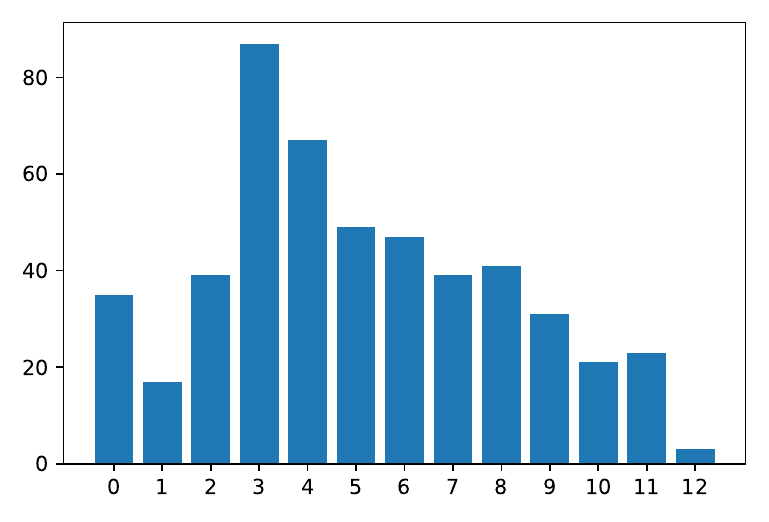}
    \caption{XLM RoBERTa}
    \label{fig:e-xlm-r}
    \end{subfigure}

    \begin{subfigure}[b]{0.30\linewidth}
    \centering
    \includegraphics[width=\linewidth]{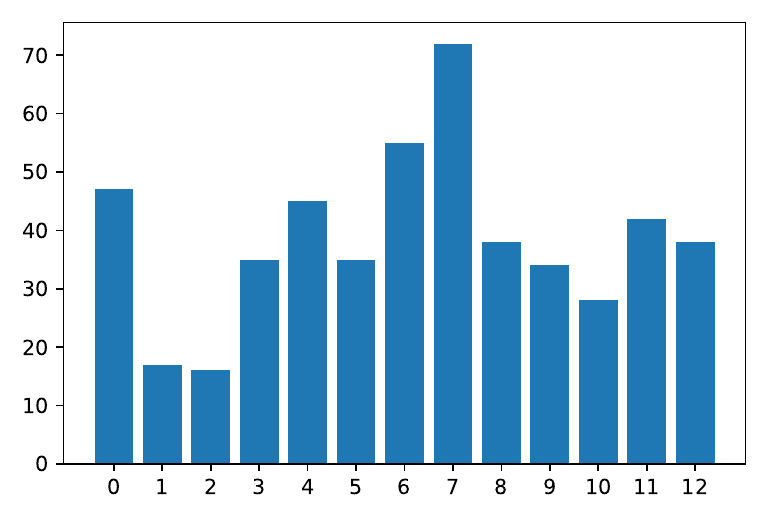}
    \caption{German  BERT}
    \label{fig:g-bert}
    \end{subfigure}
    \begin{subfigure}[b]{0.30\linewidth}
    \centering
    \includegraphics[width=\linewidth]{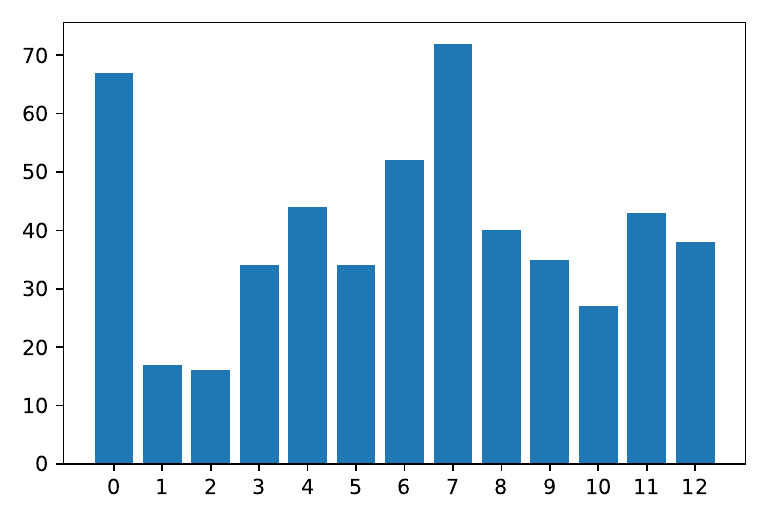}
    \caption{Multilingual BERT}
    \label{fig:g-mbert}
    \end{subfigure}    
    \begin{subfigure}[b]{0.30\linewidth}
    \centering
    \includegraphics[width=\linewidth]{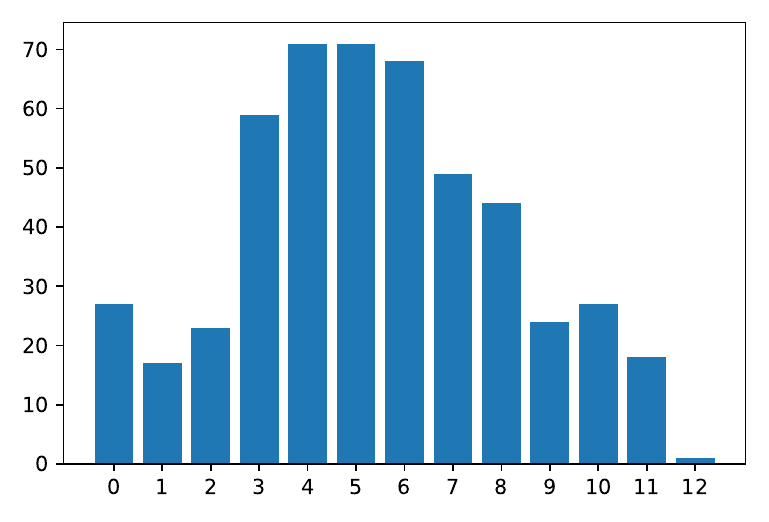}
    \caption{XLM RoBERTa}
    \label{fig:g-xlm-r}
    \end{subfigure}
    
    \begin{subfigure}[b]{0.30\linewidth}
    \centering
    \includegraphics[width=\linewidth]{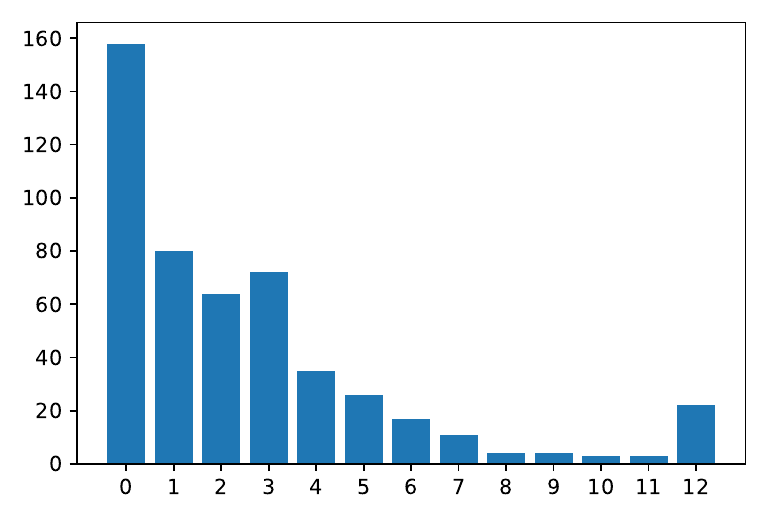}
    \caption{FlauBERT}
    \label{fig:f-bert}
    \end{subfigure}
    \begin{subfigure}[b]{0.30\linewidth}
    \centering
    \includegraphics[width=\linewidth]{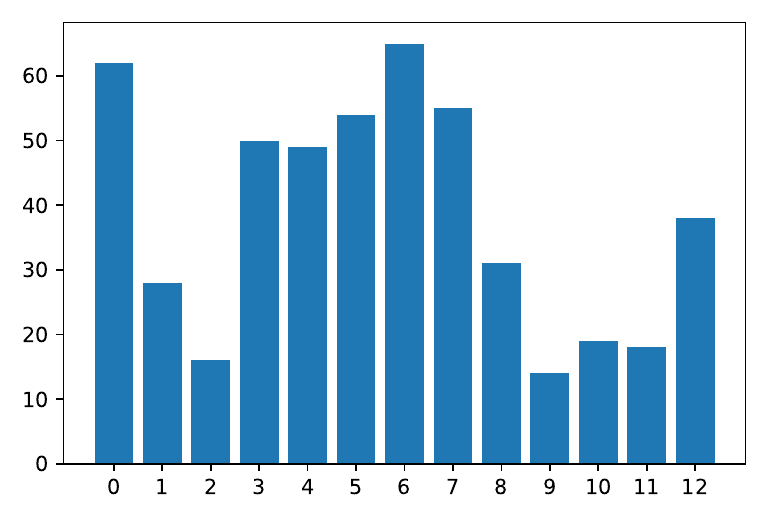}
    \caption{Multilingual BERT}
    \label{fig:f-mbert}
    \end{subfigure}    
    \begin{subfigure}[b]{0.30\linewidth}
    \centering
    \includegraphics[width=\linewidth]{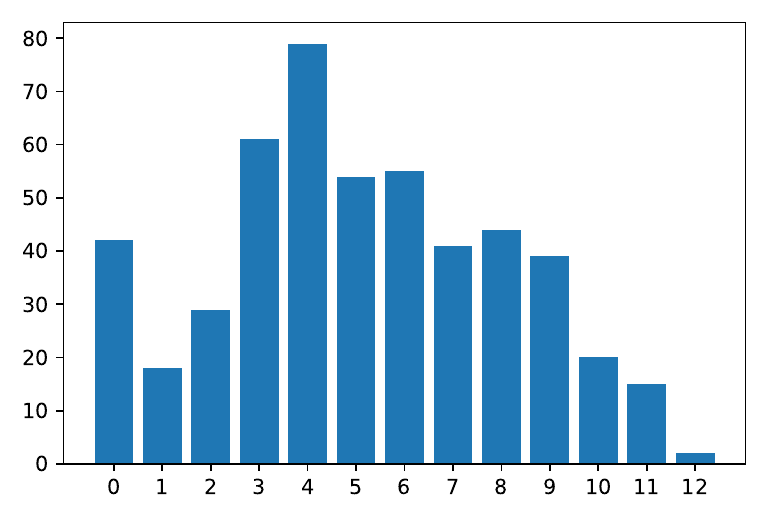}
    \caption{XLM RoBERTa}
    \label{fig:f-xlm-r}
    \end{subfigure}
    \caption{How salient neurons in POS task distribute across different layers for different models? English (a,b,c), German (d,e,f), French (g,h,i), X-axis = Layer number, Y-axis = Number of neurons selected from that layer.}
    \label{fig:layerwise-multilingual}
%\vspace{-10pt}
\end{figure*}

\begin{figure*}[t]
   \begin{subfigure}[b]{0.30\linewidth}
    \centering
    \includegraphics[width=\linewidth]{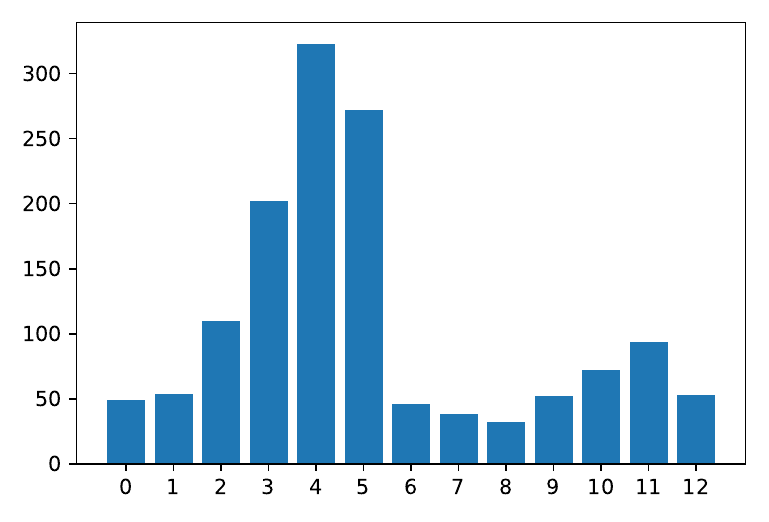}
    \caption{BERT}
    \label{fig:e-bert-syn}
    \end{subfigure}
    \begin{subfigure}[b]{0.30\linewidth}
    \centering
    \includegraphics[width=\linewidth]{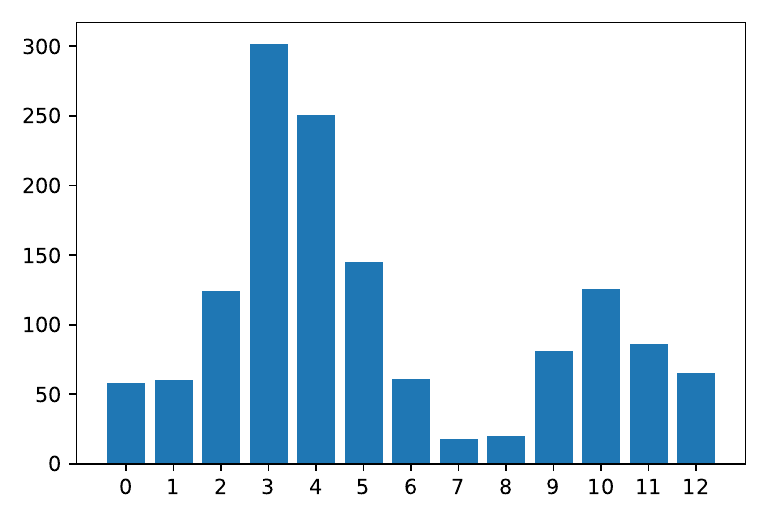}
    \caption{Multilingual BERT}
    \label{fig:e-mbert-syn}
    \end{subfigure}    
    \begin{subfigure}[b]{0.30\linewidth}
    \centering
    \includegraphics[width=\linewidth]{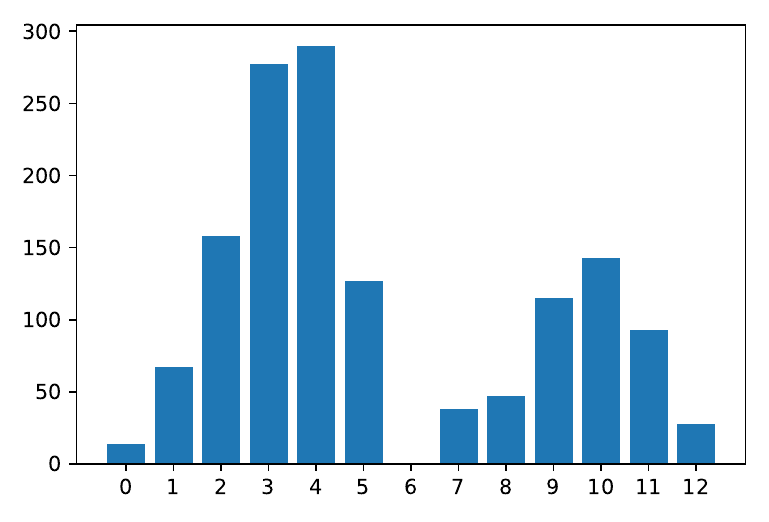}
    \caption{XLM RoBERTa}
    \label{fig:e-xlm-r-syn}
    \end{subfigure}

    \begin{subfigure}[b]{0.30\linewidth}
    \centering
    \includegraphics[width=\linewidth]{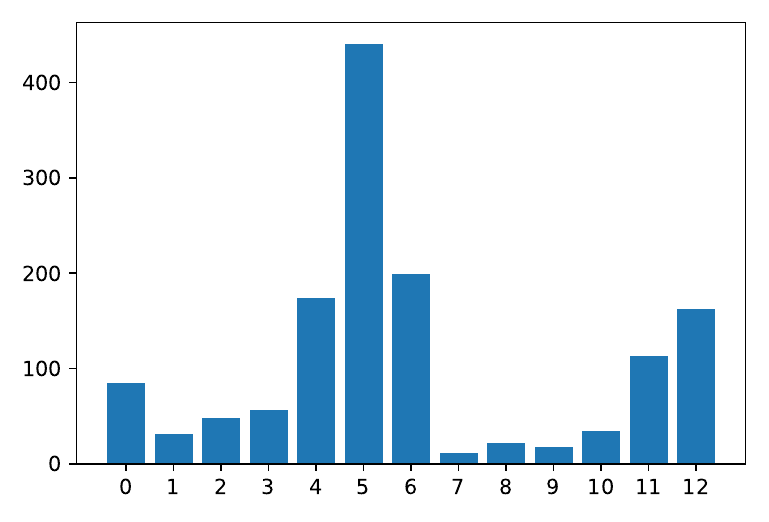}
    \caption{German  BERT}
    \label{fig:g-bert-syn}
    \end{subfigure}
    \begin{subfigure}[b]{0.30\linewidth}
    \centering
    \includegraphics[width=\linewidth]{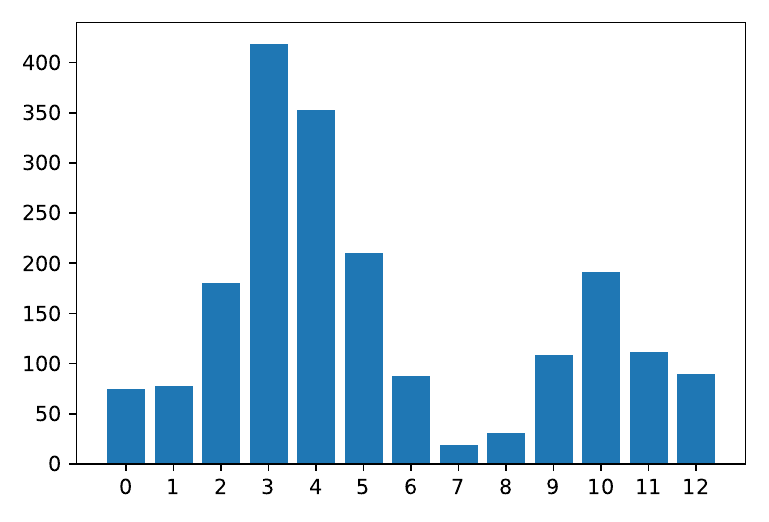}
    \caption{Multilingual BERT}
    \label{fig:g-mbert-syn}
    \end{subfigure}    
    \begin{subfigure}[b]{0.30\linewidth}
    \centering
    \includegraphics[width=\linewidth]{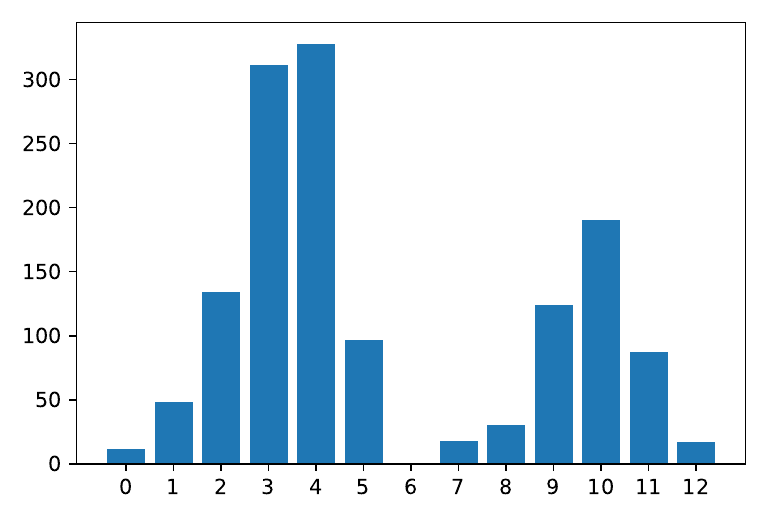}
    \caption{XLM RoBERTa}
    \label{fig:g-xlm-r-syn}
    \end{subfigure}
    
    \begin{subfigure}[b]{0.30\linewidth}
    \centering
    \includegraphics[width=\linewidth]{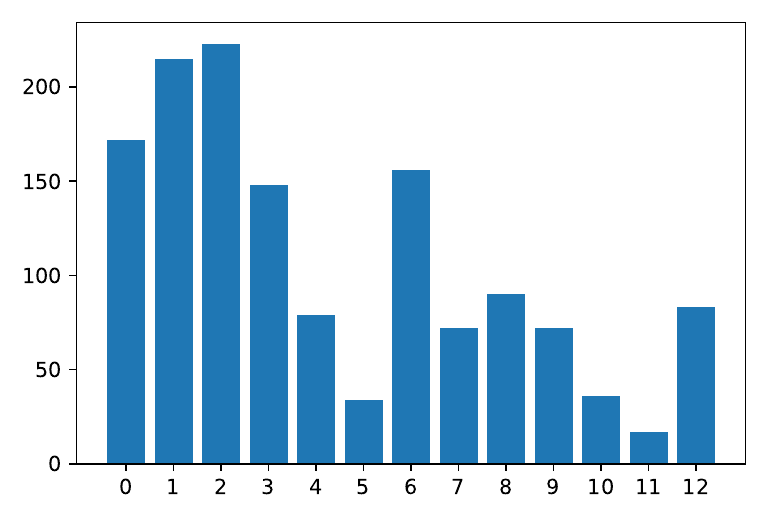}
    \caption{FlauBERT}
    \label{fig:f-bert-syn}
    \end{subfigure}
    \begin{subfigure}[b]{0.30\linewidth}
    \centering
    \includegraphics[width=\linewidth]{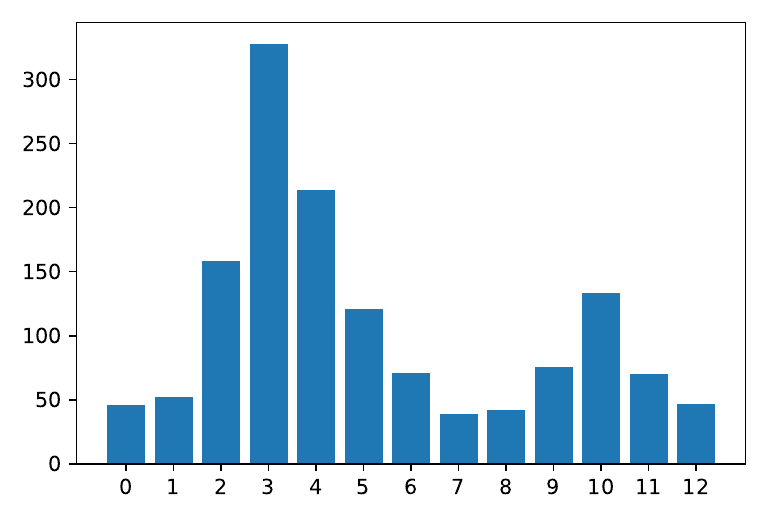}
    \caption{Multilingual BERT}
    \label{fig:f-mbert-syn}
    \end{subfigure}    
    \begin{subfigure}[b]{0.30\linewidth}
    \centering
    \includegraphics[width=\linewidth]{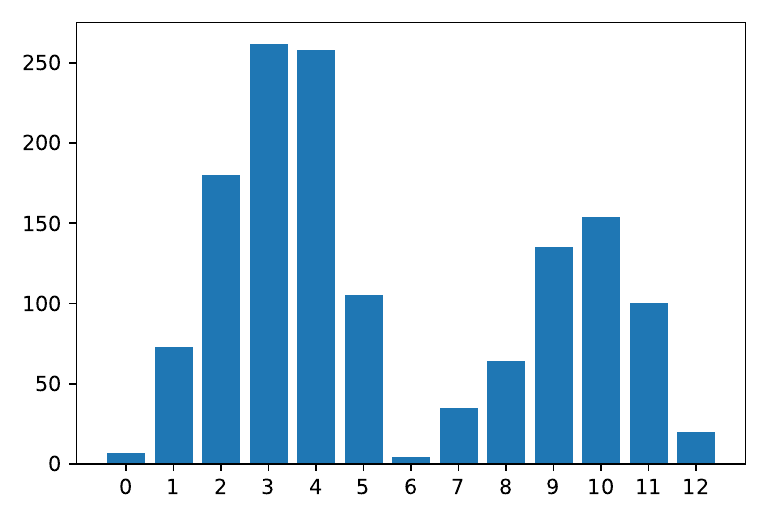}
    \caption{XLM RoBERTa}
    \label{fig:f-xlm-r-syn}
    \end{subfigure}
    \caption{How salient neurons in Syntactic Dependency task distribute across different layers for different models? English (a,b,c), German (d,e,f), French (g,h,i), X-axis = Layer number, Y-axis = Number of neurons selected from that layer.}
    \label{fig:layerwise-multilingual-syn}
%\vspace{-10pt}
\end{figure*}

\begin{figure*}[ht]
    \centering
    \begin{subfigure}[b]{0.95\linewidth}
    \centering
    \includegraphics[width=\linewidth]{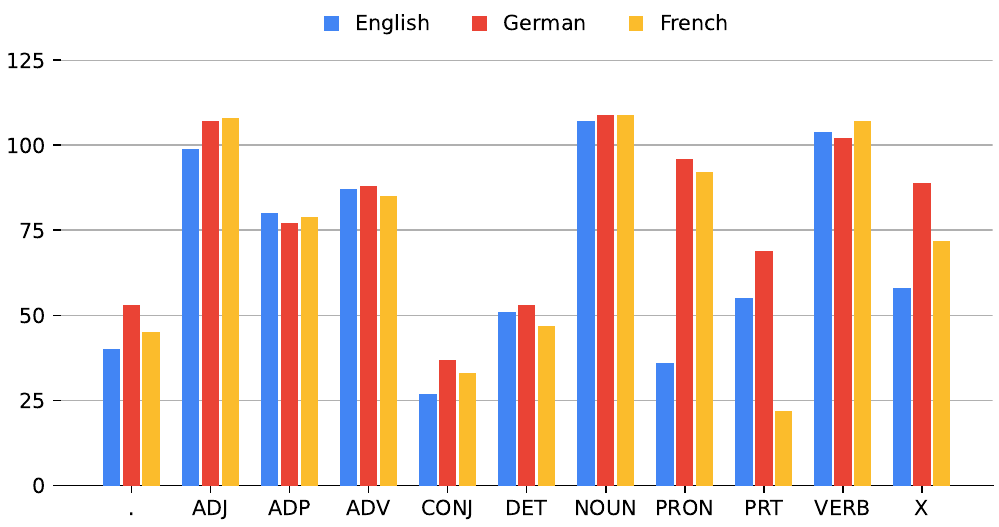}
    \caption{POS tagging}
    \label{fig:mbert_neurons}
    \end{subfigure}
    \begin{subfigure}[b]{0.95\linewidth}
    \centering
    \includegraphics[width=\linewidth]{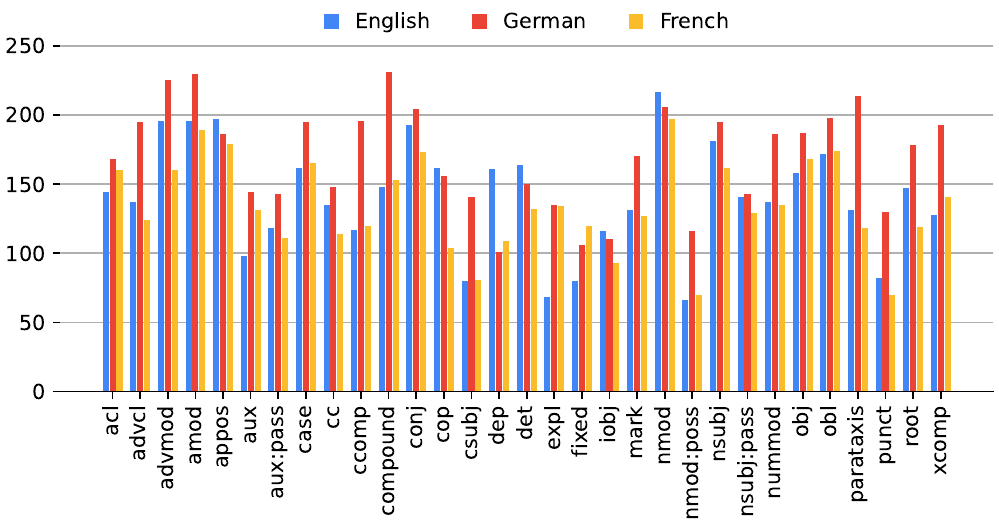}
    \caption{Syntactic Dependency Labeling}
    \label{fig:mbert_neurons_syn}
    \end{subfigure}
\caption{Distribution of Top Neurons Across POS (a) and Syntactic Dependency Properties in Multilingual BERT}
    \label{fig:propertyWise_neurons_multilingual_bert}
%\vspace{-10pt}
\end{figure*}

\vspace{2mm}
\noindent \textbf{Layer-wise Analysis:} Next, we analyze the distribution of salient neurons across layers. Our analysis reveals several findings: i) multilingual models also capture morphology in the lower layers of the model (See Figure \ref{fig:layerwise-multilingual}), ii) the neurons that capture syntactic dependency between a head and its modifier are predominantly found at layers 4 and 5 across different languages and models (See Figure \ref{fig:layerwise-multilingual-syn}), iii) the neuron distribution pattern for multilingual models is quite similar across the three languages. However, monolingual models exhibit slightly different neuron distributions across the three languages. This difference may arise due to the models being trained with different data and optimization functions. For example, FlauBERT only uses the MLM (masked language model) objective and does not include NSP (next sentence prediction) during training. Similarly, the German BERT model is trained with the WWM (whole word masking) optimization function. Such variations can cause minor differences, but the general pattern of morphology captured in the lower-middle part of the network remains consistent across languages and different models. 

\vspace{2mm}
\noindent \textbf{Property-wise Analysis:} Lastly, we analyze the distribution of neurons across individual properties within the task. To carry out this analysis, we converted POS tags for different languages into universal tags using the mapping described in \cite{petrov-etal-2012-universal}. The syntactic dependency labels are shared because we used Universal dependencies. Figure \ref{fig:propertyWise_neurons_multilingual_bert} shows the results for the multilingual BERT model.\footnote{Please see the Appendix for results on XLM-Roberta (see Figure \ref{fig:propertyWise_neurons_multilingual_xlm-r}).} We observe that different properties in the POS tagging task exhibit very consistent neuron distribution patterns across three different languages. Additionally, we found that functional properties such as determiners (DET) and conjunctions (CONJ) are localized to fewer neurons, while open categories such as nouns and verbs require a larger number of neurons to be accurately represented in the multilingual models. In comparison to POS tagging, where we observed harmony among languages, we found that a significantly larger number of neurons were necessary for German to capture different properties in the task of predicting syntactic dependency relations.
% Lastly we analyze the distribution of neurons across individual properties within the task. To carry out this analysis, we converted POS tags for different languages into universal tags using the mapping described in Petrov etl al., 2012 \citep{petrov-etal-2012-universal}. The syntactic dependency labels are shared because we used Universal dependencies. Figure %\ref{fig:mbert_neurons}
% \ref{fig:propertyWise_neurons_multilingual_bert} shows results for the  multilingual BERT model.\footnote{Please See Appendix for results on XLM-Roberta (See Figure \ref{fig:propertyWise_neurons_multilingual_xlm-r}).} We see that different properties in POS tagging task show a very consistent neuron distribution patterns across 3 different languages. We also found our previous observation of functional properties such as determiners (DET) and conjunctions (CONJ) localized to fewer neurons, as opposed to open categories such as nouns and verbs that required larger number of neurons to be true for the multilingual models. In comparison to POS tagging where we observed harmony among languages, we observed that a distinctly larger number of neurons were required for German in capturing different properties on the task of predicting syntactic dependency relation.
%we observed that distinctly larger number of neurons were required for individual properties across different languages on the task of predicting syntactic dependency relation. %More specifically we found German to require a larger number of neurons for each property compared to French and English. 
This can be attributed to the fact that German has a more complex word order. For example German has a complicated verb complex often called as ``separable verbs" that are split across the sentence, requiring the model to capture long distance dependency \citep{fraser-etal-2013-knowledge}. Similarly note how neurons that capture \texttt{compound} category is much higher in German where compounding is very productively used in forming new words \citep{koehn-knight-2003-empirical}. 

\section{Related Work}
\label{sec:literature}

A multitude of papers have delved into the interpretation of deep NLP models, with a focus on two key questions: i) comprehending the acquired knowledge within contextualized representations, known as \textbf{\emph{Concept Analysis}}, and ii) exploring how this information is utilized in the decision-making process, referred to as \textbf{\emph{Attribution Analysis}}. The former entails post-hoc decomposability, where researchers analyze representations to unveil linguistic and non-linguistic phenomena captured during the network's training for any NLP task \citep{belinkov:2017:acl, conneau2018you,sajjad-etal-2022-analyzing}. On the other hand, the latter focuses on characterizing the contributions of model components and input features toward specific predictions \citep{linzen_tacl, gulordava-etal-2018-colorless, marvin-linzen-2018-targeted}. The work done on neuron analysis predominantly falls under \emph{Concept Analysis}. We classify these into a group of three:

\vspace{2mm}

\noindent \textbf{Visualizing} provides a direct approach to understanding neuron function by identifying patterns across a set of sentences manually~\citep{li-etal-2016-visualizing,karpathy2015visualizing}. Yet, due to the sheer number of neurons in a neural network model, visualizing all of them becomes unwieldy. To address this challenge, specific cues have been employed to narrow down the selection of neurons for visualization. These cues include identifying saturated neurons, high/low variance neurons, or omitting dead neurons~\citep{karpathy2015visualizing} when utilizing the ReLU activation function. Despite the effectiveness of visualization in neuron analysis, it does have limitations. Firstly, it is a qualitative technique. Secondly, visualizing neurons with multiple meanings (polysemous neurons) proves challenging. Lastly, not all neurons can be visually interpreted. However, by employing a neuron ranking method such as LCA, as demonstrated in this paper, we can enhance the efficiency of qualitative analysis by focusing on the most prominent neurons.

\vspace{2mm}

\noindent \textbf{Corpus-based Methods} investigate the role of a neuron using techniques such as ranking sentences in a corpus, generating synthetic sentences to maximize its activation, or computing various neuron-level statistics over a corpus. \citet{kadar-etal-2017-representation} utilized corpus ranking to identify words that activate a neuron maximally by analyzing the top-k context from a corpus. In contrast, \citet{Na-ICLR} introduced lexical concepts of different granularity, such as words, phrases, and sentences, using parse trees where each node represented a concept. However, one limitation of corpus ranking is its confined analysis space, limited to the underlying corpus. To overcome this limitation, \citet{poerner-etal-2018-interpretable} generated novel sentences maximizing a neuron's activation, providing a broader understanding of the neuron's role in the model. Moreover, \cite{Mu-Nips} and \cite{suau2020finding} proposed a \emph{Masking-based} method to identify important neurons for a concept. The concept involves identifying the overlap between the presence of a concept in a set of sentences and the high activation values of a neuron. Additionally, \cite{antverg2022on} used data activations to derive a neuron ranking for a concept class, comparing differences in activation values across different concepts within the class. Corpus-based methods address some limitations of the probing framework inherited by LCA. However, unlike LCA, they do not model the selection of groups of neurons collaborating in learning a concept.

\vspace{2mm}
\noindent \textbf{Classifiers} Closer to our work is \cite{torroba-hennigen-etal-2020-intrinsic} and \cite{stanczak2023latentvariable}. They train a generative classifier with the assumption that neurons exhibit a Gaussian distribution.
They fit a multivariate Gaussian over all neurons and extracted individual probes for single neurons. A caveat to their approach is that activations do not always follow a Gaussian prior in practice -- hence restricting their analysis to only the neurons that satisfy this criteria. \citet{chowdhury2024:csl} used the \texttt{LCA} to carry out neuron-wise analyses of pretrained speech models. %Moreover, training a Gaussian-based probe for a large set of neurons 9984 (13x768) is still infeasible, despite  for a multi-class NLP tasks that we have carried out in this paper.

%\paragraph{Unsupervised Methods} 
\vspace{2mm}
% \noindent \textbf{Unsupervised Methods} A diverse set of unsupervised methods have been proposed in the literature. \cite{lakretz-etal-2019-emergence,li-etal-2016-visualizing} performed \emph{Neuron Ablation} to identify the most salient neurons for the network. One limitation of ablation is that  the ranking obtained using single neuron ablation may not be very meaningful in a large neural network. Ablating neurons in group and trying all permutations of neurons or even a subset of neurons is an NP-hard problem \citep{binshtok:2007:AAAI}. %\cite{torroba-hennigen-etal-2020-intrinsic} relied on the assumption that neuron activations follow a \emph{Gaussian distribution} to identify neurons that capture concepts like past and present verbs. 
% \citet{valipur-2019} used \emph{Random Forests} to derive neuron ranking. \cite{alammar2020explaining} used \emph{Matrix Factorization} \citep{olah2018the} to decompose a large matrix into a product of smaller matrices of factors, where each factor represents a group of elements performing similar function. \cite{mayes_under_hood:2020} used UMAP~\citep{mcinnes2020umap} to project neuron activations to a low dimensional space and then performed K-means clustering to group neurons. \cite{bau2018identifying} carried a multi-model search to identify the most salient neurons in the network. They used \emph{Pearson correlation} to compute a similarity score of each neuron of a model with respect to the neurons of other models. 

\noindent \textbf{Unsupervised Methods} The literature offers a diverse array of unsupervised techniques. Researchers such as \cite{lakretz-etal-2019-emergence} and \cite{li-etal-2016-visualizing} have employed \emph{Neuron Ablation} to pinpoint the most significant neurons in the network. However, a drawback of single neuron ablation is that the resulting ranking may lack meaningful interpretation in large neural networks. Ablating neurons in groups and exploring all permutations or a subset of neurons becomes an NP-hard problem \citep{binshtok:2007:AAAI}. Taking a different approach, \citet{valipur-2019} utilized \emph{Random Forests} to establish a neuron ranking. \cite{alammar2020explaining} applied \emph{Matrix Factorization} \citep{olah2018the} to decompose a large matrix into smaller matrices of factors, where each factor represented a group of elements performing a similar function. Similarly, \cite{mayes_under_hood:2020} utilized UMAP~\citep{mcinnes2020umap} to project neuron activations into a lower-dimensional space and subsequently applied K-means clustering to group the neurons. Additionally, \cite{bau2018identifying} conducted a multi-model search to identify the most salient neurons in the network. They employed \emph{Pearson correlation} to compute a similarity score for each neuron in a model concerning the neurons in other models.

%Please see \cite{neuronSurvey} for a comprehensive survey on the neuron analysis.

% \vspace{2mm}
% \noindent \textbf{Classifiers} Other work done using classifier-based approach to discovering salient neurons in the models include \cite{torroba-hennigen-etal-2020-intrinsic} and \cite{valipur-2019}. The former rely on the assumption that neuron activations follow a \emph{Gaussian distribution} and the latter used \emph{Random Forests} to derive neuron ranking. 

% Different from these methods, we introduce a new method called as the \emph{Linguistic Correlation Analysis} based on the diagnostic classifier framework \citep{Hupkes}. Our method trains a posthoc classifier on the underlying representations to identify neurons with respect to a pre-defined concept. We carry out a wide-scale exploration across core linguistic concepts and perform a comparative architectural analysis of popular transformer models.

\paragraph{Attribution Neurons} The work on analyzing how knowledge is encoded within the learned representations, does not address the question whether it is used by the model during prediction is a less explored area \citep{feder-etal-2021-causalm, elazar-etal-2021-amnesic}. There is however, another body of work called \textit{Attribution Analysis} that focuses on the methods that characterize the role of neurons and layers towards a specific prediction.  Some of the notable methods 
help us to identify important neurons (w.r.t prediction) in different layers of a deep neural network, include gradient and perturbation-based attribution algorithms such as Integrated Gradients~\citep{ig}, Layer Conductance~\citep{cond}, Saliency~\citep{saliency} and SHapley Additive exPlanations(SHAP)~\citep{shappely_NIPS2017_7062}. More recent and advanced attribution algorithms that take feature or neuron interactions into account include Integrated Hessians~\citep{janizek2020explaining}, Shapely Taylor index~\citep{dhamdhere2020shapley} and Archipelago~\citep{achipelago}. However, these methods do not provide linguistic explanation for the neurons. By connecting concept analysis with the attribution analysis we can bridge this gap. We leave this exploration for the future.

\section{Conclusion}
\label{sec:conclusion}

In this article, we presented \emph{Linguistic Correlation Analysis} to discover salient neurons in deep NLP models. We train linear probes with elastic-net regularization to extract a neuron ranking of a pre-trained model, with respect to any extrinsic property of interest. We discovered and analyzed individual neurons across a variety of neural language models on the task of predicting core linguistic properties (morphology, syntax and semantics). Our results reinforce previous findings and also illuminate further insights:

\begin{itemize}
    \item Although neural language models contain highly distributed information, it is feasible to distill a limited set of neurons to perform subsequent natural language processing tasks.
    \item The number of salient neurons depends on the complexity of the property in question. For instance, functional classes like closed-class words tend to be concentrated in a smaller number of neurons, while open-class words are spread across a larger neural network.
    \item Information is redundantly dispersed, with multiple subsets of neurons carrying identical information coexisting within the network.
    \item Neurons specializing in acquiring basic lexical details like suffixation or word structure are concentrated in the lower layers of the model, whereas those capturing non-local dependencies are primarily situated in the middle and higher layers of the network.
    \item During transfer learning, the neurons responsible for learning linguistic information are regressed to lower layers in the model, as the higher layers are optimized for specific downstream tasks.
    \item Fascinating distinctions between architectures were noticed: in XLNet, neurons were concentrated within fewer layers, whereas in BERT, they remained spread throughout the network even after fine-tuning.
    \item Our research revealed that multilingual models demonstrated a comparable distribution of neurons when learning concepts across various languages.

\end{itemize}

\paragraph{Future Work} Most of the work done on neuron analysis including ours only highlights the knowledge encoded within the learned representations. The question whether it is actually used by the model during prediction has been largely unanswered. A few recent studies carried to answer this question such as \cite{feder-etal-2021-causalm} and \cite{elazar-etal-2021-amnesic} analyze representations at a more holistic level and ignore individual neurons. More recently \cite{antverg2022on} showed that high probing accuracy does not necessarily mean that the information is important by the model. In the future, we would try to establish connection between encoded knowledge versus how it is utilized during prediction to enable richer explanations.

%\linenumbers
\bibliography{anthology,eacl}

\begin{thebibliography}{88}
\providecommand{\natexlab}[1]{#1}
\providecommand{\url}[1]{\texttt{#1}}
\expandafter\ifx\csname urlstyle\endcsname\relax
  \providecommand{\doi}[1]{doi: #1}\else
  \providecommand{\doi}{doi: \begingroup \urlstyle{rm}\Url}\fi

\bibitem[Abzianidze et~al.(2017)Abzianidze, Bjerva, Evang, Haagsma, van Noord, Ludmann, Nguyen, and Bos]{abzianidze-EtAl:2017:EACLshort}
Lasha Abzianidze, Johannes Bjerva, Kilian Evang, Hessel Haagsma, Rik van Noord, Pierre Ludmann, Duc-Duy Nguyen, and Johan Bos.
\newblock The parallel meaning bank: Towards a multilingual corpus of translations annotated with compositional meaning representations.
\newblock In \emph{Proceedings of the 15th Conference of the European Chapter of the Association for Computational Linguistics}, EACL~'17, pages 242--247, Valencia, Spain, 2017.

\bibitem[Adi et~al.(2016)Adi, Kermany, Belinkov, Lavi, and Goldberg]{adi2016fine}
Yossi Adi, Einat Kermany, Yonatan Belinkov, Ofer Lavi, and Yoav Goldberg.
\newblock {Fine-grained Analysis of Sentence Embeddings Using Auxiliary Prediction Tasks}.
\newblock \emph{arXiv preprint arXiv:1608.04207}, 14\penalty0 (14), 2016.

\bibitem[Alammar(2020)]{alammar2020explaining}
J~Alammar.
\newblock Interfaces for explaining transformer language models, 2020.
\newblock URL \url{https://jalammar.github.io/explaining-transformers/}.

\bibitem[Antverg and Belinkov(2022)]{antverg2022on}
Omer Antverg and Yonatan Belinkov.
\newblock On the pitfalls of analyzing individual neurons in language models.
\newblock In \emph{International Conference on Learning Representations}, 2022.
\newblock URL \url{https://openreview.net/forum?id=8uz0EWPQIMu}.

\bibitem[Bau et~al.(2019)Bau, Belinkov, Sajjad, Durrani, Dalvi, and Glass]{bau2018identifying}
Anthony Bau, Yonatan Belinkov, Hassan Sajjad, Nadir Durrani, Fahim Dalvi, and James Glass.
\newblock Identifying and controlling important neurons in neural machine translation.
\newblock In \emph{International Conference on Learning Representations}, 2019.
\newblock URL \url{https://openreview.net/forum?id=H1z-PsR5KX}.

\bibitem[Belinkov et~al.(2017{\natexlab{a}})Belinkov, Durrani, Dalvi, Sajjad, and Glass]{belinkov:2017:acl}
Yonatan Belinkov, Nadir Durrani, Fahim Dalvi, Hassan Sajjad, and James Glass.
\newblock {What do Neural Machine Translation Models Learn about Morphology?}
\newblock In \emph{Proceedings of the 55th Annual Meeting of the Association for Computational Linguistics (ACL)}, Vancouver, July 2017{\natexlab{a}}. Association for Computational Linguistics.
\newblock URL \url{https://aclanthology.coli.uni-saarland.de/pdf/P/P17/P17-1080.pdf}.

\bibitem[Belinkov et~al.(2017{\natexlab{b}})Belinkov, M\`arquez, Sajjad, Durrani, Dalvi, and Glass]{belinkov:2017:ijcnlp}
Yonatan Belinkov, Llu\'{i}s M\`arquez, Hassan Sajjad, Nadir Durrani, Fahim Dalvi, and James Glass.
\newblock {Evaluating Layers of Representation in Neural Machine Translation on Part-of-Speech and Semantic Tagging Tasks}.
\newblock In \emph{Proceedings of the 8th International Joint Conference on Natural Language Processing (IJCNLP)}, November 2017{\natexlab{b}}.

\bibitem[Belinkov et~al.(2020)Belinkov, Durrani, Dalvi, Sajjad, and Glass]{belinkov-etal-2020-analysis}
Yonatan Belinkov, Nadir Durrani, Fahim Dalvi, Hassan Sajjad, and James Glass.
\newblock On the linguistic representational power of neural machine translation models.
\newblock \emph{Computational Linguistics}, 45\penalty0 (1):\penalty0 1--57, March 2020.

\bibitem[Bentivogli et~al.(2009)Bentivogli, Dagan, Dang, Giampiccolo, and Magnini]{Bentivogli09thefifth}
Luisa Bentivogli, Ido Dagan, Hoa~Trang Dang, Danilo Giampiccolo, and Bernardo Magnini.
\newblock The fifth pascal recognizing textual entailment challenge.
\newblock In \emph{In Proc Text Analysis Conference (TAC’09}, 2009.

\bibitem[Binshtok et~al.(2007)Binshtok, Brafman, Shimony, Martin, and Boutilier]{binshtok:2007:AAAI}
Maxim Binshtok, Ronen~I Brafman, Solomon~Eyal Shimony, Ajay Martin, and Crag Boutilier.
\newblock Computing optimal subsets.
\newblock In \emph{Proceedings of the Twenty Second AAAI Conference on Artificial Intelligence (AAAI, Oral presentation)}, July 2007.

\bibitem[Bojar et~al.(2014)Bojar, Buck, Federmann, Haddow, Koehn, Leveling, Monz, Pecina, Post, Saint-Amand, Soricut, Specia, and Tamchyna]{bojar-etal-2014-findings}
Ond{\v{r}}ej Bojar, Christian Buck, Christian Federmann, Barry Haddow, Philipp Koehn, Johannes Leveling, Christof Monz, Pavel Pecina, Matt Post, Herve Saint-Amand, Radu Soricut, Lucia Specia, and Ale{\v{s}} Tamchyna.
\newblock Findings of the 2014 workshop on statistical machine translation.
\newblock In \emph{Proceedings of the Ninth Workshop on Statistical Machine Translation}, pages 12--58, Baltimore, Maryland, USA, June 2014. Association for Computational Linguistics.
\newblock \doi{10.3115/v1/W14-3302}.
\newblock URL \url{https://aclanthology.org/W14-3302}.

\bibitem[Cer et~al.(2017)Cer, Diab, Agirre, Lopez-Gazpio, and Specia]{cer-etal-2017-semeval}
Daniel Cer, Mona Diab, Eneko Agirre, I{\~n}igo Lopez-Gazpio, and Lucia Specia.
\newblock {S}em{E}val-2017 task 1: Semantic textual similarity multilingual and crosslingual focused evaluation.
\newblock In \emph{Proceedings of the 11th International Workshop on Semantic Evaluation ({S}em{E}val-2017)}, pages 1--14, Vancouver, Canada, August 2017. Association for Computational Linguistics.
\newblock \doi{10.18653/v1/S17-2001}.
\newblock URL \url{https://www.aclweb.org/anthology/S17-2001}.

\bibitem[Chan et~al.(2020)Chan, Schweter, and M{\"o}ller]{chan-etal-2020-germans}
Branden Chan, Stefan Schweter, and Timo M{\"o}ller.
\newblock {G}erman{'}s next language model.
\newblock In \emph{Proceedings of the 28th International Conference on Computational Linguistics}, pages 6788--6796, Barcelona, Spain (Online), December 2020. International Committee on Computational Linguistics.
\newblock \doi{10.18653/v1/2020.coling-main.598}.
\newblock URL \url{https://aclanthology.org/2020.coling-main.598}.

\bibitem[Chowdhury et~al.(2024)Chowdhury, Durrani, and Ali]{chowdhury2024:csl}
Shammur~Absar Chowdhury, Nadir Durrani, and Ahmed Ali.
\newblock What do end-to-end speech models learn about speaker, language and channel information? a layer-wise and neuron-level analysis.
\newblock \emph{Computer Speech and Language}, 83:\penalty0 101539, 2024.
\newblock ISSN 0885-2308.
\newblock \doi{https://doi.org/10.1016/j.csl.2023.101539}.
\newblock URL \url{https://www.sciencedirect.com/science/article/pii/S088523082300058X}.

\bibitem[Conneau et~al.(2018)Conneau, Kruszewski, Lample, Barrault, and Baroni]{conneau2018you}
Alexis Conneau, German Kruszewski, Guillaume Lample, Lo{\"\i}c Barrault, and Marco Baroni.
\newblock {What you can cram into a single vector: Probing sentence embeddings for linguistic properties}.
\newblock In \emph{Proceedings of the 56th Annual Meeting of the Association for Computational Linguistics (ACL)}, July 2018.

\bibitem[Conneau et~al.(2020)Conneau, Khandelwal, Goyal, Chaudhary, Wenzek, Guzmán, Grave, Ott, Zettlemoyer, and Stoyanov]{conneau2020unsupervised}
Alexis Conneau, Kartikay Khandelwal, Naman Goyal, Vishrav Chaudhary, Guillaume Wenzek, Francisco Guzmán, Edouard Grave, Myle Ott, Luke Zettlemoyer, and Veselin Stoyanov.
\newblock Unsupervised cross-lingual representation learning at scale, 2020.

\bibitem[Dalvi et~al.(2017)Dalvi, Durrani, Sajjad, Belinkov, and Vogel]{dalvi:2017:ijcnlp}
Fahim Dalvi, Nadir Durrani, Hassan Sajjad, Yonatan Belinkov, and Stephan Vogel.
\newblock {Understanding and Improving Morphological Learning in the Neural Machine Translation Decoder}.
\newblock In \emph{Proceedings of the 8th International Joint Conference on Natural Language Processing (IJCNLP)}, November 2017.

\bibitem[Dalvi et~al.(2019)Dalvi, Durrani, Sajjad, Belinkov, Bau, and Glass]{dalvi:2019:AAAI}
Fahim Dalvi, Nadir Durrani, Hassan Sajjad, Yonatan Belinkov, D.~Anthony Bau, and James Glass.
\newblock What is one grain of sand in the desert? analyzing individual neurons in deep nlp models.
\newblock In \emph{Proceedings of the Thirty-Third AAAI Conference on Artificial Intelligence (AAAI, Oral presentation)}, January 2019.

\bibitem[Dalvi et~al.(2020)Dalvi, Sajjad, Durrani, and Belinkov]{dalvi-2020-CCFS}
Fahim Dalvi, Hassan Sajjad, Nadir Durrani, and Yonatan Belinkov.
\newblock Analyzing redundancy in pretrained transformer models.
\newblock In \emph{Proceedings of the 2020 Conference on Empirical Methods in Natural Language Processing (EMNLP-2020)}, Online, November 2020. Association for Computational Linguistics.

\bibitem[Dalvi et~al.(2023)Dalvi, Durrani, and Sajjad]{dalvi-etal-2023-neurox}
Fahim Dalvi, Nadir Durrani, and Hassan Sajjad.
\newblock Neurox library for neuron analysis of deep nlp models.
\newblock In \emph{Proceedings of the 61st Annual Meeting of the Association for Computational Linguistics: System Demonstrations}, pages 75--83, Toronto, Canada, July 2023. Association for Computational Linguistics.

\bibitem[Devlin et~al.(2019)Devlin, Chang, Lee, and Toutanova]{devlin-etal-2019-bert}
Jacob Devlin, Ming-Wei Chang, Kenton Lee, and Kristina Toutanova.
\newblock {BERT}: Pre-training of deep bidirectional transformers for language understanding.
\newblock In \emph{Proceedings of the 2019 Conference of the North {A}merican Chapter of the Association for Computational Linguistics: Human Language Technologies, Volume 1 (Long and Short Papers)}, Minneapolis, Minnesota, 2019. Association for Computational Linguistics.

\bibitem[Dhamdhere et~al.(2018)Dhamdhere, Sundararajan, and Yan]{cond}
Kedar Dhamdhere, Mukund Sundararajan, and Qiqi Yan.
\newblock How important is a neuron?, 2018.

\bibitem[Dhamdhere et~al.(2020)Dhamdhere, Agarwal, and Sundararajan]{dhamdhere2020shapley}
Kedar Dhamdhere, Ashish Agarwal, and Mukund Sundararajan.
\newblock The shapley taylor interaction index, 2020.

\bibitem[Dolan and Brockett(2005)]{dolan-brockett-2005-automatically}
William~B. Dolan and Chris Brockett.
\newblock Automatically constructing a corpus of sentential paraphrases.
\newblock In \emph{Proceedings of the Third International Workshop on Paraphrasing ({IWP}2005)}, 2005.
\newblock URL \url{https://www.aclweb.org/anthology/I05-5002}.

\bibitem[Durrani et~al.(2019)Durrani, Dalvi, Sajjad, Belinkov, and Nakov]{durrani-etal-2019-one}
Nadir Durrani, Fahim Dalvi, Hassan Sajjad, Yonatan Belinkov, and Preslav Nakov.
\newblock One size does not fit all: Comparing {NMT} representations of different granularities.
\newblock In \emph{Proceedings of the 2019 Conference of the North {A}merican Chapter of the Association for Computational Linguistics: Human Language Technologies, Volume 1 (Long and Short Papers)}, pages 1504--1516, Minneapolis, Minnesota, June 2019. Association for Computational Linguistics.
\newblock \doi{10.18653/v1/N19-1154}.
\newblock URL \url{https://www.aclweb.org/anthology/N19-1154}.

\bibitem[Durrani et~al.(2022)Durrani, Sajjad, Dalvi, and Alam]{durrani-etal-2022-latent}
Nadir Durrani, Hassan Sajjad, Fahim Dalvi, and Firoj Alam.
\newblock On the transformation of latent space in fine-tuned nlp models.
\newblock In \emph{Proceedings of the 2022 Conference on Empirical Methods in Natural Language Processing (EMNLP)}, pages 1495--1516, Abu Dhabi, UAE, December 2022. Association for Computational Linguistics.
\newblock \doi{10.18653/v1/2020.emnlp-main.395}.
\newblock URL \url{https://aclanthology.org/2022.emnlp-main.97}.

\bibitem[Elazar et~al.(2021)Elazar, Ravfogel, Jacovi, and Goldberg]{elazar-etal-2021-amnesic}
Yanai Elazar, Shauli Ravfogel, Alon Jacovi, and Yoav Goldberg.
\newblock Amnesic probing: Behavioral explanation with amnesic counterfactuals.
\newblock \emph{Transactions of the Association for Computational Linguistics}, 9:\penalty0 160--175, 2021.
\newblock \doi{10.1162/tacl_a_00359}.
\newblock URL \url{https://aclanthology.org/2021.tacl-1.10}.

\bibitem[Feder et~al.(2021)Feder, Oved, Shalit, and Reichart]{feder-etal-2021-causalm}
Amir Feder, Nadav Oved, Uri Shalit, and Roi Reichart.
\newblock {C}ausa{LM}: Causal model explanation through counterfactual language models.
\newblock \emph{Computational Linguistics}, 47\penalty0 (2):\penalty0 333--386, June 2021.
\newblock \doi{10.1162/coli_a_00404}.
\newblock URL \url{https://aclanthology.org/2021.cl-2.13}.

\bibitem[Fraser et~al.(2013)Fraser, Schmid, Farkas, Wang, and Sch{\"u}tze]{fraser-etal-2013-knowledge}
Alexander Fraser, Helmut Schmid, Rich{\'a}rd Farkas, Renjing Wang, and Hinrich Sch{\"u}tze.
\newblock Knowledge sources for constituent parsing of {G}erman, a morphologically rich and less-configurational language.
\newblock \emph{Computational Linguistics}, 39\penalty0 (1):\penalty0 57--85, March 2013.
\newblock \doi{10.1162/COLI_a_00135}.
\newblock URL \url{https://aclanthology.org/J13-1005}.

\bibitem[Gulordava et~al.(2018)Gulordava, Bojanowski, Grave, Linzen, and Baroni]{gulordava-etal-2018-colorless}
Kristina Gulordava, Piotr Bojanowski, Edouard Grave, Tal Linzen, and Marco Baroni.
\newblock Colorless green recurrent networks dream hierarchically.
\newblock In \emph{Proceedings of the 2018 Conference of the North {A}merican Chapter of the Association for Computational Linguistics: Human Language Technologies, Volume 1 (Long Papers)}, pages 1195--1205, New Orleans, Louisiana, June 2018. Association for Computational Linguistics.
\newblock \doi{10.18653/v1/N18-1108}.
\newblock URL \url{https://www.aclweb.org/anthology/N18-1108}.

\bibitem[Hastie et~al.(2013)Hastie, Tibshirani, and Friedman]{hastie2013elements}
T.~Hastie, R.~Tibshirani, and J.~Friedman.
\newblock \emph{The Elements of Statistical Learning: Data Mining, Inference, and Prediction}.
\newblock Springer Series in Statistics. Springer New York, 2013.
\newblock ISBN 9781489905185.
\newblock URL \url{https://books.google.com.qa/books?id=DdQDmwEACAAJ}.

\bibitem[Hewitt and Liang(2019)]{hewitt-liang-2019-designing}
John Hewitt and Percy Liang.
\newblock Designing and interpreting probes with control tasks.
\newblock In \emph{Proceedings of the 2019 Conference on Empirical Methods in Natural Language Processing and the 9th International Joint Conference on Natural Language Processing (EMNLP-IJCNLP)}, pages 2733--2743, Hong Kong, China, November 2019. Association for Computational Linguistics.
\newblock \doi{10.18653/v1/D19-1275}.
\newblock URL \url{https://www.aclweb.org/anthology/D19-1275}.

\bibitem[Hockenmaier(2006)]{hockenmaier2006creating}
Julia Hockenmaier.
\newblock Creating a {CCGbank} and a wide-coverage {CCG} lexicon for {G}erman.
\newblock In \emph{Proceedings of the 21st International Conference on Computational Linguistics and 44th Annual Meeting of the Association for Computational Linguistics}, ACL~'06, pages 505--512, Sydney, Australia, 2006.

\bibitem[Hupkes et~al.(2017)Hupkes, Veldhoen, and Zuidema]{Hupkes}
Dieuwke Hupkes, Sara Veldhoen, and Willem~H. Zuidema.
\newblock Visualisation and 'diagnostic classifiers' reveal how recurrent and recursive neural networks process hierarchical structure.
\newblock \emph{CoRR}, abs/1711.10203, 2017.
\newblock URL \url{http://arxiv.org/abs/1711.10203}.

\bibitem[Janizek et~al.(2020)Janizek, Sturmfels, and Lee]{janizek2020explaining}
Joseph~D. Janizek, Pascal Sturmfels, and Su-In Lee.
\newblock Explaining explanations: Axiomatic feature interactions for deep networks, 2020.

\bibitem[K{\'a}d{\'a}r et~al.(2017)K{\'a}d{\'a}r, Chrupa{\l}a, and Alishahi]{kadar-etal-2017-representation}
{\'A}kos K{\'a}d{\'a}r, Grzegorz Chrupa{\l}a, and Afra Alishahi.
\newblock Representation of linguistic form and function in recurrent neural networks.
\newblock \emph{Computational Linguistics}, 43\penalty0 (4):\penalty0 761--780, December 2017.
\newblock \doi{10.1162/COLI_a_00300}.
\newblock URL \url{https://www.aclweb.org/anthology/J17-4003}.

\bibitem[Karpathy et~al.(2015)Karpathy, Johnson, and Fei-Fei]{karpathy2015visualizing}
Andrej Karpathy, Justin Johnson, and Li~Fei-Fei.
\newblock Visualizing and understanding recurrent networks, 2015.

\bibitem[Kim et~al.(2020)Kim, Choi, Edmiston, and goo Lee]{kim2020pretrained}
Taeuk Kim, Jihun Choi, Daniel Edmiston, and Sang goo Lee.
\newblock Are pre-trained language models aware of phrases? simple but strong baselines for grammar induction, 2020.

\bibitem[Kingma and Ba(2014)]{kingma2014adam}
Diederik Kingma and Jimmy Ba.
\newblock {Adam: A Method for Stochastic Optimization}.
\newblock \emph{arXiv preprint arXiv:1412.6980}, 14\penalty0 (14), 2014.

\bibitem[Koehn and Knight(2003)]{koehn-knight-2003-empirical}
Philipp Koehn and Kevin Knight.
\newblock Empirical methods for compound splitting.
\newblock In \emph{10th Conference of the {E}uropean Chapter of the Association for Computational Linguistics}, Budapest, Hungary, April 2003. Association for Computational Linguistics.
\newblock URL \url{https://aclanthology.org/E03-1076}.

\bibitem[Lakretz et~al.(2019)Lakretz, Kruszewski, Desbordes, Hupkes, Dehaene, and Baroni]{lakretz-etal-2019-emergence}
Yair Lakretz, German Kruszewski, Theo Desbordes, Dieuwke Hupkes, Stanislas Dehaene, and Marco Baroni.
\newblock The emergence of number and syntax units in {LSTM} language models.
\newblock In \emph{Proceedings of the 2019 Conference of the North {A}merican Chapter of the Association for Computational Linguistics: Human Language Technologies, Volume 1 (Long and Short Papers)}, pages 11--20, Minneapolis, Minnesota, June 2019. Association for Computational Linguistics.
\newblock \doi{10.18653/v1/N19-1002}.
\newblock URL \url{https://www.aclweb.org/anthology/N19-1002}.

\bibitem[Le et~al.(2020)Le, Vial, Frej, Segonne, Coavoux, Lecouteux, Allauzen, Crabb{\'e}, Besacier, and Schwab]{le-etal-2020-flaubert-unsupervised}
Hang Le, Lo{\"\i}c Vial, Jibril Frej, Vincent Segonne, Maximin Coavoux, Benjamin Lecouteux, Alexandre Allauzen, Benoit Crabb{\'e}, Laurent Besacier, and Didier Schwab.
\newblock {F}lau{BERT}: Unsupervised language model pre-training for {F}rench.
\newblock In \emph{Proceedings of the 12th Language Resources and Evaluation Conference}, pages 2479--2490, Marseille, France, May 2020. European Language Resources Association.
\newblock ISBN 979-10-95546-34-4.
\newblock URL \url{https://aclanthology.org/2020.lrec-1.302}.

\bibitem[Lertvittayakumjorn et~al.(2020)Lertvittayakumjorn, Specia, and Toni]{lertvittayakumjorn-etal-2020-find}
Piyawat Lertvittayakumjorn, Lucia Specia, and Francesca Toni.
\newblock {FIND}: {H}uman-in-the-{L}oop {D}ebugging {D}eep {T}ext {C}lassifiers.
\newblock In \emph{Proceedings of the 2020 Conference on Empirical Methods in Natural Language Processing (EMNLP)}, pages 332--348, Online, November 2020. Association for Computational Linguistics.
\newblock \doi{10.18653/v1/2020.emnlp-main.24}.
\newblock URL \url{https://aclanthology.org/2020.emnlp-main.24}.

\bibitem[Li et~al.(2016)Li, Chen, Hovy, and Jurafsky]{li-etal-2016-visualizing}
Jiwei Li, Xinlei Chen, Eduard Hovy, and Dan Jurafsky.
\newblock Visualizing and understanding neural models in {NLP}.
\newblock In \emph{Proceedings of the 2016 Conference of the North {A}merican Chapter of the Association for Computational Linguistics: Human Language Technologies}, pages 681--691, San Diego, California, June 2016. Association for Computational Linguistics.
\newblock \doi{10.18653/v1/N16-1082}.
\newblock URL \url{https://www.aclweb.org/anthology/N16-1082}.

\bibitem[Linzen et~al.(2016{\natexlab{a}})Linzen, Dupoux, and Goldberg]{linzen2016assessing}
Tal Linzen, Emmanuel Dupoux, and Yoav Goldberg.
\newblock {Assessing the Ability of LSTMs to Learn Syntax-Sensitive Dependencies}.
\newblock \emph{Transactions of the Association for Computational Linguistics}, 4:\penalty0 521--535, 2016{\natexlab{a}}.

\bibitem[Linzen et~al.(2016{\natexlab{b}})Linzen, Dupoux, and Goldberg]{linzen_tacl}
Tal Linzen, Emmanuel Dupoux, and Yoav Goldberg.
\newblock {Assessing the ability of LSTMs to learn syntax-sensitive dependencies}.
\newblock \emph{Transactions of the Association for Computational Linguistics}, 4:521– 535, 2016{\natexlab{b}}.

\bibitem[Liu et~al.(2019{\natexlab{a}})Liu, Gardner, Belinkov, Peters, and Smith]{liu-etal-2019-linguistic}
Nelson~F. Liu, Matt Gardner, Yonatan Belinkov, Matthew~E. Peters, and Noah~A. Smith.
\newblock Linguistic knowledge and transferability of contextual representations.
\newblock In \emph{Proceedings of the 2019 Conference of the North {A}merican Chapter of the Association for Computational Linguistics: Human Language Technologies, Volume 1 (Long and Short Papers)}, pages 1073--1094, Minneapolis, Minnesota, June 2019{\natexlab{a}}. Association for Computational Linguistics.
\newblock URL \url{https://www.aclweb.org/anthology/N19-1112}.

\bibitem[Liu et~al.(2019{\natexlab{b}})Liu, Ott, Goyal, Du, Joshi, Chen, Levy, Lewis, Zettlemoyer, and Stoyanov]{liu2019roberta}
Yinhan Liu, Myle Ott, Naman Goyal, Jingfei Du, Mandar Joshi, Danqi Chen, Omer Levy, Mike Lewis, Luke Zettlemoyer, and Veselin Stoyanov.
\newblock Roberta: A robustly optimized bert pretraining approach, 2019{\natexlab{b}}.

\bibitem[Lundberg and Lee(2017)]{shappely_NIPS2017_7062}
Scott~M Lundberg and Su-In Lee.
\newblock A unified approach to interpreting model predictions.
\newblock In I.~Guyon, U.~V. Luxburg, S.~Bengio, H.~Wallach, R.~Fergus, S.~Vishwanathan, and R.~Garnett, editors, \emph{Advances in Neural Information Processing Systems 30}, pages 4765--4774. Curran Associates, Inc., 2017.
\newblock URL \url{http://papers.nips.cc/paper/7062-a-unified-approach-to-interpreting-model-predictions.pdf}.

\bibitem[Ma et~al.(2018)Ma, Liu, Lee, Zhang, and Grama]{MODE}
Shiqing Ma, Yingqi Liu, {Wen Chuan} Lee, Xiangyu Zhang, and Ananth Grama.
\newblock Mode: Automated neural network model debugging via state differential analysis and input selection.
\newblock In Alessandro Garci, {Corina S.} Pasareanu, and {Gary T.} Leavens, editors, \emph{ESEC/FSE 2018 - Proceedings of the 2018 26th ACM Joint Meeting on European So ftware Engineering Conference and Symposium on the Foundations of So ftware Engineering}, ESEC/FSE 2018 - Proceedings of the 2018 26th ACM Joint Meeting on European Software Engineering Conference and Symposium on the Foundations of Software Engineering, pages 175--186. Association for Computing Machinery, Inc, October 2018.
\newblock \doi{10.1145/3236024.3236082}.
\newblock Publisher Copyright: {\textcopyright} 2018 Association for Computing Machinery.; 26th ACM Joint European Software Engineering Conference and Symposium on the Foundations of Software Engineering, ESEC/FSE 2018 ; Conference date: 04-11-2018 Through 09-11-2018.

\bibitem[Marcus et~al.(1993)Marcus, Santorini, and Marcinkiewicz]{marcus-etal-1993-building}
Mitchell~P. Marcus, Beatrice Santorini, and Mary~Ann Marcinkiewicz.
\newblock Building a large annotated corpus of {E}nglish: The {P}enn {T}reebank.
\newblock \emph{Computational Linguistics}, 19\penalty0 (2):\penalty0 313--330, 1993.
\newblock URL \url{https://www.aclweb.org/anthology/J93-2004}.

\bibitem[Marvin and Linzen(2018)]{marvin-linzen-2018-targeted}
Rebecca Marvin and Tal Linzen.
\newblock Targeted syntactic evaluation of language models.
\newblock In \emph{Proceedings of the 2018 Conference on Empirical Methods in Natural Language Processing}, pages 1192--1202, Brussels, Belgium, October-November 2018. Association for Computational Linguistics.
\newblock \doi{10.18653/v1/D18-1151}.
\newblock URL \url{https://www.aclweb.org/anthology/D18-1151}.

\bibitem[Mclnnes et~al.(2020)Mclnnes, Healy, and Melville]{mcinnes2020umap}
Leland Mclnnes, John Healy, and James Melville.
\newblock {UMAP}: Uniform manifold approximation and projection for dimension reduction, 2020.

\bibitem[Meyes et~al.(2020)Meyes, de~Puiseau, Posada{-}Moreno, and Meisen]{mayes_under_hood:2020}
Richard Meyes, Constantin~Waubert de~Puiseau, Andres Posada{-}Moreno, and Tobias Meisen.
\newblock Under the hood of neural networks: Characterizing learned representations by functional neuron populations and network ablations.
\newblock \emph{CoRR}, abs/2004.01254, 2020.

\bibitem[Michel et~al.(2019)Michel, Levy, and Neubig]{sixteenHeads}
Paul Michel, Omer Levy, and Graham Neubig.
\newblock Are sixteen heads really better than one?
\newblock \emph{CoRR}, abs/1905.10650, 2019.
\newblock URL \url{http://arxiv.org/abs/1905.10650}.

\bibitem[Mousi et~al.(2023)Mousi, Durrani, and Dalvi]{mousi2023llms}
Basel Mousi, Nadir Durrani, and Fahim Dalvi.
\newblock Can llms facilitate interpretation of pre-trained language models?
\newblock In \emph{Proceedings of the 2023 Conference on Empirical Methods in Natural Language Processing (EMNLP)}, Singapore, dec 2023. Association for Computational Linguistics.

\bibitem[Mu and Andreas(2020)]{Mu-Nips}
Jesse Mu and Jacob Andreas.
\newblock Compositional explanations of neurons.
\newblock \emph{CoRR}, abs/2006.14032, 2020.
\newblock URL \url{https://arxiv.org/abs/2006.14032}.

\bibitem[Na et~al.(2019)Na, Choe, Lee, and Kim]{Na-ICLR}
Seil Na, Yo~Joong Choe, Dong{-}Hyun Lee, and Gunhee Kim.
\newblock Discovery of natural language concepts in individual units of cnns.
\newblock \emph{CoRR}, abs/1902.07249, 2019.
\newblock URL \url{http://arxiv.org/abs/1902.07249}.

\bibitem[Nguyen et~al.(2014)Nguyen, Nguyen, Pham, and Pham]{nguyen-EtAl:2014:Demos}
Dat~Quoc Nguyen, Dai~Quoc Nguyen, Dang~Duc Pham, and Son~Bao Pham.
\newblock Rdrpostagger: A ripple down rules-based part-of-speech tagger.
\newblock In \emph{Proceedings of the Demonstrations at the 14th Conference of the European Chapter of the Association for Computational Linguistics}, pages 17--20, Gothenburg, Sweden, April 2014. Association for Computational Linguistics.
\newblock URL \url{http://www.aclweb.org/anthology/E14-2005}.

\bibitem[Olah et~al.(2018)Olah, Satyanarayan, Johnson, Carter, Schubert, Ye, and Mordvintsev]{olah2018the}
Chris Olah, Arvind Satyanarayan, Ian Johnson, Shan Carter, Ludwig Schubert, Katherine Ye, and Alexander Mordvintsev.
\newblock The building blocks of interpretability.
\newblock \emph{Distill}, 14\penalty0 (14), 2018.
\newblock \doi{10.23915/distill.00010}.
\newblock https://distill.pub/2018/building-blocks.

\bibitem[Petrov et~al.(2012)Petrov, Das, and McDonald]{petrov-etal-2012-universal}
Slav Petrov, Dipanjan Das, and Ryan McDonald.
\newblock A universal part-of-speech tagset.
\newblock In \emph{Proceedings of the Eighth International Conference on Language Resources and Evaluation ({LREC}'12)}, pages 2089--2096, Istanbul, Turkey, May 2012. European Language Resources Association (ELRA).
\newblock URL \url{http://www.lrec-conf.org/proceedings/lrec2012/pdf/274_Paper.pdf}.

\bibitem[Poerner et~al.(2018)Poerner, Roth, and Sch{\"u}tze]{poerner-etal-2018-interpretable}
Nina Poerner, Benjamin Roth, and Hinrich Sch{\"u}tze.
\newblock Interpretable textual neuron representations for {NLP}.
\newblock In \emph{Proceedings of the 2018 {EMNLP} Workshop {B}lackbox{NLP}: Analyzing and Interpreting Neural Networks for {NLP}}, pages 325--327, Brussels, Belgium, November 2018. Association for Computational Linguistics.
\newblock \doi{10.18653/v1/W18-5437}.
\newblock URL \url{https://www.aclweb.org/anthology/W18-5437}.

\bibitem[Prasanna et~al.(2020)Prasanna, Rogers, and Rumshisky]{prasanna-etal-2020-bert}
Sai Prasanna, Anna Rogers, and Anna Rumshisky.
\newblock {W}hen {BERT} {P}lays the {L}ottery, {A}ll {T}ickets {A}re {W}inning.
\newblock In \emph{Proceedings of the 2020 Conference on Empirical Methods in Natural Language Processing (EMNLP)}, pages 3208--3229, Online, November 2020. Association for Computational Linguistics.
\newblock \doi{10.18653/v1/2020.emnlp-main.259}.
\newblock URL \url{https://aclanthology.org/2020.emnlp-main.259}.

\bibitem[Qian et~al.(2016{\natexlab{a}})Qian, Qiu, and Huang]{qian-qiu-huang:2016:EMNLP2016}
Peng Qian, Xipeng Qiu, and Xuanjing Huang.
\newblock {Analyzing Linguistic Knowledge in Sequential Model of Sentence}.
\newblock In \emph{Proceedings of the 2016 Conference on Empirical Methods in Natural Language Processing}, pages 826--835, Austin, Texas, November 2016{\natexlab{a}}. Association for Computational Linguistics.
\newblock URL \url{https://aclweb.org/anthology/D16-1079}.

\bibitem[Qian et~al.(2016{\natexlab{b}})Qian, Qiu, and Huang]{qian-qiu-huang:2016:P16-11}
Peng Qian, Xipeng Qiu, and Xuanjing Huang.
\newblock {Investigating Language Universal and Specific Properties in Word Embeddings}.
\newblock In \emph{Proceedings of the 54th Annual Meeting of the Association for Computational Linguistics (Volume 1: Long Papers)}, pages 1478--1488, Berlin, Germany, August 2016{\natexlab{b}}. Association for Computational Linguistics.
\newblock URL \url{http://www.aclweb.org/anthology/P16-1140}.

\bibitem[Rajpurkar et~al.(2016)Rajpurkar, Zhang, Lopyrev, and Liang]{rajpurkar-etal-2016-squad}
Pranav Rajpurkar, Jian Zhang, Konstantin Lopyrev, and Percy Liang.
\newblock {SQ}u{AD}: 100,000+ questions for machine comprehension of text.
\newblock In \emph{Proceedings of the 2016 Conference on Empirical Methods in Natural Language Processing}, pages 2383--2392, Austin, Texas, November 2016. Association for Computational Linguistics.
\newblock \doi{10.18653/v1/D16-1264}.
\newblock URL \url{https://www.aclweb.org/anthology/D16-1264}.

\bibitem[Rangamani et~al.(2023)Rangamani, Galanti, and Poggio]{neuralCollapse}
Marius Rangamani, Akshay;~Lindegaard, Tomer Galanti, and Tomaso Poggio.
\newblock Feature learning in deep classifiers through intermediate neural collapse.
\newblock 2023.

\bibitem[Rethmeier et~al.(2020)Rethmeier, Saxena, and Augenstein]{rethmeier2019txray}
Nils Rethmeier, Vageesh~Kumar Saxena, and Isabelle Augenstein.
\newblock Tx-ray: Quantifying and explaining model-knowledge transfer in (un-)supervised {NLP}.
\newblock In Ryan~P. Adams and Vibhav Gogate, editors, \emph{Proceedings of the Thirty-Sixth Conference on Uncertainty in Artificial Intelligence, {UAI} 2020, virtual online, August 3-6, 2020}, page 197. {AUAI} Press, 2020.
\newblock URL \url{http://www.auai.org/uai2020/proceedings/197\_main\_paper.pdf}.

\bibitem[Sajjad et~al.(2022)Sajjad, Durrani, Dalvi, Alam, Khan, and Xu]{sajjad-etal-2022-analyzing}
Hassan Sajjad, Nadir Durrani, Fahim Dalvi, Firoj Alam, Abdul Khan, and Jia Xu.
\newblock Analyzing encoded concepts in transformer language models.
\newblock In \emph{Proceedings of the 2022 Conference of the North American Chapter of the Association for Computational Linguistics: Human Language Technologies}, pages 3082--3101, Seattle, United States, July 2022. Association for Computational Linguistics.
\newblock \doi{10.18653/v1/2022.naacl-main.225}.
\newblock URL \url{https://aclanthology.org/2022.naacl-main.225}.

\bibitem[Sajjad et~al.(2023)Sajjad, Dalvi, Durrani, and Nakov]{sajjad2023:csl}
Hassan Sajjad, Fahim Dalvi, Nadir Durrani, and Preslav Nakov.
\newblock On the effect of dropping layers of pre-trained transformer models.
\newblock \emph{Computer Speech and Language}, 77\penalty0 (C):\penalty0 101429, 2023.
\newblock ISSN 0885-2308.
\newblock \doi{https://doi.org/10.1016/j.csl.2022.101429}.
\newblock URL \url{https://www.sciencedirect.com/science/article/pii/S0885230822000596}.

\bibitem[Shi et~al.(2016)Shi, Padhi, and Knight]{shi-padhi-knight:2016:EMNLP2016}
Xing Shi, Inkit Padhi, and Kevin Knight.
\newblock Does string-based neural {MT} learn source syntax?
\newblock In \emph{Proceedings of the 2016 Conference on Empirical Methods in Natural Language Processing}, EMNLP~'16, pages 1526--1534, Austin, TX, USA, 2016.

\bibitem[Simonyan et~al.(2014)Simonyan, Vedaldi, and Zisserman]{saliency}
Karen Simonyan, Andrea Vedaldi, and Andrew Zisserman.
\newblock Deep inside convolutional networks: Visualising image classification models and saliency maps, 2014.

\bibitem[Socher et~al.(2013)Socher, Perelygin, Wu, Chuang, Manning, Ng, and Potts]{socher-etal-2013-recursive}
Richard Socher, Alex Perelygin, Jean Wu, Jason Chuang, Christopher~D. Manning, Andrew Ng, and Christopher Potts.
\newblock Recursive deep models for semantic compositionality over a sentiment treebank.
\newblock In \emph{Proceedings of the 2013 Conference on Empirical Methods in Natural Language Processing}, pages 1631--1642, Seattle, Washington, USA, October 2013. Association for Computational Linguistics.
\newblock URL \url{https://www.aclweb.org/anthology/D13-1170}.

\bibitem[Stańczak et~al.(2023)Stańczak, Hennigen, Williams, Cotterell, and Augenstein]{stanczak2023latentvariable}
Karolina Stańczak, Lucas~Torroba Hennigen, Adina Williams, Ryan Cotterell, and Isabelle Augenstein.
\newblock A latent-variable model for intrinsic probing, 2023.

\bibitem[Suau et~al.(2020)Suau, Zappella, and Apostoloff]{suau2020finding}
Xavier Suau, Luca Zappella, and Nicholas Apostoloff.
\newblock Finding experts in transformer models.
\newblock \emph{CoRR}, abs/2005.07647, 2020.
\newblock URL \url{https://arxiv.org/abs/2005.07647}.

\bibitem[Sundararajan et~al.(2017)Sundararajan, Taly, and Yan]{ig}
Mukund Sundararajan, Ankur Taly, and Qiqi Yan.
\newblock Axiomatic attribution for deep networks, 2017.

\bibitem[Tenney et~al.(2019)Tenney, Das, and Pavlick]{tenney-etal-2019-bert}
Ian Tenney, Dipanjan Das, and Ellie Pavlick.
\newblock {BERT} rediscovers the classical {NLP} pipeline.
\newblock In \emph{Proceedings of the 57th Annual Meeting of the Association for Computational Linguistics}, pages 4593--4601, Florence, Italy, July 2019. Association for Computational Linguistics.
\newblock \doi{10.18653/v1/P19-1452}.
\newblock URL \url{https://www.aclweb.org/anthology/P19-1452}.

\bibitem[Tjong Kim~Sang and Buchholz(2000)]{tjong-kim-sang-buchholz-2000-introduction}
Erik~F. Tjong Kim~Sang and Sabine Buchholz.
\newblock Introduction to the {C}o{NLL}-2000 shared task chunking.
\newblock In \emph{Fourth Conference on Computational Natural Language Learning and the Second Learning Language in Logic Workshop}, 2000.
\newblock URL \url{https://www.aclweb.org/anthology/W00-0726}.

\bibitem[Torroba~Hennigen et~al.(2020)Torroba~Hennigen, Williams, and Cotterell]{torroba-hennigen-etal-2020-intrinsic}
Lucas Torroba~Hennigen, Adina Williams, and Ryan Cotterell.
\newblock Intrinsic probing through dimension selection.
\newblock In \emph{Proceedings of the 2020 Conference on Empirical Methods in Natural Language Processing (EMNLP)}, pages 197--216, Online, November 2020. Association for Computational Linguistics.
\newblock \doi{10.18653/v1/2020.emnlp-main.15}.
\newblock URL \url{https://www.aclweb.org/anthology/2020.emnlp-main.15}.

\bibitem[Tsang et~al.(2020)Tsang, Rambhatla, and Liu]{achipelago}
Michael Tsang, Sirisha Rambhatla, and Yan Liu.
\newblock How does this interaction affect me? interpretable attribution for feature interactions, 2020.

\bibitem[Valipour et~al.(2019)Valipour, Lee, Jamacaro, and Bessega]{valipur-2019}
Mehrdad Valipour, En{-}Shiun~Annie Lee, Jaime~R. Jamacaro, and Carolina Bessega.
\newblock Unsupervised transfer learning via {BERT} neuron selection.
\newblock \emph{CoRR}, abs/1912.05308, 2019.
\newblock URL \url{http://arxiv.org/abs/1912.05308}.

\bibitem[Voita et~al.(2019)Voita, Talbot, Moiseev, Sennrich, and Titov]{voita-etal-2019-analyzing}
Elena Voita, David Talbot, Fedor Moiseev, Rico Sennrich, and Ivan Titov.
\newblock Analyzing multi-head self-attention: Specialized heads do the heavy lifting, the rest can be pruned.
\newblock In \emph{Proceedings of the 57th Annual Meeting of the Association for Computational Linguistics}, pages 5797--5808, Florence, Italy, July 2019. Association for Computational Linguistics.
\newblock \doi{10.18653/v1/P19-1580}.
\newblock URL \url{https://www.aclweb.org/anthology/P19-1580}.

\bibitem[Vylomova et~al.(2016)Vylomova, Cohn, He, and Haffari]{vylomova2016word}
Ekaterina Vylomova, Trevor Cohn, Xuanli He, and Gholamreza Haffari.
\newblock {Word Representation Models for Morphologically Rich Languages in Neural Machine Translation}.
\newblock \emph{arXiv preprint arXiv:1606.04217}, 14\penalty0 (14), 2016.

\bibitem[Wambsganss et~al.(2021)Wambsganss, Engel, and Fromm]{Wambsganss2021ImprovingEA}
Thiemo Wambsganss, Christiane Engel, and Hansj{\"o}rg Fromm.
\newblock Improving explainability and accuracy through feature engineering: A taxonomy of features in nlp-based machine learning.
\newblock In \emph{International Conference on Interaction Sciences}, 2021.

\bibitem[Wang et~al.(2018)Wang, Singh, Michael, Hill, Levy, and Bowman]{wang-etal-2018-glue}
Alex Wang, Amanpreet Singh, Julian Michael, Felix Hill, Omer Levy, and Samuel Bowman.
\newblock {GLUE}: A multi-task benchmark and analysis platform for natural language understanding.
\newblock In \emph{Proceedings of the 2018 {EMNLP} Workshop {B}lackbox{NLP}: Analyzing and Interpreting Neural Networks for {NLP}}, pages 353--355, Brussels, Belgium, November 2018. Association for Computational Linguistics.
\newblock \doi{10.18653/v1/W18-5446}.
\newblock URL \url{https://www.aclweb.org/anthology/W18-5446}.

\bibitem[Williams et~al.(2018)Williams, Nangia, and Bowman]{williams-etal-2018-broad}
Adina Williams, Nikita Nangia, and Samuel Bowman.
\newblock A broad-coverage challenge corpus for sentence understanding through inference.
\newblock In \emph{Proceedings of the 2018 Conference of the North {A}merican Chapter of the Association for Computational Linguistics: Human Language Technologies, Volume 1 (Long Papers)}, pages 1112--1122, New Orleans, Louisiana, June 2018. Association for Computational Linguistics.
\newblock \doi{10.18653/v1/N18-1101}.
\newblock URL \url{https://www.aclweb.org/anthology/N18-1101}.

\bibitem[Yang et~al.(2019)Yang, Dai, Yang, Carbonell, Salakhutdinov, and Le]{yang2019xlnet}
Zhilin Yang, Zihang Dai, Yiming Yang, Jaime~G. Carbonell, Ruslan Salakhutdinov, and Quoc~V. Le.
\newblock Xlnet: Generalized autoregressive pretraining for language understanding.
\newblock In Hanna~M. Wallach, Hugo Larochelle, Alina Beygelzimer, Florence d'Alch{\'{e}}{-}Buc, Emily~B. Fox, and Roman Garnett, editors, \emph{Advances in Neural Information Processing Systems 32: Annual Conference on Neural Information Processing Systems 2019, NeurIPS 2019, 8-14 December 2019, Vancouver, BC, Canada}, pages 5754--5764, 2019.
\newblock URL \url{http://papers.nips.cc/paper/8812-xlnet-generalized-autoregressive-pretraining-for-language-understanding}.

\bibitem[Zou and Hastie(2005)]{Zou05regularizationand}
Hui Zou and Trevor Hastie.
\newblock Regularization and variable selection via the elastic net.
\newblock \emph{Journal of the Royal Statistical Society, Series B}, 67:\penalty0 301--320, 2005.

\end{thebibliography}

\label{lastpage}

%\acks{}


\newpage

\appendix
\section*{Appendix}

\subsection*{Appendix A: Data and Representations}
\label{sec:appendix:data}

\begin{figure*}[t]
    \centering
    \begin{subfigure}[b]{0.32\linewidth}
    \centering
    \includegraphics[width=\linewidth]{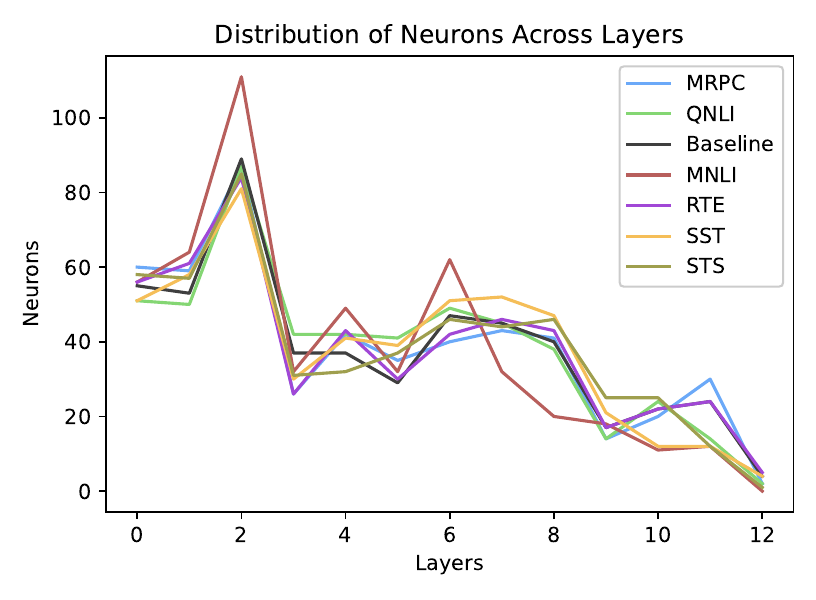}
    \caption{BERT -- POS}
    \label{fig:bert_pos_neurons-A}
    \end{subfigure}
    \begin{subfigure}[b]{0.32\linewidth}
    \centering
    \includegraphics[width=\linewidth]{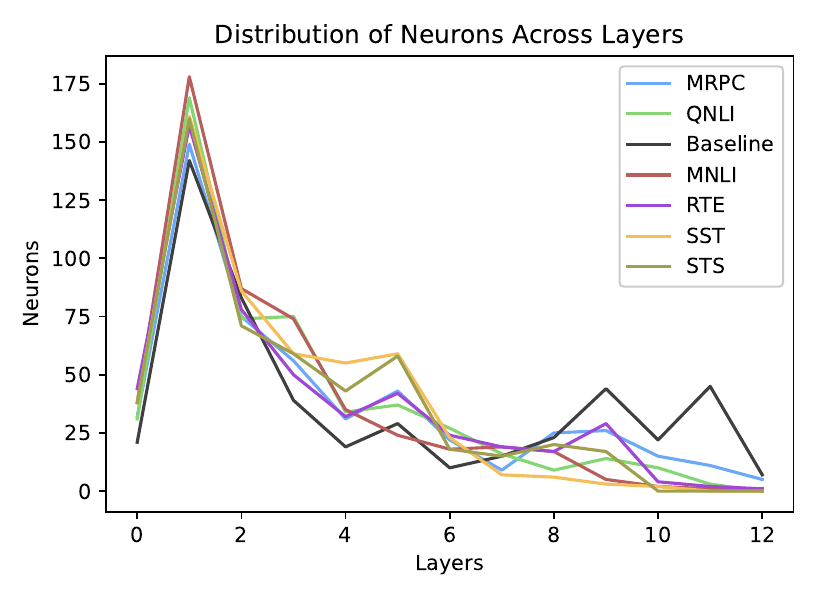}
    \caption{RoBERTa -- POS}
    \label{fig:roberta_pos_neurons-A}
    \end{subfigure}
    \begin{subfigure}[b]{0.32\linewidth}
    \centering
    \includegraphics[width=\linewidth]{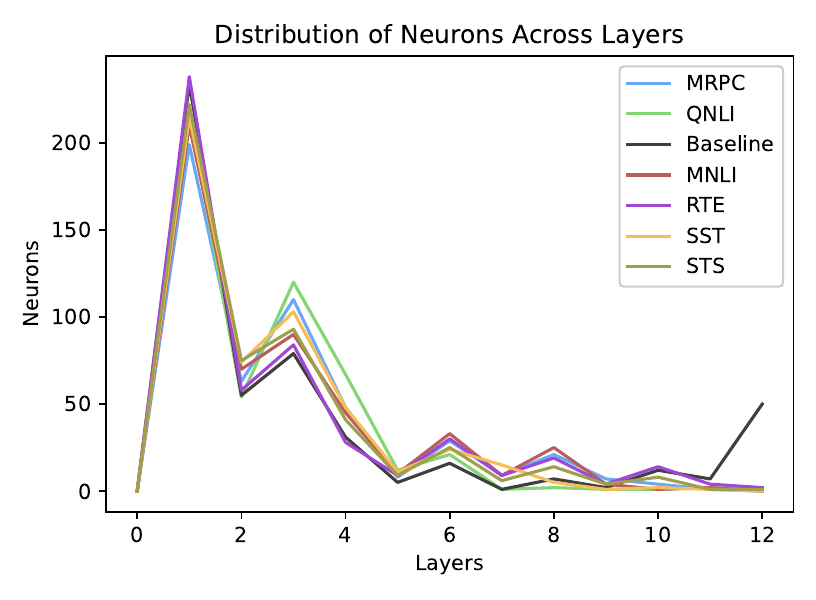}
    \caption{XLNet -- POS}
    \label{fig:xlnet_pos_neurons-A}
    \end{subfigure}
    \centering
    \begin{subfigure}[b]{0.32\linewidth}
    \centering
    \includegraphics[width=\linewidth]{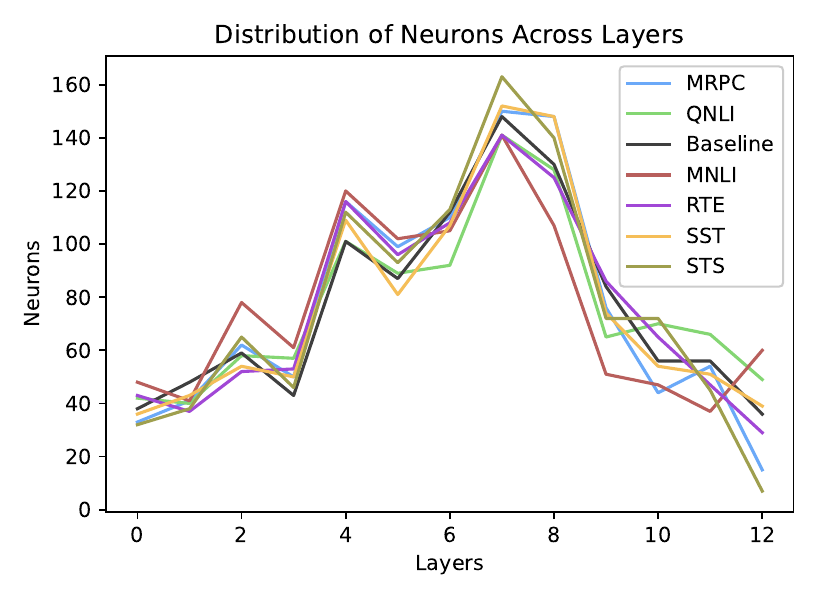}
    \caption{BERT -- Chunking}
    \label{fig:bert_chunking_neurons-A}
    \end{subfigure}
    \begin{subfigure}[b]{0.32\linewidth}
    \centering
    \includegraphics[width=\linewidth]{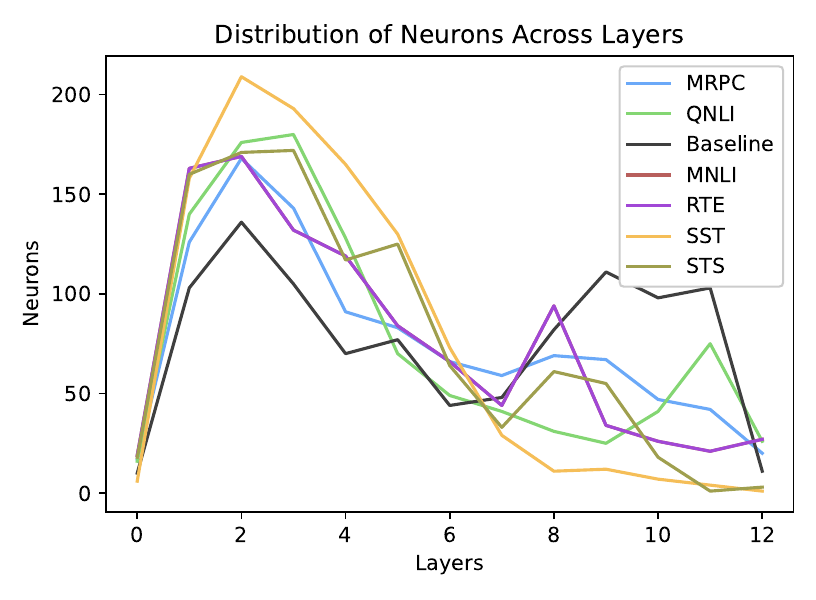}
    \caption{RoBERTa -- Chunking}
    \label{fig:roberta_chunking_neurons-A}
    \end{subfigure}
    \begin{subfigure}[b]{0.32\linewidth}
    \centering
    \includegraphics[width=\linewidth]{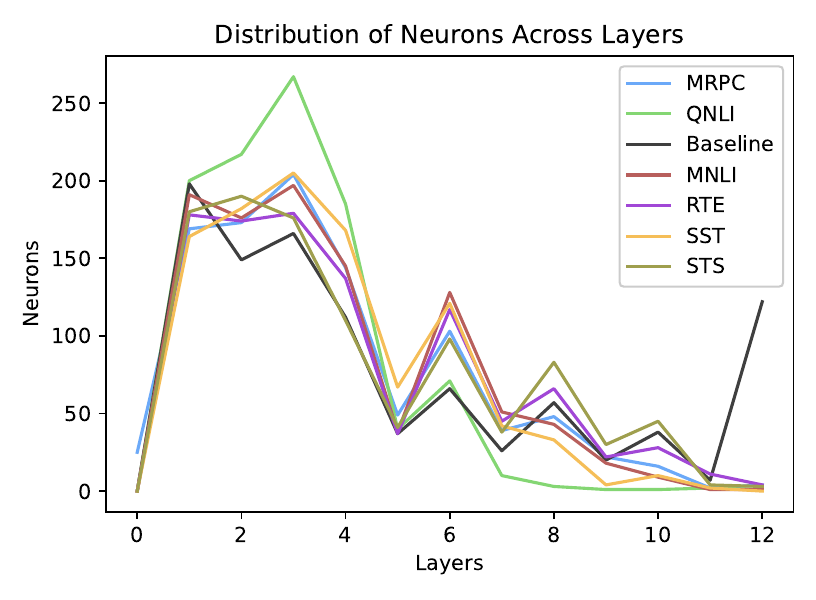}
    \caption{XLNet -- Chunking}
    \label{fig:xlnet_chunking_neurons-A}
    \end{subfigure}
    \centering
    \begin{subfigure}[b]{0.32\linewidth}
    \centering
    \includegraphics[width=\linewidth]{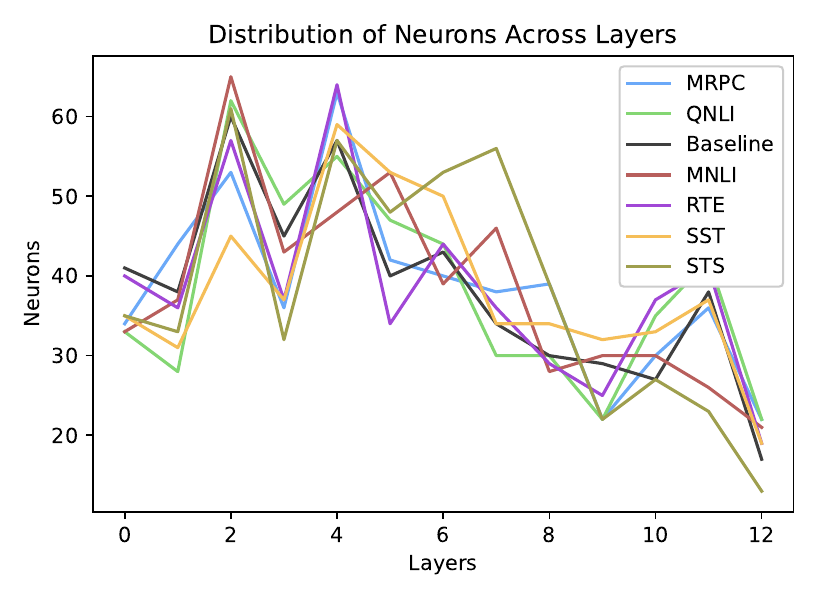}
    \caption{BERT -- SEM}
    \label{fig:bert_sem_neurons-A}
    \end{subfigure}
    \begin{subfigure}[b]{0.32\linewidth}
    \centering
    \includegraphics[width=\linewidth]{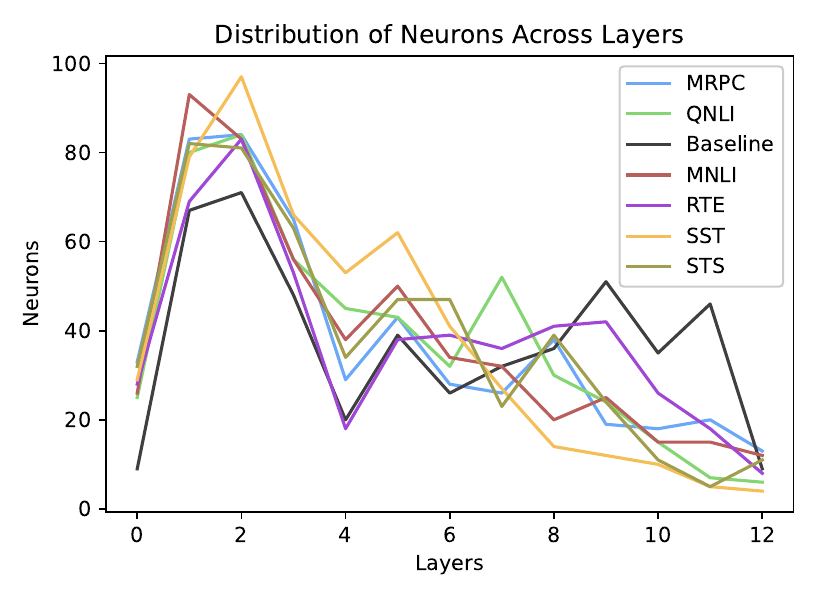}
    \caption{RoBERTa -- SEM}
    \label{fig:roberta_sem_neurons-A}
    \end{subfigure}
    \begin{subfigure}[b]{0.32\linewidth}
    \centering
    \includegraphics[width=\linewidth]{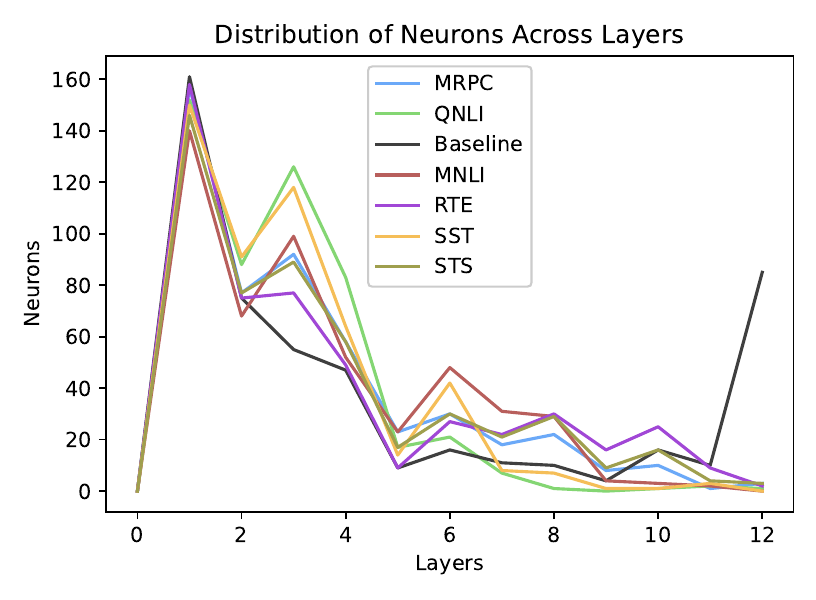}
    \caption{XLNet -- SEM}
    \label{fig:xlnet_sem_neurons-A}
    \end{subfigure}
    \caption{Distribution of Top Neurons across Layers.}
    \label{fig:layerwise_neurons_appendix}
%\vspace{-10pt}
\end{figure*}

We used standard splits for training, development and test data for the 4 linguistic tasks (POS, SEM, Chunking and CCG super tagging) that we used to carry out our analysis on. The splits to preprocess the data are available through git repository\footnote{\url{https://github.com/nelson-liu/contextual-repr-analysis}} released with \cite{liu-etal-2019-linguistic}. See Table \ref{tab:dataStats} for statistics. We obtained the understudied pre-trained models from the authors of the paper, through personal communication.

\begin{table}[ht]                                   
\centering  
    \caption{Data statistics (number of sentences) on training, development and test sets using in the experiments and the number of tags to be predicted.}
%\footnotesize
%\resizebox{\columnwidth}{!}{                                   
    \begin{tabular}{l|cccc}                                 
    \toprule                                    
Task    & Train & Dev & Test & Tags \\      
\midrule
    Suffix & 40000 & 5000 & 5000 & 58 \\
    POS & 36557 & 1802 & 1963 & 44 \\
    SEM & 36928 & 5301 & 10600 & 73 \\
    Chunking &  8881 &  1843 &  2011 & 22 \\
    CCG &  39101 & 1908 & 2404 & 1272 \\
    \bottomrule
    \end{tabular}
    % \caption{Data statistics (number of sentences) on training, development and test sets using in the experiments and the number of tags to be predicted}
    \label{tab:dataStats}                       
\end{table}

\begin{table}[ht]                                   
\centering  
\caption{Data statistics (number of sentences) on training, development and test sets using in the experiments and the number of POS tags and Syntactic Dependency Relations to be predicted in multilingual experiments.}
%\footnotesize
%\resizebox{\columnwidth}{!}{                                   
    \begin{tabular}{l|cccc}                                 
    \toprule                                    
Task    & Train & Dev & Test & Tags \\      
\midrule
    POS (en) & 14498 & 3000 & 8172 &  44\\
    POS (de) & 14498 & 3000 & 8172 &  52 \\
    POS (fr) & 11495 & 3000  & 3003 & 13 \\
    \midrule
    Syntactic Dependency (en) & 11663 & 1914 & 3828 &  49\\
    Syntactic Dependency (de) & 14118 & 1775 & 1776 &  35 \\
    Syntactic Dependency (fr) & 14552 & 1895  & 1894 & 40 \\
    \bottomrule
    \end{tabular}
    % \caption{Data statistics (number of sentences) on training, development and test sets using in the experiments and the number of POS tags and Syntactic Dependency Relations to be predicted in multilingual experiments}
\label{tab:dataStats2}                      
\end{table}

\subsection*{Appendix B: Spread of Neurons}
\label{sec:appendix:neuronSpread}

\begin{figure*}[t]
    \centering
    \begin{subfigure}[b]{0.65\linewidth}
    \centering
    \includegraphics[width=\linewidth]{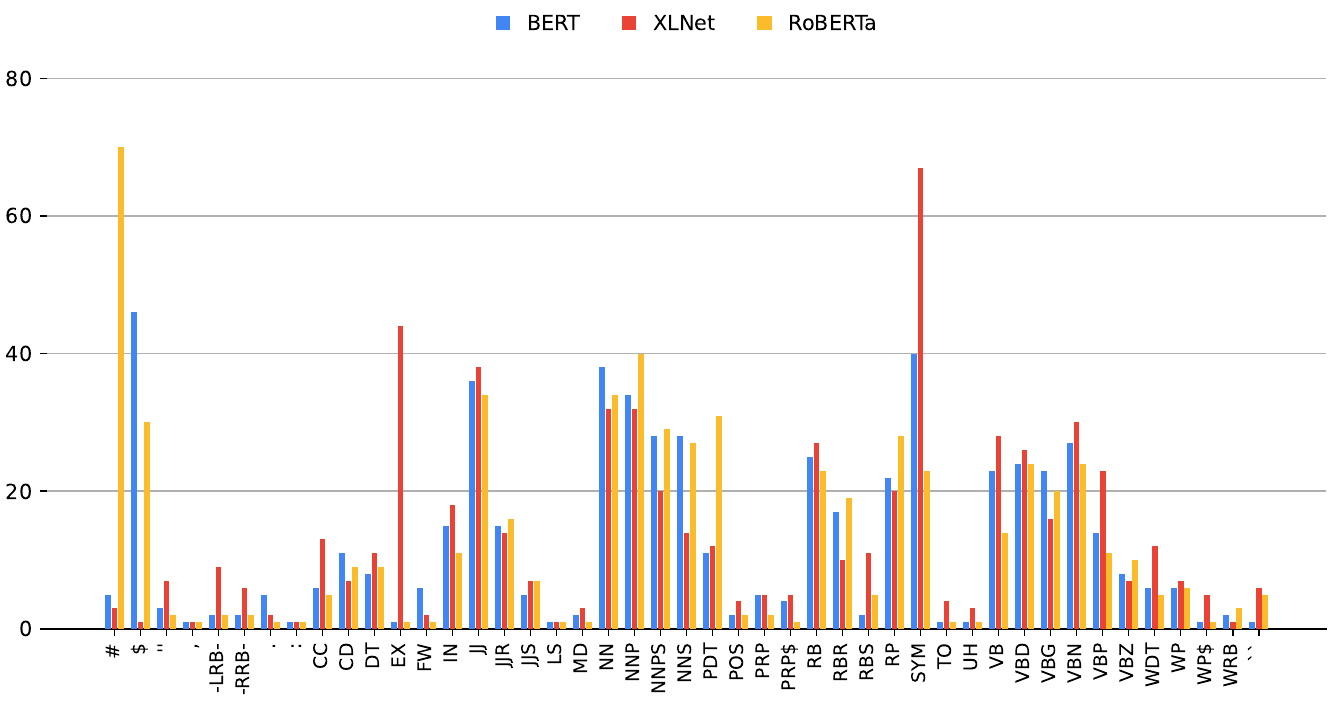}
    \caption{POS}
    \label{fig:pos_neurons-A}
    \end{subfigure}
    \begin{subfigure}[b]{0.65\linewidth}
    \centering
    \includegraphics[width=\linewidth]{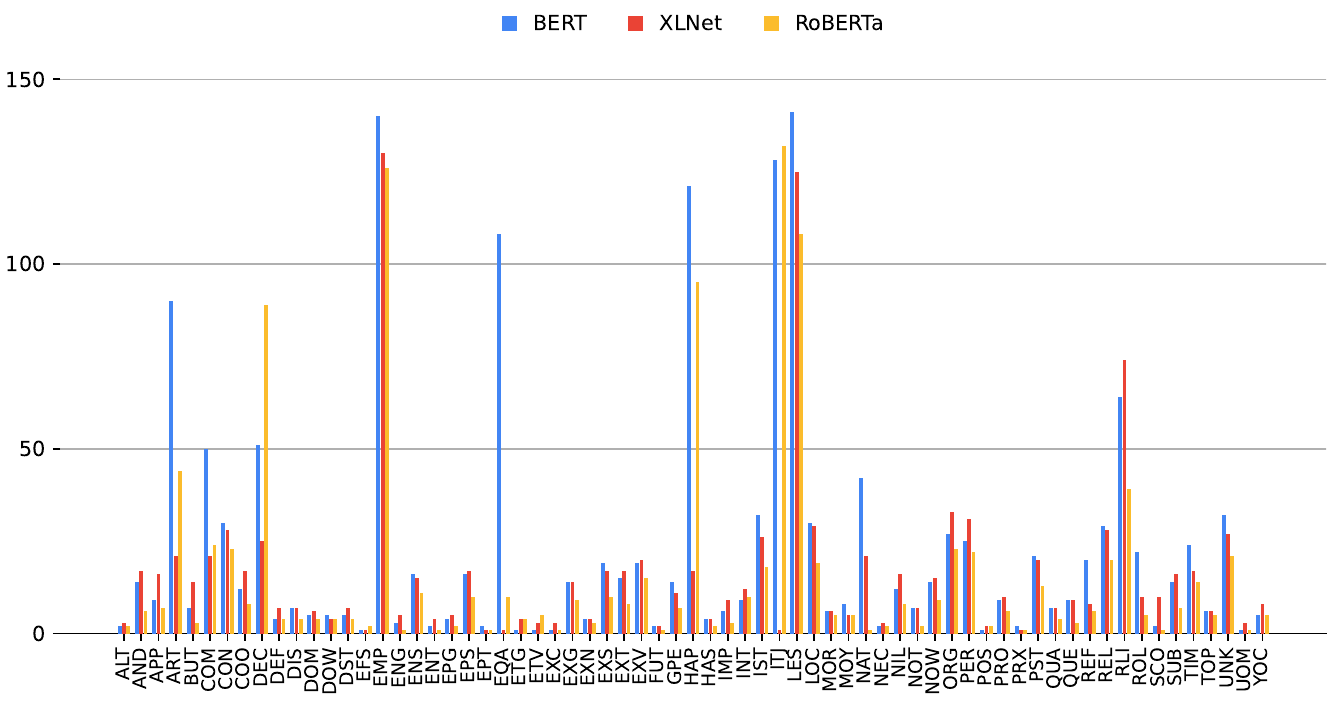}
    \caption{SEM}
    \label{fig:sem_neurons-A}
    \end{subfigure}
    \begin{subfigure}[b]{0.65\linewidth}
    \centering
    \includegraphics[width=\linewidth]{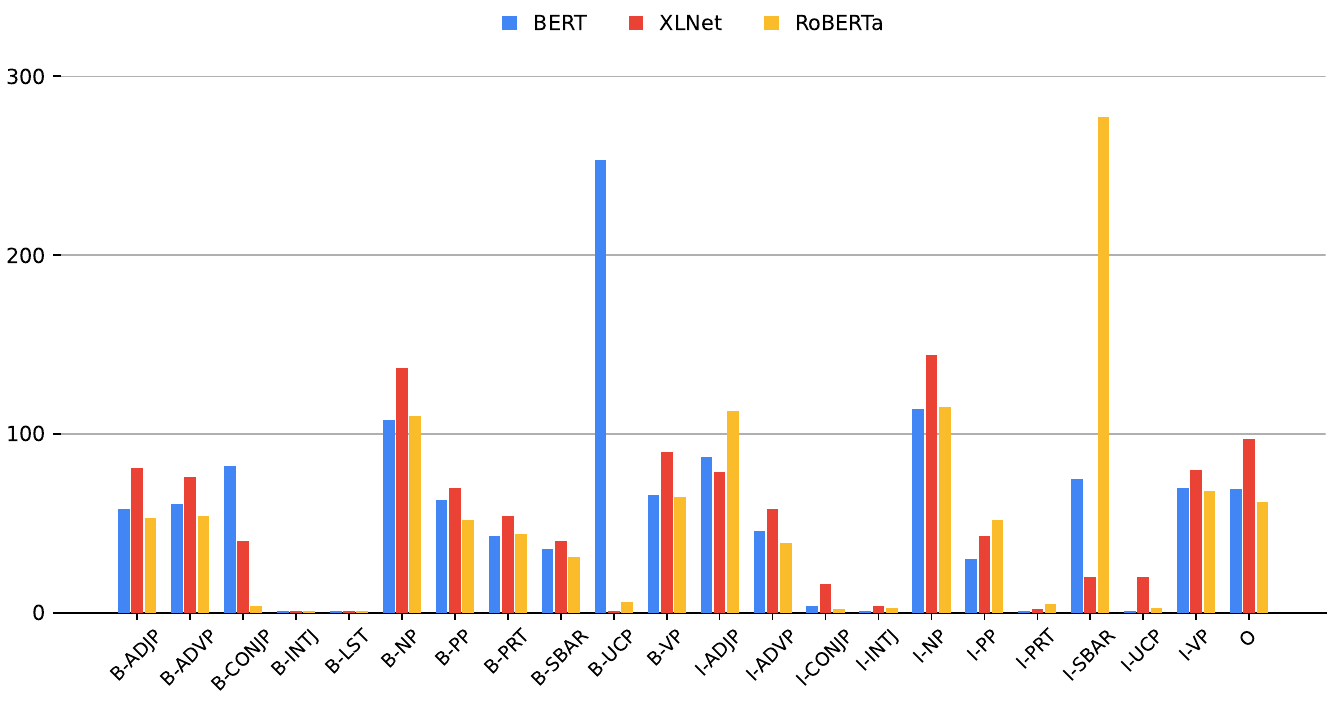}
    \caption{Chunking}
    \label{fig:chunking_neurons-A}
    \end{subfigure}
    \caption{Distribution of Top Neurons Properties.}
    \label{fig:propertyWise_neurons_appendix}
%\vspace{-10pt}
\end{figure*}

\begin{figure*}[t]
    \centering
    \begin{subfigure}[b]{0.95\linewidth}
    \centering
    \includegraphics[width=\linewidth]{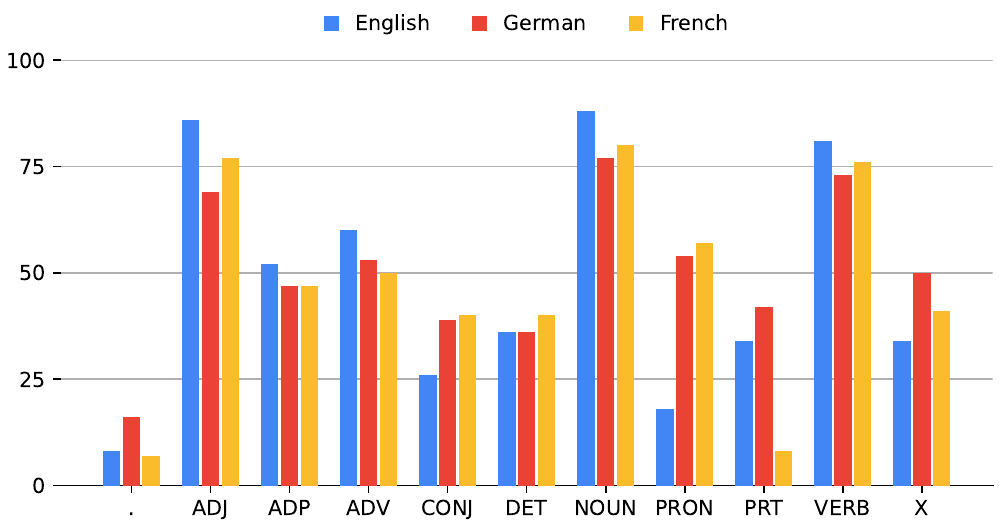}
    \caption{XLM-RoBERTa -- POS tagging}
    \label{fig:xlm-r_neurons}
    \end{subfigure}
    \begin{subfigure}[b]{0.95\linewidth}
    \centering
    \includegraphics[width=\linewidth]{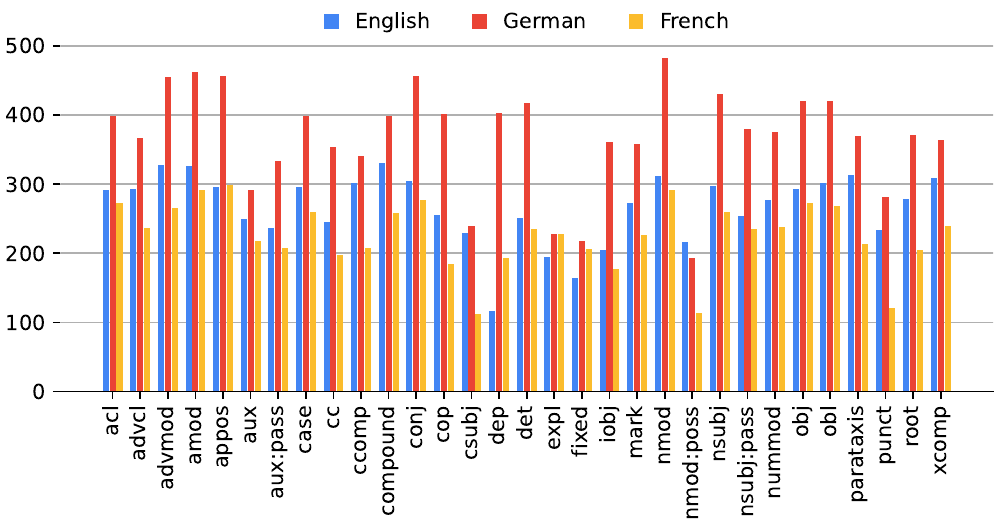}
    \caption{XLM-RoBERTa -- Syntactic Dependency Labeling}
    \label{fig:xlm-r_neurons_syn}
    \end{subfigure}
\caption{Distribution of Top Neurons Across POS (a) and Syntactic Dependency Properties in XLM-Roberta.}
    \label{fig:propertyWise_neurons_multilingual_xlm-r}
%\vspace{-10pt}
\end{figure*}

In section \ref{sec:layerWise} we presented how distributed or localized neurons are across different properties. We showed that some properties, including closed class words are captured in fewer neurons where as open class words or words that occur in many varying contexts require more neruons. See Figure \ref{fig:propertyWise_neurons_appendix} for all the properties. In section \ref{sec:multilingual} we presented if properties within a task behave similarly across different languages and architectures. In Figure \ref{fig:propertyWise_neurons_multilingual_xlm-r} we present results on XLM-RoBERTa where we see a similar evidence as seen in the main results.

\subsection*{Appendix C: Transfer Learning}
\label{sec:appendix:transferLearning}

Section \ref{sec:fineTuning} presented neuron-wise probing results for selected GLUE tasks. We see a similar pattern across architectures as in Figure \ref{fig:layerwise_neurons_appendix}. As the model is fine-tuned towards downstream, number of salient neurons towards a linguistic property, in the lower layers increase.

% Note: in this sample, the section number is hard-coded in. Following
% proper LaTeX conventions, it should properly be coded as a reference:

%In this appendix we prove the following theorem from
%Section~\ref{sec:textree-generalization}:

\end{document}